%% file: preprint/preprint.tex
\documentclass{article} 
\usepackage{preprint/arxiv}

\input{packages_and_commands}

\usepackage{hyperref}
\usepackage{url}
\usepackage{multirow}
\usepackage{placeins}
\usepackage{bbm}

\title{Improved probabilistic regression using\\  diffusion models}

\author{Carlo Kneissl$^*$, Christopher Bülte$^*$, Philipp Scholl\thanks{Equal contribution.} \\
 Ludwig-Maximilians-Universit\"at M\"unchen \\
     Munich Center for Machine Learning (MCML) \\
 	Munich, Germany\\
\texttt{\{kneissl, buelte, scholl\}@math.lmu.de} \\
\AND
  {Gitta Kutyniok}\\
  Ludwig-Maximilians-Universit\"at M\"unchen \\
      University of Troms\o{} \\
  DLR-German Aerospace Center \\
  Munich Center for Machine Learning (MCML) \\
	Munich, Germany\\
	\texttt{kutyniok@math.lmu.de} \\
}

\date{}

\renewcommand{\headerleft}{Improved probabilistic regression using diffusion models}
\renewcommand{\headeright}{}
\renewcommand{\undertitle}{}
\renewcommand{\shorttitle}{}

\begin{document}

\maketitle

\begin{abstract}
Probabilistic regression models the entire predictive distribution of a response variable, offering richer insights than classical point estimates and directly allowing for uncertainty quantification. While diffusion-based generative models have shown remarkable success in generating complex, high-dimensional data, their usage in general regression tasks often lacks uncertainty-related evaluation and remains limited to domain-specific applications. We propose a novel diffusion-based framework for probabilistic regression that learns predictive distributions in a nonparametric way. More specifically, we propose to model the full distribution of the diffusion noise, enabling adaptation to diverse tasks and enhanced uncertainty quantification. We investigate different noise parameterizations, analyze their trade-offs, and evaluate our framework across a broad range of regression tasks, covering low- and high-dimensional settings. For several experiments, our approach shows superior performance against existing baselines, while delivering calibrated uncertainty estimates, demonstrating its versatility as a tool for probabilistic prediction.
\end{abstract}

\section{Introduction}

Supervised regression concerns predicting a response variable $\bm y \in \mathcal{Y}$ from covariates $\bm c \in \mathcal{C}$. Classical approaches estimate the conditional mean $\mathbb{E}[\bm y \mid \bm c]$, whereas probabilistic regression models the full predictive distribution $p_\mathcal{Y}(\bm y \mid \bm c)$ \citep{bishop, 10.1093/jrsssb/qkae108}. The latter provides calibrated uncertainty estimates and captures complex data-generating processes, including multimodal outcomes and nonlinear noise. We adopt a broad notion of regression, encompassing heterogeneous data types such as scalars, images, and trajectories. A key modeling choice is the specification of $p_\mathcal{Y}(\cdot \mid \bm c)$: while Gaussian assumptions remain widespread \citep{lakshminarayanan_simple_2017, 374138}, recent work has emphasized more flexible non-parametric and distribution-free alternatives \citep{kelenDistributionFreeDataUncertainty2024a, 10.1093/jrsssb/qkae108}.

Diffusion-based generative models have emerged as state-of-the-art approaches for high-dimensional data generation, achieving remarkable results in tasks such as photorealistic image \citep{rombach2022high} and video synthesis \citep{ho2022videodiffusionmodels}. These models are typically formulated as diffusion probabilistic models \citep{sohl-dicksteinDeepUnsupervisedLearning2015, hoDenoisingDiffusionProbabilistic2020}, where data are gradually perturbed by Gaussian noise (the forward process) and new samples are generated via the time-reversed dynamics. Prior work has advanced diffusion models through improved noise schedules \citep{nicholImprovedDenoisingDiffusion2021, karrasElucidatingDesignSpace2022} and sampling algorithms \citep{songDenoisingDiffusionImplicit2022, bortoli2025distributional}, where the focus has largely been on accelerating the diffusion process while achieving a high generational quality \citep{NEURIPS2019_3001ef25, nicholImprovedDenoisingDiffusion2021}.


Recently, diffusion models have been applied to various regression tasks such as depth estimation \citep{ke2024repurposing}, autoregressive flow prediction \citep{kohl_benchmarking_2024, pmlr-v202-finzi23a}, and weather forecasting \citep{priceProbabilisticWeatherForecasting2025, couairon2024archesweatherarchesweathergendeterministic}, often achieving state-of-the-art performance. Despite their inherently probabilistic nature, evaluations rarely emphasize uncertainty-related metrics or calibration. Recent efforts that aim to extract uncertainty estimates from diffusion models \citep{shu2024zeroshotuncertaintyquantificationusing, chan2024estimating, berry2024shedding} typically rely on training multiple networks, incurring substantial computational overhead. Furthermore, the intimate relation between uncertainty quantification and the noise modeling within the diffusion process remains underexplored.


\textbf{Contributions: }In this work, we address these limitations by adapting the diffusion process to yield calibrated probabilistic predictions. Building on recent advances from generative modeling \citep{bortoli2025distributional}, we introduce a novel loss for diffusion-based regression models that enables learning a flexible noise distribution, moving beyond optimization of the conditional mean. We derive several trainable parameterizations that offer task-specific trade-offs between expressivity and computational efficiency while admitting closed-form sampling at the same time. These extensions provide principled epistemic uncertainty estimation within diffusion models, and we extensively validate our method across diverse regression tasks\textemdash{}including the UCI benchmark, flow prediction, weather forecasting, and depth estimation\textemdash{}showing consistent improvement in predictive performance and substantial gains in uncertainty quantification and calibration.



\section{Background}

\subsection{Probabilistic regression}
Let $\bm y \in \mathcal{Y} \subseteq \mathbb{R}^{d_y}$ denote the response variables of interest and $\bm c \in \mathcal{C} \subseteq \mathbb{R}^{d_c}$ the corresponding conditioning variables. Classical regression focuses on estimating the conditional mean $\mathbb{E}[\bm y \mid \bm c]$, whereas probabilistic regression seeks to model the full conditional distribution $p_{\mathcal{Y}}(\bm y \mid \bm c)$, thereby explicitly capturing predictive uncertainty. Given training data $\mathcal{D} = {(\bm c_i, \bm y_i)}_{i=1}^N$, the goal is to recover the predictive distribution $p_\mathcal{Y}(\bm y \mid \bm c, \mathcal{D})$.

The setting is deliberately chosen to be general, including classical regression approaches, but also autoregressive prediction tasks, for example, when choosing $\bm y = \bm y_{t+1}, \bm c = \bm y_{t}, \ \bm y_t \in \mathcal{Y}, \ \forall t=1,\ldots, T$. A common strategy is to parameterize the conditional distribution via a generative mapping
\begin{equation}
f_{\theta}: \mathcal{C} \times \mathcal{Z} \to \mathcal{Y}, \quad \theta \in \Theta \subseteq \mathbb{R}^p,
\end{equation}
where $\bm z \sim p_\mathcal{Z}$ is drawn from a tractable prior, typically Gaussian. The parameters $\theta$ are then optimized such that $f_{\theta}(\bm c, \cdot)\approx p_\mathcal{Y}(\cdot \mid \bm c)$.

In this work, we depart from classical parameterizations and instead employ diffusion-based generative models to represent $p_\mathcal{Y}(\bm y \mid \bm c, \mathcal{D})$. For notational simplicity, we omit the explicit dependence on $\mathcal{D}$ in the following.

\subsection{Diffusion models} \label{sec:background-diffusion-models}
Diffusion probabilistic models (DPMs) aim to learn a target distribution $p_0$ on $\mathbb{R}^d$ from samples by estimating the reverse dynamics of a diffusion process. We follow the non-Markovian formulation of denoising diffusion implicit models (DDIM) \citep{songDenoisingDiffusionImplicit2022}.  

Let $T \in \mathbb{N}$, $\bm x_0 \sim p_0$, and $\beta_{1:T} \in [0,1]^T$ denote a noise schedule. Define $\alpha_t \coloneqq 1-\beta_t$ and $\bar{\alpha}_t \coloneqq \prod_{i=1}^t \alpha_i$. The \emph{forward process} is specified as
$
    p(\bm x_{1:T} \mid \bm x_0) \coloneq p(\bm x_T \mid \bm x_0) \prod_{t=2}^T p(\bm x_{t-1} \mid \bm x_t, \bm x_0),
$
where, for $t > 1$,
\begin{equation}
\label{background:diffusion:reverse_process_def}
p(\bm x_{t-1} \mid \bm x_t, \bm x_0) = \cN(\underbrace{\sqrt{\bar{\alpha}_{t-1}}\bm x_0 + 
    \sqrt{ 1 - \bar{\alpha}_{t-1} - \sigma_t^2} \cdot 
    \frac{\bm x_t - \sqrt{\bar{\alpha}_t}\bm x_0}{\sqrt{1 - \bar{\alpha}_t}} }_{\eqcolon \bm \mu(\bm x_0, \bm x_t)}
    , \sigma_t^2 \mathbf{I}).
\end{equation}

We set $p(\bm x_T \mid \bm x_0) \coloneq \cN \left(\sqrt{\bar{\alpha}_T}\bm x_0, (1-\bar{\alpha}_T)\mathbf{I}\right)$ and require $\bar{\alpha}_T$ to be sufficiently small such that $p(\bm x_T \mid \bm x_0)\approx\cN(\bm 0, \mathbf{I})$.
The variance schedule is parameterized as
$\sigma_t \coloneq \eta \sqrt{\tilde{\beta}} \coloneq \eta \sqrt{\frac{1 - \bar{\alpha}_{t-1}}{1 - \bar{\alpha}_t} \beta_t}$
with $\eta \in [0,1]$ interpolating between the deterministic DDIM process ($\eta=0$) and the stochastic DDPM process ($\eta=1$) \citep{hoDenoisingDiffusionProbabilistic2020}.  

An important feature of this construction is that for all choices of $\sigma_t$ and for all $t=1, \ldots, T$,  we obtain the property
\begin{equation}    \label{background:diffusion:nice_property}
    p(\bm x_t \mid \bm x_0) = \cN \left(\sqrt{\bar{\alpha}_t} \bm x_0, (1-\bar{\alpha}_t) \mathbf{I}\right),
\end{equation}
which yields the identity
\begin{equation}
    \label{background:diffusion:rv_formula}
    \bm X_t = \sqrt{\bar{\alpha}_t} \bm X_0 + \sqrt{1 - \bar{\alpha}_t} \epsilon_t, \quad \epsilon_t \sim \cN(\bm 0, \bf{I}),
\end{equation}
for $\epsilon_t$ independent from $\bm X_0$, highlighting that $\bm X_t$ is defined by adding a specified amount of noise $\epsilon_t$ to the original input $\bm X_0$.


To generate samples, one must approximate the reverse process $p(\bm x_{t-1}\mid \bm x_t)$, which is intractable in general. DPMs approximate it with a latent-variable model
\begin{equation}
\label{eq:reverse-diffusion}
    p_\theta(\bm x_{0:T}) \coloneqq p_\theta(\bm x_T) \prod_{t=1}^T p_\theta(\bm x_{t-1} \mid \bm x_t),
\end{equation}
which is a Markov chain that samples from $\bm x_T$ to $\bm x_0$, that is referred to as the \emph{generative process}. For sufficiently small $\beta_t$, the reverse transition $p(\bm x_{t-1} \mid \bm x_t)$ is well approximated by a Gaussian, thus allowing to set $p_\theta(\bm x_T)\sim\mathcal{N}(\bm 0,\bm I)$ and specifying the latent variable model as a neural network, $p_\theta(\bm x_{t-1}| \bm x_t)=\mathcal{N}(\bm x_{t-1};\bm \mu_\theta(\bm x_t,t),\bm \Sigma_\theta(\bm x_t,t))$.

Training proceeds by minimizing the variational lower bound (VLB), which reduces to a KL divergence between $p_\theta(\bm x_{t-1}\mid \bm x_t)$ and the true posterior $p(\bm x_{t-1}\mid \bm x_t,\bm x_0)$. DDIM fixes the covariance to $\bm \Sigma_\theta = \sigma_t \mathbf{I}$, while DDPM considers $\bm \Sigma_\theta \in \{\beta_t \mathbf{I}, \tilde{\beta}_t \mathbf{I}\}$. With these simplifications, the VLB objective reduces (after reweighting) to a mean-squared error:

\begin{equation} \label{eq:L_VLB}
    L_{\text{simple}}(\theta) = 
    \mathbb{E}_{t, x_0, \epsilon_t} \left[ 
    \lVert\epsilon_t - \epsilon_\theta(\bm x_t,t)\rVert_2^2
    \right].
\end{equation}

To perform conditional modeling with covariates $\bm c$, the target distribution is set as $p_0 = p_\mathcal{Y}(\cdot \mid \bm c)$, yielding an approximate predictive distribution $p_\theta(\cdot \mid \bm c) \approx p_\mathcal{Y}(\cdot \mid \bm c)$. For simplicity, we omit explicit conditioning in the notation.  

While DDPM and DDIM have shown great success in conditional and unconditional generative modeling, note that due to the objective in Equation~\eqref{eq:L_VLB}, the latent variable model only learns to approximate $\epsilon_\theta(\bm x_t, t) \approx \mathbb{E} \left[ \epsilon_t \mid \bm x_t \right]$ and does not capture information about the full distribution $p(\epsilon_t \mid \bm x_t)$.

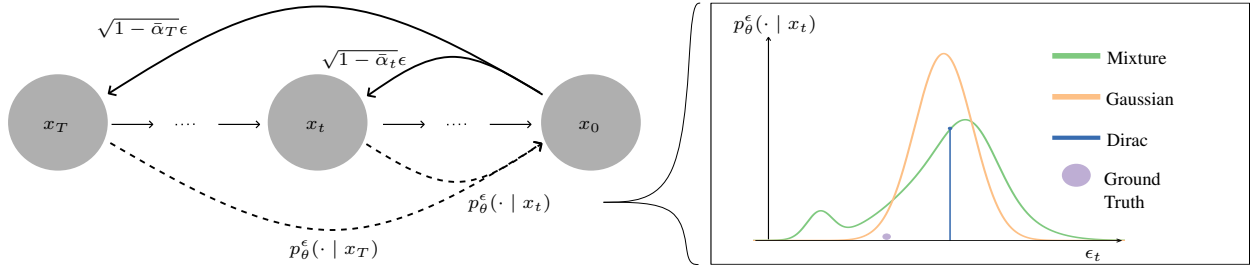
\begin{figure}[t]
    \scriptsize
  \centering
  \def\svgwidth{1.0\linewidth}
  \input{preprint/DistrDiffusionDrawing.tex}
  \caption{Method overview. At any given time $t = 1, \ldots, T$ in the diffusion process, we need a prediction of the noise $\epsilon_t$ given our current state $x_t$. Traditionally, this is achieved by a network that approximates the conditional mean $\mathbb{E}[\epsilon_t \mid x_t]$, which, when viewed under a probabilistic viewpoint, treats the distribution $p^\epsilon(\cdot \mid x_t)$ as a Dirac distribution. We propose to model this distribution by a Gaussian Mixture family.}
  \label{fig:overview}
\end{figure}
\section{Our methodology}
\input{ICLR_Submission/methodology.tex}

\section{Numerical experiments}
We now evaluate the proposed methodology across several regression tasks where diffusion models have previously demonstrated competitive or even state-of-the-art performance. Specifically, we consider (i) UCI regression benchmarks \citep{hanCARDClassificationRegression2022}, (ii) autoregressive prediction tasks \citep{rasulAutoregressiveDenoisingDiffusion, kohl_benchmarking_2024, priceProbabilisticWeatherForecasting2025}, and (iii) depth estimation \citep{ke2024repurposing}. 

For each task, we use our previously proposed instantiations: \emph{Univariate Gaussian} $(\bm\Sigma_\theta^{\mathrm{diag}})$, \emph{univariate Gaussian mixture} $(\bm\Sigma_\theta^{\mathrm{mix}})$, and \emph{multivariate Gaussian} ($\bm\Sigma_\theta^{\mathrm{mv}}$) with \emph{low-rank plus diagonal} due to its superior performance, see Appendix~\ref{app:hyperparameters}. As a baseline, we use established diffusion models $(\bm\delta_\theta)$ from the literature.
In addition, we compare against the sample-based approach ($\bm\epsilon_t^{\mathrm{ES}}$) of \citet{bortoli2025distributional}.
Where possible, we apply minor hyperparameter tuning (Appendix~\ref{app:hyperparameters}). Model performance is assessed using a variety of metrics, with particular emphasis on probabilistic calibration and distributional accuracy. Further experimental details are provided in Appendix~\ref{app:experiment_details}, and additional results and visualizations in Appendix~\ref{app:additional_results}. \footnote{The code for the experiments in Section~\ref{numerical_exps:uci} and \ref{sec:autoregressive-prediction-tasks} can be found in \url{https://github.com/Philipp238/diffusionUQ} and for Section~\ref{sec:depth-estimation} in \url{https://github.com/Philipp238/Marigold}.}

\subsection{UCI regression} \label{numerical_exps:uci}

We first evaluate our approach on the UCI regression benchmarks \citep{Dua:2019}, comparing it against CARD \citep{hanCARDClassificationRegression2022}, which serves as the baseline $\bm\delta_\theta$. Additional details on CARD are provided in Appendix~\ref{appendix:proofs}, and full experimental settings are given in Appendix~\ref{app:experiment_details:uci}.


\begin{table}[ht] 
\centering
\footnotesize
\caption{RMSE ($\downarrow$) and CRPS ($\downarrow$) on the UCI datasets of the different methods. 
For better readability, the results for Naval and Kin8nm have been scaled by a factor of 10000 and 100, respectively.}
\label{numerical_exps:uci:card_rmse_and_crps_table}
\begin{tabular}{|l|cccc|cccc|}
\toprule
 & \multicolumn{4}{c|}{RMSE} & \multicolumn{4}{c|}{CRPS} \\
\cmidrule(lr){2-5}\cmidrule(lr){6-9}
 & $\bm\delta_\theta$ & $\bm\Sigma_\theta^{\mathrm{diag}}$ & $\bm\Sigma_\theta^{\mathrm{mix}}$ & $\bm\epsilon_t^\mathrm{ES}$ & $\bm\delta_\theta$ & $\bm\Sigma_\theta^{\mathrm{diag}}$ & $\bm\Sigma_\theta^{\mathrm{mix}}$ & $\bm\epsilon_t^\mathrm{ES}$ \\
\midrule
Concrete & 4.81 & \textbf{4.72} & 4.73 & 4.81 & 2.60 & 2.46 & \textbf{2.45} & 2.50 \\
Energy   & 0.45 & 0.45 & \textbf{0.44} & 0.45 & 0.30 & 0.25 & \textbf{0.24} & 0.25 \\
Kin8nm   & 6.91 & \textbf{6.88} & 6.89 & 6.90 & 3.93 & \textbf{3.81} & 3.82 & 3.83 \\
Naval    & 1.35 & 1.24 & \textbf{1.23} & \textbf{1.23} & 0.79 & 0.56 & 0.54 & \textbf{0.53} \\
Power    & 3.86 & 3.64 & \textbf{3.59} & 3.78 & 2.04 & 1.84 & \textbf{1.81} & 1.93 \\
Protein  & 3.76 & \textbf{3.71} & 3.72 & 3.76 & 1.71 & 1.66 & \textbf{1.65} & 1.69 \\
Wine     & 0.67 & 0.67 & \textbf{0.66} & \textbf{0.66} & 0.34 & 0.34 & 0.34 & \textbf{0.33} \\
Yacht    & 0.74 & \textbf{0.70} & 0.78 & 0.80 & 0.36 & \textbf{0.28} & 0.31 & 0.32 \\
\midrule
Avg. Rank & 3.44 & 1.94 & \textbf{1.75} & 2.875 & 3.875 & 2.0625 & \textbf{1.625} & 2.4375 \\
\bottomrule
\end{tabular}
\end{table}

Table~\ref{numerical_exps:uci:card_rmse_and_crps_table} reports test performance in terms of RMSE and CRPS across datasets. All distributional variants perform at least on par with the baseline, with the exception of $\bm\Sigma_\theta^{\mathrm{mix}}$ and $\bm\epsilon_t^\mathrm{ES}$ on the \emph{yacht} dataset.
Still, even here, the simple diagonal Gaussian parameterization $\bm\Sigma_\theta^{\mathrm{diag}}$ consistently improves over $\delta_\theta$, indicating that modest extensions of the noise model can already yield measurable benefits.  


Improvements are particularly strong for CRPS, which directly evaluates distributional fit. Among the competitors, the Gaussian mixture parameterization $\bm\Sigma_\theta^{\mathrm{mix}}$ achieves the best overall performance, closely followed by the diagonal Gaussian $\bm\Sigma_\theta^{\mathrm{diag}}$. 
In fact, both methods rank first and second on average across datasets, clearly outperforming all other approaches.
This demonstrates that richer parameterized noise distributions can substantially improve probabilistic calibration without compromising predictive accuracy.

\subsection{Autoregressive prediction tasks} \label{sec:autoregressive-prediction-tasks}
Autoregressive prediction is a key benchmark for probabilistic modeling, as small errors accumulate over rollout length, making uncertainty quantification particularly important. We next evaluate our approach on this setting, where the goal is to generate temporal trajectories $u_1, \ldots, u_S$ from an initial state $u_0$\footnote{To avoid confusion with diffusion timesteps $t \in 1, \ldots, T$, rollout times are denoted by $s \in 1,\ldots,S$.}. At each step, the diffusion model $f_\theta$ samples from a conditional distribution $~{u_s \sim f_\theta (\cdot \mid u_{s-1}, u_{s-2})}$,
predicting the dynamics $u_s - u_{s-1}$ for three different systems:

\textbf{1D PDEs.}
We consider the Burgers' and Kuramoto--Sivashinsky (KS) equations,
\begin{align}
        \partial_s u(s,x) + \partial_x \big(u^2(s,x)/2\big) - \nu / \pi \, \partial_{xx}u(s,x)&= 0 \tag{Burgers'}\\
        \partial_s u(s,x)+ u \, \partial_x u(s,x) + \partial_x^2 u(s,x) + \partial_x^4 u(s,x) &= 0, \tag{KS}
\end{align}
with random initial conditions and a spatial resolution of $256$.

\textbf{Surface temperature prediction.}

As a more complex task, we predict 2-meter surface temperature (T2M) based on several meteorological input variables. We fix a 6-hour forecast horizon, initialized at 00UTC, and train on data from 2011 to 2020, evaluating on the period from 2021 to 2022.

For all tasks, we employ a U-Net diffusion backbone~\citep{songDenoisingDiffusionImplicit2022, karrasElucidatingDesignSpace2022}, adapted to arbitrary input shapes and 1D convolutions. PDE experiments are averaged over five runs, whereas T2M is evaluated once due to computational cost. Full experimental details are provided in Appendix~\ref{app:details_autoregressive}.

\begin{table}[ht] 
    \footnotesize
    \centering
        \caption{\emph{Autoregressive prediction results}\textemdash{}showing RMSE ($\downarrow$), CRPS ($\downarrow$) and $\mathcal{C}_{0.95}$ only. The RMSE and CRPS are scaled by the factor 1000 (100) for the Burgers' (KS) equation. Best values per dataset are in bold. A lower average rank indicates better performance.}
    \label{tab:autoregressive_results}
    \begin{tabular}{|l|ccc|ccc|ccc|c|}
    \toprule
    & \multicolumn{3}{c|}{Burgers'} 
    & \multicolumn{3}{c|}{KS} 
    & \multicolumn{3}{c|}{T2M} &\\
    \cmidrule(lr){2-4}\cmidrule(lr){5-7}\cmidrule(lr){8-10}
    Model & RMSE & CRPS & $\mathcal{C}_{0.95}$ & RMSE & CRPS & $\mathcal{C}_{0.95}$ & RMSE & CRPS & $\mathcal{C}_{0.95}$ & Avg. Rank \\
    \midrule
    $\bm\delta_\theta$ & 0.95 & 0.16 & \textbf{0.94} & 0.56 & 0.24 & 0.84 & 0.77 & 0.40 & 0.97 & 3.56 \\
    $\bm\Sigma_\theta^{\mathrm{diag}}$ & 0.81 & \textbf{0.12} & 1.00 & 0.39 & 0.21 & 1.00 & \textbf{0.71} & \textbf{0.35} & 0.84 & \textbf{2.39} \\
    $\bm\Sigma_\theta^{\mathrm{mix}}$  & 0.81 & 0.13 & 1.00 & \textbf{0.35} & \textbf{0.20} & 1.00 & \textbf{0.71}& \textbf{0.35 }& 0.83 & \textbf{2.39} \\
    $\bm\Sigma_\theta^{\mathrm{mv}}$   & \textbf{0.70} & 0.24 & 1.00 & 0.49 & 0.34 & 0.99 & 0.76 & 0.38 &\textbf{ 0.94} & 2.94 \\
    $\bm\epsilon_t^\mathrm{ES}$& 0.81 & 0.14 & 0.99 & 0.59 & 0.34 & \textbf{0.98} & -    & -    & -  & 3.08 \\
    \bottomrule
    \end{tabular}
\end{table}

\begin{figure}[ht]
\begin{subfigure}{\linewidth}
        \centering
    \includegraphics[width=\linewidth]{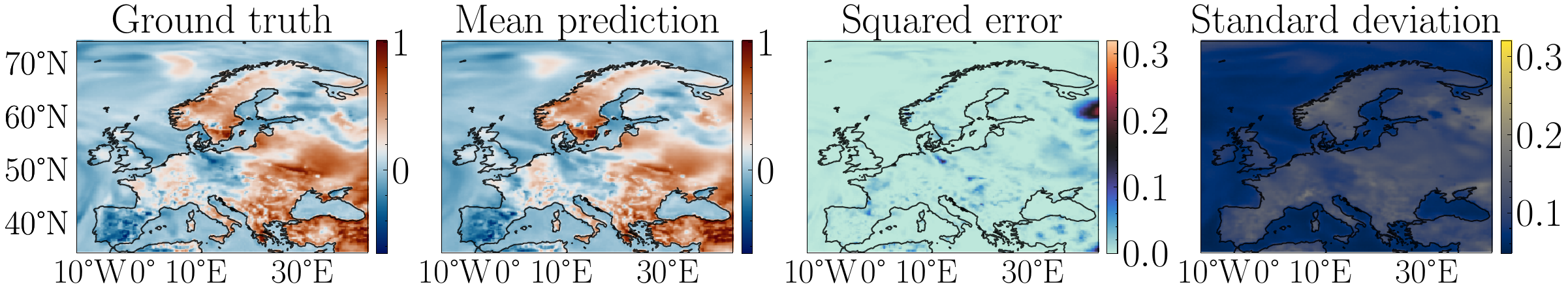}
\end{subfigure}
\begin{subfigure}{\linewidth}
        \centering
    \includegraphics[width=\linewidth]{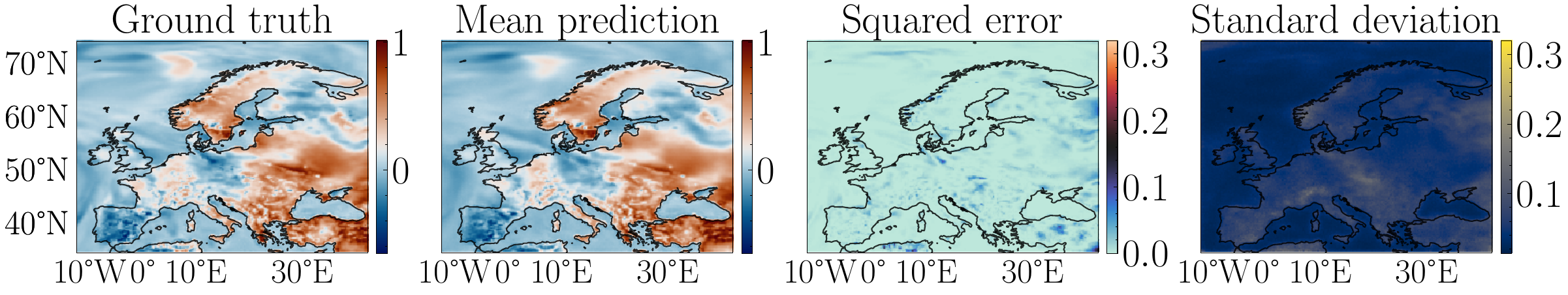}
\end{subfigure}
    \caption{Comparison of the $\bm \delta_\theta$ model (upper) and $\bm\Sigma_\theta^\mathrm{mix}$ model (lower) predictions for a selected sample of the dynamics of the 2-meter surface temperature.}
    \label{fig:t2m_visualization}
\end{figure}

Table~\ref{tab:autoregressive_results} summarizes the results. On Burgers', no single variant dominates all metrics, but every probabilistic model improves upon the deterministic baseline $\bm\delta_\theta$ for some metric, with particularly strong gains in the RMSE. For KS, the Gaussian mixture parameterization $\bm\Sigma_\theta^{\mathrm{mix}}$ achieves the best overall performance, clearly outperforming $\bm\delta_\theta$. Across both PDEs, coverage remains close to one, indicating slight underconfidence and imperfect marginal calibration; we analyze this behavior further in Appendix~\ref{app:calibration} and provide a principled way to further improve performance for our method.\\
For T2M, all models improve upon $\bm\delta_\theta$, with the strongest improvements by $\bm\Sigma_\theta^{\mathrm{diag}}$ and $\bm\Sigma_\theta^{\mathrm{mix}}$. Figure~\ref{fig:t2m_visualization} highlights this performance difference: while both models capture the mean structure, $\bm\delta_\theta$ leaves large residuals in the top-right region of the domain, which $\bm\Sigma_\theta^{\mathrm{mix}}$ successfully resolves.

\begin{figure}[t]
    \centering
    \includegraphics[width=\linewidth]{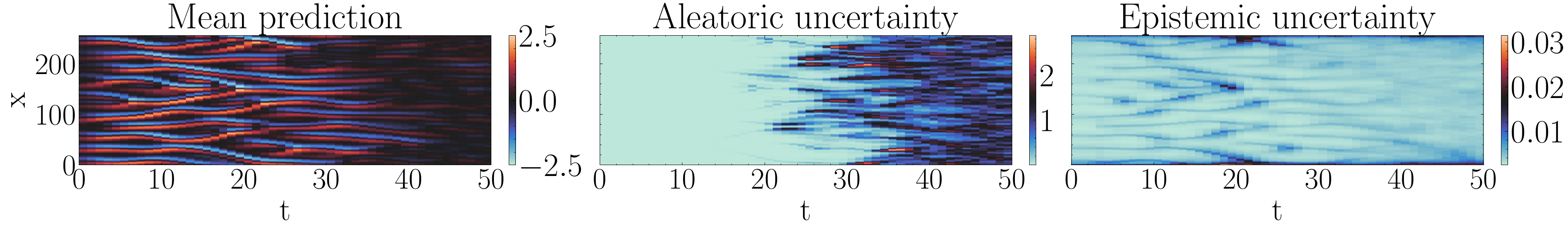}
    \caption{Comparison of aleatoric and epistemic uncertainty estimates for a sample trajectory of the $\bm\Sigma_\theta^\mathrm{diag}$ model.}
    \label{fig:ks_eu_main}
\end{figure}

One additional advantage of our framework is that, by learning $p_\theta^\epsilon(\epsilon_t \mid \bm x_t)$, it naturally provides a second-order distribution and hence estimates of epistemic (or model-) uncertainty \hbox{\citep{hullermeier2021aleatoric}}—something unavailable in standard diffusion models. Figure~\ref{fig:ks_eu_main} demonstrates this for the KS equation: While aleatoric uncertainty grows with rollout length, due to the chaotic nature of the governing equation, epistemic uncertainty remains structurally consistent and could potentially be removed with more available data. More details and a theoretical derivation are given in Appendix~\ref{app:epistemic_uncertainty}.

\subsection{Monocular depth estimation} \label{sec:depth-estimation}

Finally, we evaluate our method on monocular depth estimation, where the goal is to recover a dense depth map from a single RGB image. Recent work has shown that diffusion models achieve state-of-the-art performance on this task~\citep{ke2024repurposing}. In particular, Marigold~\citep{ke2024repurposing} adapts Stable Diffusion~\citep{rombach2022high} by finetuning its U-Net, concatenating image and depth embeddings, and modifying the first convolutional layer. We adopt this setup, additionally replacing the final convolution layer with our method-specific alternatives. \\
Following Marigold, we finetune on the synthetic datasets Hypersim~\citep{roberts2021hypersim} and VKitti2~\citep{cabon2020virtual}, and evaluate on five real-world benchmarks: NYUv2~\citep{silberman2012nyu}, ScanNet~\citep{dai2017scannet}, KITTI~\citep{geiger2012we}, ETH3D~\citep{schops2017multi}, and DIODE~\citep{vasiljevic2019diode}. Dataset splits and further details are provided in Appendix~\ref{app:experiment-details-depth}.

%
\renewcommand{\arraystretch}{0.85}
\begin{table}[ht] 
    \footnotesize
    \centering
    \caption{\emph{Depth estimation results} \textemdash{}showing AbsRel~($\downarrow$) and CRPS~($\downarrow$) only. Best values per dataset are in bold. Lower Avg. Rank indicates better overall performance.} 
    \label{tab:depth-estimation-absrel-crps}
    \begin{tabular}{|l|cc|cc|cc|cc|}
    \toprule
    & \multicolumn{2}{c|}{$\bm\delta_\theta$} 
    & \multicolumn{2}{c|}{$\bm\Sigma_\theta^{\mathrm{diag}}$} 
    & \multicolumn{2}{c|}{$\bm\Sigma_\theta^{\mathrm{mix}}$} 
    & \multicolumn{2}{c|}{$\bm\Sigma_\theta^{\mathrm{mv}}$} \\
    \cmidrule(lr){2-3}\cmidrule(lr){4-5}\cmidrule(lr){6-7}\cmidrule(lr){8-9}
    Experiment & AbsRel & CRPS & AbsRel & CRPS & AbsRel & CRPS & AbsRel & CRPS \\
    \midrule
    NYUv2   & 5.96 & 11.32 & 5.90 & 11.32 & 5.89 & 11.35 & \textbf{5.67} & \textbf{11.02} \\
    KITTI   & 10.32 & 142.92 & 10.07 & 138.24 & \textbf{9.89} & \textbf{137.60} & 10.14 & 142.28 \\
    ETH3D   & 6.82 & 29.23 & 6.57 & \textbf{27.89} & 6.72 & 28.66 & \textbf{6.47} & 29.43 \\
    ScanNet & 7.10 & 9.19 & 6.84 & 8.86 & 6.96 & 8.99 & \textbf{6.79} & \textbf{8.85} \\
    DIODE   & 30.60 & 191.39 & 31.52 & 196.08 & 31.09 & 190.04 & \textbf{29.82} & \textbf{186.63} \\
    \midrule
    Avg. Rank & 3.6 & 3.3 & 2.6 & 2.3 & 2.4 & 2.4 & \textbf{1.4} & \textbf{2.0} \\
    \bottomrule
    \end{tabular}
\end{table}

\begin{figure}
    \centering
    \begin{subfigure}{\linewidth}
        \includegraphics[width=\linewidth]{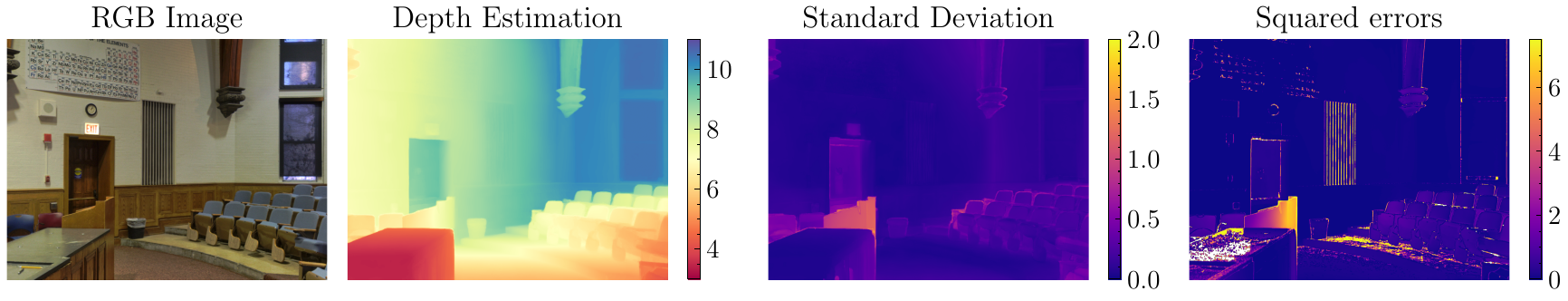}
    \end{subfigure}
    \begin{subfigure}{\linewidth}
        \includegraphics[width=\linewidth]{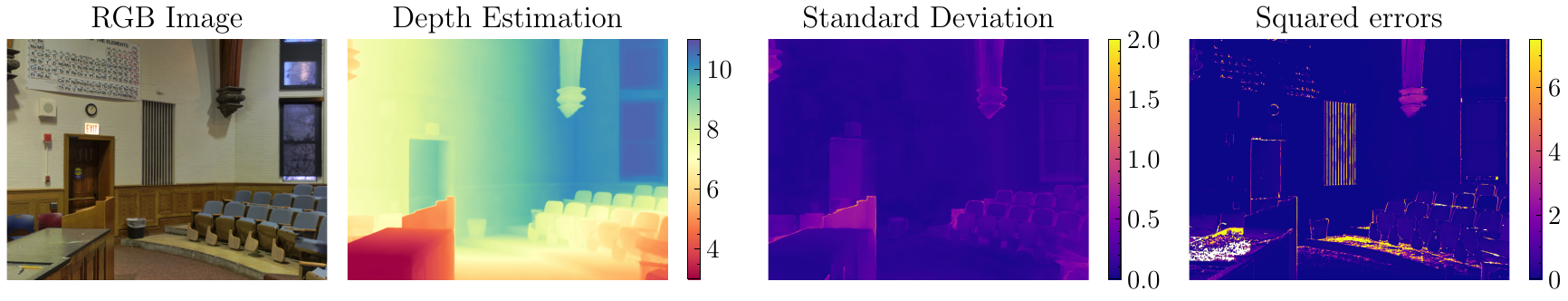}
    \end{subfigure}
    \caption{Comparison of the $\bm \delta_\theta$ model (upper) and $\bm\Sigma_\theta^\mathrm{mv}$ model (lower) predictions for the depth estimation task on DIODE \citep{vasiljevic2019diode}.}
    \label{fig:depth-estimation}
\end{figure}

Table~\ref{tab:depth-estimation-absrel-crps} shows that the multivariate Gaussian parameterization $\bm\Sigma_\theta^{\mathrm{mv}}$ achieves the best overall performance, consistently improving upon the deterministic Marigold baseline $\bm\delta_\theta$ across nearly all datasets and metrics. 
Figure~\ref{fig:depth-estimation} illustrates the improvement in prediction quality and uncertainty estimation over $\bm\delta_\theta$. Additional experiments, metrics, and further experiment details are reported in Appendix~\ref{app:additional_results}.

\section{Related work}

\textbf{Approximation of the noise distribution.}
This paper proposes to explicitly model the full predictive distribution of $\epsilon_t$ rather than only its conditional mean. Prior work has mainly focused on parameterizing the process variance as a diagonal Gaussian, using a variety of strategies~\citep{hoDenoisingDiffusionProbabilistic2020, nicholImprovedDenoisingDiffusion2021, bao2022analyticdpmanalyticestimateoptimal, bao2022estimatingoptimalcovarianceimperfect, ou2025improvingprobabilisticdiffusionmodels, zhang2024momentmatchingdenoisinggibbs}. These methods, however, remain tailored to generative tasks and limited to the Gaussian assumption, since they rely on the standard diffusion loss. Other lines of work challenge the Gaussian assumption in the reverse process and propose to replace it with alternative parametric distributions \hbox{\citep{nachmani2021nongaussiandenoisingdiffusion, guo2023gaussian, pandey2025heavytailed}}. Closest to our proposed approach is the recent work by \citet{bortoli2025distributional}, who propose to learn the noise distribution non-parametrically via proper scoring rules to reduce the number of denoising steps in generative modeling. In contrast, we focus on regression tasks, where accurate uncertainty quantification is essential, and propose a parameterized framework for specifying the noise distribution, allowing practitioners to balance expressivity and computational cost.

\textbf{Regression with diffusion.}
Diffusion-based models have also been adapted for regression tasks, ranging from classical UCI benchmarks~\citep{hanCARDClassificationRegression2022}, to autoregressive prediction~\citep{kohl_benchmarking_2024, priceProbabilisticWeatherForecasting2025, couairon2024archesweatherarchesweathergendeterministic, larsson2025diffusionlamprobabilisticlimitedarea}, and image-to-image regression such as depth estimation~\citep{ke2024repurposing}. In this context, several works have explored the connection between diffusion models and uncertainty quantification, for example via ensembling or Bayesian approaches \citep{shu2024zeroshotuncertaintyquantificationusing, chan2024estimating, berry2024shedding}. Our work complements these existing approaches by showcasing that our proposed framework can improve performance across a variety of regression tasks, independent of the underlying model architecture, and allows for assessing predictive uncertainty.

\section{Conclusion}

We proposed a framework for extending diffusion models and explicitly parameterizing the full distribution of $\epsilon_t$ within the diffusion process. This yields a general approach to probabilistic regression that improves predictive performance across a diverse set of benchmarks. We suggested concrete parameterizations based on multivariate Gaussian mixtures and showed that the corresponding backward sampling distributions admit a closed form. Training with proper scoring rules, our method consistently outperforms deterministic diffusion baselines\textemdash{}state-of-the-art on some datasets\textemdash{}while additionally providing estimates of epistemic uncertainty, a capability unavailable in standard diffusion models. 
Furthermore, we demonstrated that existing diffusion architectures can be finetuned within our framework with minimal modification. 

\textbf{Limitations \& future work}

Our approach introduces several design choices regarding the parameterization of the noise distribution. While experiments confirm consistent gains over standard diffusion models, the best-performing variant depends on the task. At present, there is no principled guidance for selecting a parameterization; the choice reflects a trade-off between distributional expressivity and computational cost. 
Apart from automating the selection process, future work could revolve around improving the different parameterizations and analyzing new ones, such as a multivariate Gaussian mixture. In addition, the combination with different diffusion frameworks and noise schedulers, such as those of \citet{karrasElucidatingDesignSpace2022} could be developed. Another very interesting avenue for future research would be to dive deeper into the capability of our approach to estimate epistemic uncertainty and to analyze whether the model can successfully detect out-of-distribution shifts.



\section*{Acknowledgements}
C.Kneissl, P. Scholl, and G. Kutyniok acknowledge support by the project "Genius Robot" (01IS24083), funded by the Federal Ministry of Education and Research (BMBF), as well as the ONE Munich Strategy Forum (LMU Munich, TU Munich, and the Bavarian Ministery for Science and Art).

C. Bülte and G. Kutyniok acknowledge support by the DAAD programme Konrad Zuse Schools of Excellence in Artificial Intelligence, sponsored by the Federal Ministry of Education and Research.

G. Kutyniok acknowledges partial support by the Munich Center for Machine Learning (BMBF), as well as the German Research Foundation under Grants DFG-SPP-2298, KU 1446/31-1 and KU 1446/32-1. G. Kutyniok also acknowledges support by the gAIn project, which is funded by the Bavarian Ministry of Science and the Arts (StMWK Bayern) and the Saxon Ministry for Science, Culture and Tourism (SMWK Sachsen). Furthermore, G. Kutyniok is supported by LMUexcellent, funded by the Federal Ministry of Education and Research (BMBF) and the Free State of Bavaria under the Excellence Strategy of the Federal Government and the Länder as well as by the Hightech Agenda Bavaria.

\bibliography{references}
\bibliographystyle{plainnat}

\clearpage
\appendix

\section{Approximations of the noise distribution}
\label{app:approximation}

\subsection{Approximations of the covariance matrix}
\label{app:covariance_approximations}
As already mentioned, apart from only modeling the diagonal, we consider two efficient ways of parameterizing the full covariance matrix $\Sigma_{\theta,k} \in \Ree^{D \times D}$.
\paragraph{Cholesky decomposition}
First, we represent $\Sigma_{\theta,k}$ as
\[
\Sigma_{\theta,k} = L_{\theta,k} L_{\theta,k}^\top,
\]
where $L_{\theta,k}$ is a lower triangular matrix with positive diagonal. This reduces the number of predicted parameters from $D^2$ to $D(D-1)/2$ while ensuring symmetry. To guarantee positive definiteness, we apply a softplus transformation to the diagonal of $ L_{\theta,k}$, normalize the strictly lower part columnwise, and add a small stability constant $\varepsilon = 10e-6$. In practice, we found this to be sufficient to ensure a positive definite covariance matrix. The Cholesky decomposition offers several advantages \citep{muschinskiCholeskybasedMultivariateGaussian2024}:
\begin{itemize}
    \item Sampling is efficient and requires only a matrix-vector product.
    \item The determinant can be computed as $\det(\Sigma) = (\prod_{i=1}^D L_{ii})^2$.
    \item Although computing the full inverse $\Sigma^{-1}$ has cost $\mathcal{O}(D^3)$, inverse-vector products $\Sigma^{-1} v$ require only $\mathcal{O}(D^2)$.
\end{itemize}
Nevertheless, for very high-dimensional data, these computations may still be too expensive.

\paragraph{Low-rank + diagonal}
As a more efficient alternative, we consider the low-rank + diagonal structure \citep{rezendeStochasticBackpropagationApproximate2014}
\[
\Sigma_{\theta,k} = U_{\theta,k} U_{\theta,k}^\top + D_{\theta,k},
\]
where $U_{\theta,k} \in \Ree^{D\times r}$, with $\mathrm{rank}(U_{\theta,k}) = r \ll D$ and $D_{\theta,k} \in \Ree^{D \times D}$ is a (positive) diagonal matrix. Here, $U_{\theta,k}$ and the diagonal $D_{\theta,k}$ are both predicted by the neural network.
Using the matrix determinant lemma, we obtain $\det(\Sigma) =  \det(D_{\theta,k}) \det (\mathbf{I}_r + U_{\theta,k}^\top D_{\theta,k}^{-1} U_{\theta,k})$, which can be computed in $\mathcal{O}(Dr^2 + r^3)$. Similarly, inverse-vector products $\Sigma^{-1}v$ cost $\mathcal{O}(Dr^2 + r^3 + Dr)$, while sampling is also efficient. When $r \ll D$, this parameterization is nearly as efficient as the diagonal case, though less expressive than the full Cholesky form.

\begin{figure}[ht]
    \centering
    \includegraphics[width=\linewidth]{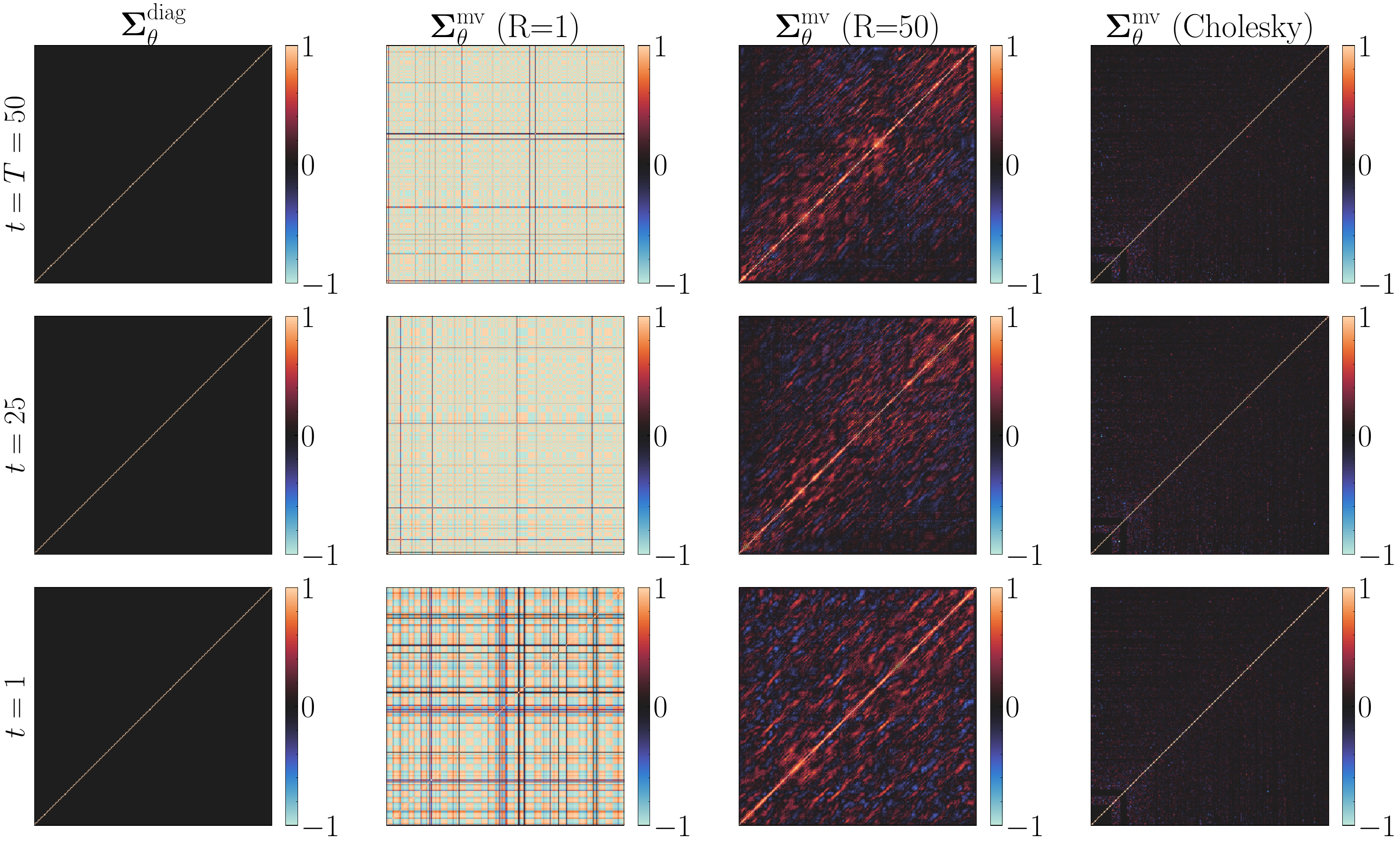}
    \caption{Correlation matrices for different covariance parameterizations across several time steps $t$ for a selected sample of the Kuramoto--Sivashinsky equation.}
    \label{fig:correlation_comparison}
\end{figure}

Table~\ref{tab:mv_approximations} summarizes the computational complexity of common operations. The low-rank plus diagonal scales favorably when $r \ll D$, while Cholesky remains more costly. Figure~\ref{fig:correlation_comparison} illustrates the different correlation structures for each method during different diffusion timesteps for a selected sample of the Kuramoto--Sivashinsky equation.

\begin{table}[ht]
\centering
\caption{Computational complexity for common operations on covariance matrices $\Sigma$ under different factorizations. 
$D$ is the data dimension, $r$ is the rank of $U$ in the low-rank$+$diagonal case.}
\label{tab:mv_approximations}
\begin{tabular}{lcccc}
\hline
\textbf{Method} 
& \textbf{Sampling from $\mathcal{N}(0,\Sigma)$} 
& \textbf{$\det(\Sigma)$} 
& \textbf{$\Sigma^{-1}$} 
& \textbf{$\Sigma^{-1} v$} \\
\hline
Cholesky & $\mathcal{O}(D^2)$ & $\mathcal{O}(D)$ & $\mathcal{O}(D^3)$ & $\mathcal{O}(D^2)$ \\
Low-rank $+$ diag & $\mathcal{O}(Dr)$ & $\mathcal{O}(Dr^2 + r^3)$ & $\mathcal{O}(D^2 r)$ & $\mathcal{O}(Dr^2 + r^3)$ \\
Diagonal & $\mathcal{O}(D)$ & $\mathcal{O}(D)$ & $\mathcal{O}(D)$ & $\mathcal{O}(D)$ \\
\hline
\end{tabular}
\end{table}

\subsection{Closed-form expressions for scoring rules}
\label{app:closed_form}
In this section, we provide closed-form expressions for different scoring rules for the different instantiations of our proposed predictive distribution.
\paragraph{Univariate Gaussians}
For univariate Gaussians, the energy score reduces to the continuous ranked probability score (CRPS), which admits a closed-form expression \citep{Gneiting.2007}:
\begin{equation}
    \mathrm{CRPS}(\mathcal{N}(\mu, \sigma^2), y) = \sigma \left( \frac{y-\mu}{\sigma} \left( \Phi \left(\frac{y-\mu}{\sigma} \right) -1\right) + 2\varphi\left(\frac{y-\mu}{\sigma} \right) - \frac{1}{\sqrt{\pi}}  \right),
\end{equation}
where $\Phi$ and $\varphi$ denote the CDF and PDF of a standard normal distribution, respectively. For a Gaussian kernel with bandwidth $\gamma$, the corresponding kernel score admits a closed-form solution as
\begin{equation}
    S_k(\mathcal{N}(\mu. \sigma^2), y) =
- \frac{1}{\sqrt{1 + 2\sigma^2/\gamma^2}}
\exp\left(
- \frac{(y - \mu)^2}{\gamma^2 + 2\sigma^2}
\right)
+ \frac{1}{2 \sqrt{1 + 4\sigma^2/\gamma^2}}
+ \frac{1}{2}.
\end{equation}
\begin{proof}
The proof follows a similar direction to \citet{rustamov2020closedformexpressionsmaximummean}. By definition, the Gaussian kernel score is given as $$S_k(\mathcal{N}(\mu, \sigma^2),y) =  \frac{1}{2} \bE_{X,X' \sim\mathcal{N}(\mu, \sigma^2)} [k(X,X')] -\bE_{X\sim \mathcal{N}(\mu, \sigma^2)}[k(X,y)] + \frac{1}{2}k(y,y),$$ with $k = \exp(- \|x - x'\|^2 / 2\gamma^2)$.

We start with the second term, which we can write as
\begin{align*}
    \bE_{\mathcal{N}(\mu, \sigma^2)}[k(X,y)] &= \int_{\Ree} e^{- (x-y)^2 / \gamma^2} (2\pi \sigma^2)^{-1/2} e^{-(x-\mu)^2/(2\sigma^2)} \, dx \\
    &= (\pi \gamma^2)^{1/2} \int_{\Ree} (\pi \gamma^2)^{-1/2} e^{- (x-y)^2 / \gamma^2} (2\pi \sigma^2)^{-1/2} e^{(x-\mu)^2/(2\sigma^2)} \, dx.
\end{align*}
Since $(x-y)^2 = (y-x)^2$, this can be expressed as the convolution of two densities
\begin{align*}
    \bE_{\mathcal{N}(\mu, \sigma^2)}[k(X,y)] &= (\pi \gamma^2)^{1/2} (f_{\mu, \sigma^2} \ast f_{0, \gamma^2/2})(y),
\end{align*}
where $f_{\mu, \sigma^2}$ denotes the pdf of a Gaussian. By the convolution theorem of Gaussian random variables, we know that $(f_{\mu, \sigma^2} \ast f_{0, \gamma^2/2})(y) = f_{\mu, \sigma^2 + \gamma^2/2}(y)$. Therefore, we obtain
\begin{align*}
 \bE_{\mathcal{N}(\mu, \sigma^2)}[k(X,y)] &= (\pi \gamma^2)^{1/2} (\pi(\gamma^2+2\sigma^2))^{-1/2} e^{-(y-\mu)^2/(2\sigma^2 + \gamma^2)}\\
 &= \frac{1}{\sqrt{1 + 2\sigma^2/\gamma^2}} \exp \left(- \frac{(y-\mu)^2}{\gamma^2 + 2\sigma^2} \right).
\end{align*}

The first term can be obtained by recognizing $X-X' \sim \mathcal{N}(0, 2\sigma^2)$ and following the same steps as above.
\end{proof}

\paragraph{Univariate Gaussian mixture}
Let $F_K(x) = \sum_{i=1}^K w_i \mathcal{N}(x; \mu_i, \sigma_i^2)$ denote a $K$-component normal mixture distribution.  Then, a closed-form expression for the CRPS is given as
\begin{equation}
    \mathrm{CRPS}(F_K, y) = \sum_{i=1}^K w_i A(y-\mu_i, \sigma_i^2) - \frac{1}{2} \sum_{i=1}^K \sum_{j=1}^K w_j w_i A(\mu_i - \mu_j, \sigma_i^2 + \sigma_j^2),
\end{equation}
where $A(\mu, \sigma^2) = \mu \left(2\Phi\left( \frac{\mu}{\sigma} \right)-1 \right) + 2\sigma \varphi \left( \frac{\mu}{\sigma^2} \right)$ \citep{https://doi.org/10.1256/qj.05.235}.

\paragraph{Multivariate Gaussian}
In the case of the multivariate Gaussian $\mathcal{N}(\bm \mu, \bm\Sigma)$, the energy score does not admit an analytical expression \citep{szekely_new_2005}. However, the kernel score with a Gaussian kernel does admit a closed-form, given as
\begin{equation}
\begin{split}
S_k(\mathcal{N}(\bm \mu, \bm\Sigma), \bm{y}) &=
- \frac{1}{\sqrt{\det\left( I + \tfrac{2}{\gamma^2} \bm\Sigma \right)}}
 \exp\left( -\frac{1}{\gamma^2} (\bm{y} - \bm{\mu})^\top \left( I + \tfrac{2}{\gamma^2} \bm\Sigma \right)^{-1} (\bm{y} - \bm{\mu}) \right) \\
&+ \frac{1}{2\sqrt{\det\left( I + \tfrac{4}{\gamma^2} \bm\Sigma \right)}}
+ \frac{1}{2},
\end{split}
\end{equation}
where $\gamma$ is the bandwidth of the kernel.
\begin{proof}
Let $X,X' \sim \mathcal{N}(\bm \mu, \bm\Sigma)$ and $Z \coloneq X - \bm y \sim \mathbb{P}_Z \coloneq \mathcal{N}(\bm \mu - \bm y, \bm\Sigma)$. Then $\|Z \|^2 = Z^\top Z = Z^\top A Z$ with $A = \boldsymbol{I}_d$ follows a generalized chi-squared distribution \citep{das2024methodsintegratemultinormalscompute}, i.e., $Z^\top A Z \sim \Tilde{\chi}(\boldsymbol{I}_d, \boldsymbol{0},0, \boldsymbol{\mu} - \boldsymbol{y}, \boldsymbol{\bm\Sigma})$. Then, we can express the kernel score in terms of the corresponding moment-generating function, via
    \begin{align*}
        \bE_{X \sim \mathcal{N}(\bm \mu, \bm\Sigma)}[k(X,\bm y)] & = \bE_{\bP_Z}\left[\exp\left(- \frac{Z^\top Z}{\gamma^2} \right)\right] =  M_{\Tilde{\chi}(\boldsymbol{I}_d, \boldsymbol{0},0, \boldsymbol{\mu} - \bm y, \bm\Sigma)}\left(-\frac{1}{\gamma^2}\right),
    \end{align*}
    with $t = - \frac{1}{\gamma^2}$. In the case of the generalized chi-squared distribution, the moment-generating function is given as
    \begin{align*}
        M_{\Tilde{\chi}(\boldsymbol{I}_d, \boldsymbol{0},0, \boldsymbol{\mu}-\bm y, \boldsymbol{\Sigma})}\left(t\right) = \frac{1}{\sqrt{\det(\boldsymbol{I}_d - 2 t\boldsymbol{\Sigma})}} \exp \left( t (\boldsymbol{\mu}-\bm y)^\top (\boldsymbol{I}_d - 2t \boldsymbol{\Sigma})^{-1} (\boldsymbol{\mu} - \bm y) \right),
    \end{align*}
    from which we obtain
    \begin{align*}
          \bE_{X \sim \mathcal{N}(\bm \mu, \bm\Sigma)}[k(X,\bm y)] & =  \frac{1}{\sqrt{\det \left( \boldsymbol{I}_d + \frac{2}{\gamma^2} \boldsymbol{\Sigma} \right)}} \exp \left( -\frac{1}{\gamma^2} (\boldsymbol{\mu}-\bm y)^\top \left( \boldsymbol{I}_d + \frac{2}{\gamma^2} \boldsymbol{\Sigma}\right)^{-1}(\boldsymbol{\mu}-\bm y) \right).
    \end{align*}
    Repeating these steps for $Z \coloneq X - X' \sim \mathbb{P}_Z \coloneq \mathcal{N}(0, 2\bm\Sigma)$ gives the expression for $\bE_{X,X' \sim \mathcal{N}(\bm \mu, \bm\Sigma)}[k(X, X')]$ and therefore the full expression for $S_k(\mathcal{N}(\bm \mu, \bm\Sigma), \bm{y})$.
\end{proof}

\FloatBarrier
\subsection{Proof of Theorem \ref{methodology:thm:closed_form}}\label{app:proof_of_thm1}
We start by stating a theorem concerning the distribution of a Gaussian random variable whose mean is the linear combination of another Gaussian distribution, which will serve as an important tool in many of the upcoming proofs.
\begin{theorem} \label{appendix:thm:gaussian_thm}
    Let $p(x) = \mathcal{N}(\mu_x, \Sigma_x)$, with mean $\mu_x\in \mathbb{R}^d$ and positive semi-definite covariance matrix $\Sigma_x \in \mathbb{R}^{d \times d}$.
    Further, let $y$ be a random vector with conditional distribution $p(y \mid x) = \mathcal{N}(Ax + b, \Sigma_y)$, for $A \in \mathbb{R}^{d \times d}, b \in \mathbb{R}^d$.
    Then $p(y) = \mathcal{N}(A\mu_x + b, A \Sigma_x A^T + \Sigma_y)$.
\end{theorem}

For a proof, compare Theorem 4.4.1 in \citet{Murphy2014}, "Machine Learning: A Probabilistic Perspective".
We now give the proof of Theorem~\ref{methodology:thm:closed_form}.
\begin{proof}
    First, we obtain an equivalent expression for \eqref{background:diffusion:reverse_process_def} which is now conditioned on $\epsilon_t$ instead of $x_0$. 
    Since $\epsilon_t = \frac{1}{\sqrt{1 - \bar{\alpha}_t}} 
    \left( 
    X_t - \sqrt{\bar{\alpha}}X_0
    \right)$, we can write 
    \begin{equation*}
        \begin{split}
            \mu(x_0, x_t) &= 
            \sqrt{\bar{\alpha}_{t-1}}x_0 + \sqrt{1 - \bar{\alpha}_{t-1} - \sigma_t^2}
            \cdot \frac{x_t - \sqrt{\bar{\alpha}_t}x_0}{\sqrt{1 - \bar{\alpha}_t}} \\
            &= \sqrt{\bar{\alpha}_{t-1}} 
            \cdot \frac{x_t - \sqrt{1 - \bar{\alpha}_t} \epsilon_t}{\sqrt{\bar{\alpha}_t}}
            + \sqrt{1 - \bar{\alpha}_{t-1} - \sigma_t^2} \epsilon_t \\
            &\eqcolon \tilde{\mu}(\epsilon_t, x_t) \eqcolon a(x_t, t)\epsilon_t + b(x_t, t)
        \end{split}
    \end{equation*}
    where we introduce the functions $a(x_t, t)$ and $b(x_t, t)$ for notational simplicity. 
    We obtain $p(x_{t-1} \mid x_t, \epsilon_t) = \cN (\tilde{\mu}(\epsilon_t, x_t), \sigma_t^2 \mathbf{I})$.
    Now,
    \begin{equation*}
        \begin{split}
            p_\theta^{(t)}(x_{t-1}\mid x_t)
            &= \int q(x_{t-1} \mid x_t, \epsilon_t) p_\theta(\epsilon_t \mid x_t) d \epsilon_t. \\
            &= \int \cN(x_{t-1}; a(x_t, t)\epsilon_t + b(x_t, t), \sigma_t^2 \mathbf{I} )
              \left(
              \sum_{k=1}^K \pi_\theta^k \cN (\epsilon_t; \mu^{\epsilon}_{\theta, k}(x_t), \Sigma^{\epsilon}_{\theta, k}(x_t))
              \right) d\epsilon_t \\
            &= \sum_{k=1}^K \pi_\theta^k \int \cN(x_{t-1}; a(x_t, t)\epsilon_t + b(x_t, t), \sigma_t^2 \mathbf{I} ) 
            \cN (\epsilon_t; \mu^{\epsilon}_{\theta, k}(x_t), \Sigma^{\epsilon}_{\theta, k}(x_t)) d\epsilon_t \\
            &= \sum_{k=1}^K \pi_\theta^k 
            \cN(x_{t-1}; 
            a(x_t, t)\mu^{\epsilon}_{\theta, k}(x_t) + b(x_t, t), 
            a(x_t, t)^2 \Sigma^{\epsilon}_{\theta, k}(x_t) + \sigma_t^2 \mathbf{I} 
            ) \\
            &= 
            \sum_{k=1} ^K\pi_\theta^k \mathcal{N}
            \left(x_{t-1};
             \sqrt{\bar{\alpha}_{t-1}} \hat{x}_0 + 
            \sqrt{ 1 - \bar{\alpha}_{t-1} - \sigma_t^2}\mu^{\epsilon}_{\theta, k}(x_t) ,
            \gamma_t^2 \Sigma^{\epsilon}_{\theta, k}(x_t) + \sigma_t^2 \mathbf{I}
            \right),
        \end{split}
    \end{equation*}
    where the fourth equality follows by Theorem~\ref{appendix:thm:gaussian_thm} and the last by plugging in the constants.
\end{proof}

\section{Details on CARD} \label{appendix:proofs}
CARD was introduced by \citet{hanCARDClassificationRegression2022} as a framework for classification and regression tasks using diffusion.
They assume to have a pre-trained regressor $f_\phi : \mathcal{C} \to \mathcal{Y}$ available, which is trained on $\mathcal{D}$ and approximates the conditional expectation $\bE [\bm{y} \mid \bm{c}]$ given covariates $\bm c \in \mathcal{C}$.
Its core idea is, that a diffusion process is used to interpolate between the noise distribution $p(\bm{y_t} \mid \bm{c}) = \cN(f_\phi(\bm{c}), \mathbf{I})$ and the data distribution $p(\bm{y_0} \mid \bm{c})$.
We will provide a formulation of CARD embedded in the DDIM framework introduced earlier.
To this end, we set the mean in \eqref{background:diffusion:reverse_process_def} to
\begin{equation}
    \label{numerical_exps:uci:card_mean_formulation}
    \mu(\bm y_0, \bm y_t) = \sqrt{\bar{\alpha}_{t-1}}\bm y_0 + 
    \left( 1 - \sqrt{\bar{\alpha}_{t-1}} \right)f_\phi(\bm c) + 
    \sqrt{ 1 - \bar{\alpha}_{t-1} - \sigma_t^2} \cdot 
    \frac{\bm y_t - \left(\sqrt{\bar{\alpha}_t}\bm y_0 +  
    \left(1 - \sqrt{\bar{\alpha}_t} \right)f_\phi(\bm c) \right)}
    {\sqrt{1 - \bar{\alpha}_t}}
\end{equation}
which yields a modified version of \eqref{background:diffusion:nice_property}, namely
\begin{equation}
\label{numerical_exps:uci:card_nice_property}
    p(\bm y_t \mid \bm y_0, f_\phi(\bm c)) 
    = \cN \left(\sqrt{\bar{\alpha}_t}\bm y_0 + \left(1 - \sqrt{\bar{\alpha}_t}f_\phi(\bm c)) \right),
    \left(1 - \bar{\alpha}_t \right) \mathbf{I} \right).
\end{equation}

We will prove this and the fact that we can recover the original CARD method by an appropriate choice of $\{\sigma_t\}_{t=1}^T$. \\
The original CARD algorithm for sampling $y_{t-1}$ given $y_t$ is specified as follows
\begin{equation}
    \begin{cases}
      \hat{y}_0 = \frac{1}{\sqrt{\bar{\alpha}_t}}
      \left(y_t - \left(1 - \sqrt{\bar{\alpha}_t}\right)f_\theta(x) 
      - \sqrt{1 - \bar{\alpha}_t}\epsilon_\theta \right)  \\
      y_{t-1} = \gamma_0 \hat{y}_0 + \gamma_1 y_t + \gamma_2 f_\phi(x) + \sqrt{\tilde{\beta}}z, \quad z \sim \cN(0, \mathbf{I})
    \end{cases}\,
\end{equation}
where
\begin{equation}
    \gamma_0 \coloneq \frac{\beta_t \sqrt{\bar{\alpha}_{t-1}}}{1 - \bar{\alpha}_t}, \quad
    \gamma_1 \coloneq \frac{\left(1 - \bar{\alpha}_{t-1} \right) \sqrt{\alpha_t}}{1 - \bar{\alpha}_t}, \quad
    \gamma_2 \coloneq \left(1 + 
    \frac{\left( \sqrt{\bar{\alpha}_t} - 1 \right)
    \left( \sqrt{\alpha_t} + \sqrt{\bar{\alpha}_{t-1}} \right)}{1 - \bar{\alpha}_t} \right).
\end{equation}
We have 
\begin{align}
    y_{t-1} &= \gamma_0 \frac{1}{\sqrt{\bar{\alpha}_t}}
      \left(y_t - \left(1 - \sqrt{\bar{\alpha}_t}\right)f_\theta(x) 
      - \sqrt{1 - \bar{\alpha}_t}\epsilon_\theta \right) + \gamma_1 y_t + \gamma_2 f_\phi(x) + \sqrt{\tilde{\beta}}z \\
      &= \underbrace{\left( \frac{\gamma_0}{\sqrt{\bar{\alpha}_t}} + \gamma_1 \right)}_{=:A}y_t + 
      \underbrace{\left( -\frac{\gamma_0\left(1 - \sqrt{\bar{\alpha}_t}\right)}{\sqrt{\bar{\alpha}_t}} + \gamma_2 \right)}_{=:B}f_\phi(x) - 
      \underbrace{\frac{\gamma_0 \sqrt{1 - \bar{\alpha}_t}}{\sqrt{\bar{\alpha}_t}}}_{=:C}\epsilon_\theta
        + \sqrt{\tilde{\beta}}z
\end{align}
and calculate
\begin{align}
    A &= \frac{\beta_t \sqrt{\bar{\alpha}_{t-1}}}{\left(1 - \bar{\alpha}_t\right)\sqrt{\bar{\alpha}_t}} + \frac{\left(1 - \bar{\alpha}_{t-1} \right) \sqrt{\alpha_t}}{1 - \bar{\alpha}_t}
      = \frac{\beta_t}{\left(1 - \bar{\alpha}_t\right)\sqrt{\alpha_t}} + \frac{\left(1 - \bar{\alpha}_{t-1} \right)\alpha_t}{\left(1 - \bar{\alpha}_t \right)\sqrt{\alpha_t}}\\
      &= \frac{1 - \alpha_t + \alpha_t - \bar{\alpha}_t}{\left(1 - \bar{\alpha}_t \right)\sqrt{\alpha_t}} 
      = \frac{1}{\sqrt{\alpha_t}}
\end{align}
as well as
\begin{align}
    B &= - \frac{\beta_t \sqrt{\bar{\alpha}_{t-1}}\left(1 - \sqrt{\bar{\alpha}_t}\right)}{\sqrt{\bar{\alpha}_t}\left(1 - \bar{\alpha}_t\right)} + 1 + 
    \frac{\left( \sqrt{\bar{\alpha}_t} - 1 \right)
    \left( \sqrt{\alpha_t} + \sqrt{\bar{\alpha}_{t-1}} \right)}{1 - \bar{\alpha}_t} \\
      &= \frac{-(1 - \alpha_t)\left(1 - \sqrt{\bar{\alpha}_t}\right) +  
      \left( \sqrt{\bar{\alpha}_t} - 1 \right)
    \left( \sqrt{\alpha_t} + \sqrt{\bar{\alpha}_{t-1}} \right) \sqrt{\alpha_t}}
      {\sqrt{\alpha_t}\left(1 - \bar{\alpha}_t\right)} +1 \\
      &= \frac{-1 + \alpha_t + \sqrt{\bar{\alpha}_t} - \alpha_t \sqrt{\bar{\alpha}_t}
      + \left( \sqrt{\bar{\alpha}_t} - 1 \right) \left( \alpha_t + \sqrt{\bar{\alpha}_t} \right)} 
      {\sqrt{\alpha_t}\left(1 - \bar{\alpha}_t\right)} +1 \\
      &= \frac{-1 + \alpha_t + \sqrt{\bar{\alpha}_t} - \alpha_t \sqrt{\bar{\alpha}_t}
      + \sqrt{\bar{\alpha}_t} \alpha_t + \bar{\alpha}_t - \alpha_t - \sqrt{\bar{\alpha}_t} }
      {\sqrt{\alpha_t}\left(1 - \bar{\alpha}_t\right)} + 1 \\
      &= 
      \frac{\bar{\alpha}_t - 1 }{\sqrt{\alpha_t}\left(1 - \bar{\alpha}_t\right)} + 1 \\
      &=  - \frac{1}{\sqrt{\alpha_t}} \left(1 - \sqrt{\alpha_t} \right)
\end{align}
and lastly
\begin{align}
    C &= \frac{\beta_t \sqrt{\bar{\alpha}_{t-1}} \sqrt{1 - \bar{\alpha}_t}}
    {\left( 1 - \bar{\alpha}_t \right)\sqrt{\bar{\alpha}_t}}
    = \frac{1}{\sqrt{\alpha_t}} \cdot \frac{1 - \alpha_t}{\sqrt{1 - \bar{\alpha}_t} }.
\end{align}
In total, we obtain
\begin{equation}
    y_{t-1} = \frac{1}{\sqrt{\alpha}_t} 
    \left(
        y_t - \frac{1 - \alpha_t}{\sqrt{1 - \bar{\alpha}_t} } \epsilon_\theta
        - \left( 1 - \sqrt{\alpha_t}\right) f_\phi(x) 
    \right)
\end{equation}
To finish the argument we substitute $\sigma = \sqrt{\frac{1 - \bar{\alpha}_{t-1}}{1 - \bar{\alpha}_t} \beta_t}$ in \eqref{numerical_exps:uci:card_mean_formulation} and obtain
\begin{align}
    & \sqrt{\bar{\alpha}_{t-1}}
    \frac{y_t - \left(\sqrt{1 - \bar{\alpha}_t} \epsilon_\theta +  
    \left(1 - \sqrt{{\bar{\alpha}}_t} \right)f_\phi(x) \right)}
    {\sqrt{\bar{\alpha}}_t}  + 
    \sqrt{ 1 - \bar{\alpha}_{t-1} - \frac{1 - \bar{\alpha}_{t-1}}{1 - \bar{\alpha}_t} (1 -\alpha_t)} \cdot \epsilon_\theta + \\
    &\quad
    \left(1 - \sqrt{\bar{\alpha}_{t-1}} \right)f_\phi(x) \\
    &= \frac{1}{\sqrt{\alpha}_t} 
    \left(
        y_t - \sqrt{1 - \bar{\alpha}_t}  \epsilon_\theta
        - \left( 1 -\sqrt{{\bar{\alpha}}_t}\right) f_\phi(x) 
    \right) + \\
    &\quad
    \sqrt{\frac{1 - \bar{\alpha}_{t-1} - \bar{\alpha}_t + \bar{\alpha}_{t-1}\bar{\alpha}_t
    - 1 + \alpha_t + \bar{\alpha}_{t-1} - \bar{\alpha}_t }
    {1 - \bar{\alpha}_t}} \cdot \epsilon_\theta + 
    \left(1 - \sqrt{\bar{\alpha}_{t-1}} \right)f_\phi(x) \\
    &= \frac{1}{\sqrt{\alpha}_t} 
    \left(
        y_t - \sqrt{1 - \bar{\alpha}_t}  \epsilon_\theta
        - \left( 1 - \sqrt{{\bar{\alpha}}_t}\right) f_\phi(x) 
    \right) + 
    \sqrt{\frac{ - \bar{\alpha}_t + \bar{\alpha}_{t-1}\bar{\alpha}_t
     + \alpha_t  - \bar{\alpha}_t }
    {1 - \bar{\alpha}_t}} \cdot \epsilon_\theta +\\
    &\quad
   \left(1 - \sqrt{\bar{\alpha}_{t-1}} \right)f_\phi(x)\\
    &= \frac{1}{\sqrt{\alpha}_t} 
    \left(
        y_t - \sqrt{1 - \bar{\alpha}_t}  \epsilon_\theta
        - \left( 1 - \sqrt{{\bar{\alpha}}_t}\right) f_\phi(x) 
    \right) + 
    \sqrt{\frac{ \left( \alpha_t - \bar{\alpha}_t \right) \left(1 - \bar{\alpha}_{t-1} \right) }
    {1 - \bar{\alpha}_t}} \cdot \epsilon_\theta + \\
    &\quad
    \left(1 - \sqrt{\bar{\alpha}_{t-1}} \right)f_\phi(x)\\
    &= \frac{1}{\sqrt{\alpha}_t} 
    \left(
        y_t - \sqrt{1 - \bar{\alpha}_t}  \epsilon_\theta
        - \left( 1 - \sqrt{{\bar{\alpha}}_t}\right) f_\phi(x) 
    \right) + 
    \sqrt{\frac{ \left( \alpha_t - \bar{\alpha}_t \right)^2 }
    {\alpha_t \left( 1 - \bar{\alpha}_t \right)}} \cdot \epsilon_\theta + 
    \left(1 - \sqrt{\bar{\alpha}_{t-1}} \right)f_\phi(x)\\
    &= \frac{1}{\sqrt{\alpha}_t} 
    \left(
        y_t + \left(\frac{ \left( \alpha_t - \bar{\alpha}_t \right)}
    {\sqrt{ 1 - \bar{\alpha}_t }} -
    \sqrt{1 - \bar{\alpha}_t} \right) \epsilon_\theta
        - \left( 1 - \sqrt{{\bar{\alpha}}_t} +
        \sqrt{\alpha_t} - \sqrt{\alpha_t \bar{\alpha}_{t-1}}  \right) f_\phi(x) 
    \right) \\
    &= \frac{1}{\sqrt{\alpha}_t} 
    \left(
        y_t - \frac{1 - \alpha_t}{\sqrt{1 - \bar{\alpha}_t}}
        \epsilon_\theta - \left( 1 - \sqrt{{\bar{\alpha}}_t}\right) f_\phi(x) 
    \right) 
\end{align}
\\
We also want to show that \eqref{numerical_exps:uci:card_nice_property} holds, which we will do by an induction argument, similar in style to \citet{songDenoisingDiffusionImplicit2022}.
\begin{proof}
    Assume that 
    \begin{equation}
    \label{appendix:card_proofs:induction_hyp}
        p(y_t \mid y_0, f_\phi(x)) 
        = \cN \left(\sqrt{\bar{\alpha}_t}y_0 + \left(1 - \sqrt{\bar{\alpha}_t}f_\phi(x) \right),
        \left(1 - \bar{\alpha}_t \right) \mathbf{I} \right).
    \end{equation}
    holds for $t = 1, \ldots, T$. We will show that it is also satisfied for $t-1$.
    Since it is true by assumption for $t=T$, this would finish the proof.
    \\
    We have 
    \begin{equation}
        p(y_{t-1} \mid y_0, f_\phi(x)) = 
        \int_{\Ree^d} p(y_{t-1} \mid y_0, y_t, f_\phi(x)) 
        p(y_t \mid y_0, f_\phi(x)) dy_t,
    \end{equation}
    where
    \begin{equation}
        p(y_{t-1} \mid y_0, y_t, f_\phi(x)) = 
        \cN(
        \sqrt{\bar{\alpha}_{t-1}}y_0 + 
        \sqrt{ 1 - \bar{\alpha}_{t-1} - \sigma_t^2} \cdot 
        \frac{y_t - \left(\sqrt{\bar{\alpha}_t}y_0 +  
        \left(1 - \sqrt{\bar{\alpha}_t} \right)f_\phi(x) \right)}
        {\sqrt{1 - \bar{\alpha}_t}} , \sigma_t^2 \mathbf{I}),
    \end{equation}
    corresponds to \eqref{background:diffusion:reverse_process_def} just using the mean defined in \eqref{numerical_exps:uci:card_mean_formulation}.
    Now,
    \begin{equation}
         p(y_t \mid y_0, f_\phi(x)) = \cN \left(\sqrt{\bar{\alpha}_t}y_0 + \left(1 - \sqrt{\bar{\alpha}_t}f_\phi(x) \right),
        \left(1 - \bar{\alpha}_t \right) \mathbf{I} \right)
    \end{equation}
    holds because of the induction hypothesis.
    By theorem~\ref{appendix:thm:gaussian_thm}, we have that $p(y_{t-1} \mid y_0, y_t, f_\phi(x))$ is Gaussian with mean $\mu_{t-1}$ and covariance matrix $\Sigma_{t-1}$, where
    \begin{align}
        \mu_{t-1} &= 
            \sqrt{\bar{\alpha}_{t-1}}y_0 + 
            \left( 1 - \sqrt{\bar{\alpha}_{t-1}} \right)f_\phi(x) + \\
            &\quad
            \sqrt{ 1 - \bar{\alpha}_{t-1} - \sigma_t^2} \cdot 
            \frac{\sqrt{\bar{\alpha}_t}y_0 + \left(1 - \sqrt{\bar{\alpha}_t}f_\phi(x) \right) - \left(\sqrt{\bar{\alpha}_t}y_0 +  
            \left(1 - \sqrt{\bar{\alpha}_t} \right)f_\phi(x) \right)}
            {\sqrt{1 - \bar{\alpha}_t}} \\
            &= \sqrt{\bar{\alpha}_{t-1}}y_0 + 
            \left( 1 - \sqrt{\bar{\alpha}_{t-1}} \right)f_\phi(x)
    \end{align}
    and 
    \begin{align}
        \Sigma_{t-1} &=
        \sigma_t^2 \mathbf{I} + 
        \frac{1 - \bar{\alpha}_{t-1} - \sigma_t^2}{1 - \bar{\alpha}_t}
        (1- \bar{\alpha}_t) \mathbf{I} = (1-  \bar{\alpha}_{t-1}) \mathbf{I}
    \end{align}
    which finishes the proof.
\end{proof}

\section{Experiment details}
\label{app:experiment_details}

\subsection{Evaluation metrics}
Consider the true target $\bm y \in \mathcal{Y}\subseteq \Ree^{d_y}$, the corresponding marginals $y^k, \ k=1,\ldots, d_y$, the true predictive distribution $p_\mathcal{Y}(\cdot \mid \bm c)$ and the learned approximate predictive distribution $p_\theta(\cdot \mid \bm c)$. Furthermore, define $\hat{\bm y}_i \sim p_\theta(\cdot \mid \bm c), \ i = 1, \ldots, M$ and the empirical predictive distribution $p_\theta^M \coloneq \{ \hat{\bm y}_m \}_{m=1}^M$. Finally, let $\bar{\bm y} \coloneq \frac{1}{M} \sum_{m=1}^M \hat{\bm y}_m$ denote the empirical mean, $\hat{\sigma}^2_k \coloneq \frac{1}{M-1} \sum_{m=1}^M (\hat{\bm y}_m - \bar{\bm y})^2$ denote the empirical variance of the $k^{th}$ marginal and $(\hat{q}_\theta^\alpha)^k$ denote the empirical quantiles of the $k^{th}$ marginal of $p_\theta^M$ at the level $\alpha$. We use the following evaluation metrics:
\begin{align}
    \mathrm{RMSE}(p_\theta^M, \bm y) &\coloneq \|\bar{\bm y} - \bm y \|_2, \\
    \mathrm{ES}(p_\theta^M, \bm y) & \coloneq \frac{1}{M} \sum_{m=1}^M \|\hat{\bm y}_i - \bm y \|_2 - \frac{1}{2M(M-1)}\sum_{\substack{m, h = 1 \\ m \neq h}}^M \|\hat{\bm y}_m - \hat{\bm y }_h \|_2, \\
    \mathrm{CRPS}(p_\theta^M, \bm y) & \coloneq \frac{1}{d_y} \sum_{k=1}^{d_y} \left( \frac{1}{M} \sum_{m=1}^M \left|\hat{y}_i^k -  y^k \right| - \frac{1}{2M(M-1)}\sum_{\substack{m, h = 1 \\ m \neq h}}^M \left|\hat{ y}_m^k - \hat{ y }_h^k \right| \right), \\
     \mathrm{NLL}(p_\theta^M, \bm y) & \coloneq  \frac{1}{d_y} \sum_{k=1}^{d_y} \log \left( 2 \pi \hat{\sigma}^2_k \right) + \frac{((\bar{\bm y})^k - y^k)^2}{\hat{\sigma}^2_k}, \\
     \mathcal{C}_\alpha(p_\theta^M, \bm y) & \coloneq \frac{1}{d_y} \sum_{k=1}^{d_y} \mathbbm{1} \left\{ y^k \in [(\hat{q}_\theta^{\alpha/2})^k , (\hat{q}_\theta^{1-\alpha/2})^k] \right\}.
\end{align}
The RMSE evaluates the match between the mean prediction $\bar{\bm y}$ and the true observation, while the energy score (ES) evaluates the match for the predictive distribution as a whole. The continuous ranked probability score (CRPS) \citep{Gneiting.2007} evaluates the predictive distribution at a pointwise level and assesses whether the predicted uncertainty fits the observations at all quantile levels for each marginal $y^k$, i.e., whether the predictive distribution is well-calibrated. Furthermore, we analyze the negative log-likelihood (NLL) of a predictive pointwise Gaussian distribution. Although the prediction can be inherently non-Gaussian, this criterion is commonly used and describes the fit of the predictive distribution in terms of the first two estimated moments. Finally, we report the coverage $\mathcal{C}_\alpha$ of the predictive distribution for selected $\alpha$-quantile levels, averaged over all marginals.

\subsection{Training details} \label{app:experiment-details-training-details}
All of our proposed methods require some restricted output for the parameters, which we realize in the following way. For the marginal $\sigma_{\theta,k}^2(\bm x_t, t)$ we use one additional last layer with a softplus activation and add a threshold of $10^{-6}$ for numerical stability. For the mixture weights $\pi_{\theta, k}$ we apply a softmax activation to guarantee the summation constraint. For the $\bm \epsilon_t^\mathrm{ES}$ method we concatenate the random noise as an additional channel dimension. For the multivariate normal method, implementation is already described in Section~\ref{app:covariance_approximations}. The remaining training details, such as batch size or number of epochs, differ across experiments and are described in the next sections, respectively.

\subsection{UCI regression} \label{app:experiment_details:uci}
To obtain a fair comparison, we take the train-test splits from \citet{pmlr-v37-hernandez-lobatoc15} and do not finetune our methods on a separate validation set but use sensible hyperparameters.
We will copy the training process from the CARD method and apply it, except for some minor modifications to the number of training epochs used for our distributional methods.
In particular, we train all of our methods for $5000$ epochs except on the yacht dataset, where we train for $10000$.
We only deviate from this for the larger kin8nm and protein datasets, where we train for $1000$ and $2000$ epochs, respectively.
We use the Adam optimizer \citep{kingma2017adammethodstochasticoptimization} with a learning rate of $0.001$.
For the diffusion process, we decided to use $50$ timesteps, both during training and inference.
We use a linear noise schedule with $\beta_1 = 0.001$ and $\beta_{50}=0.35$, which is the noise schedule used to train the CARD method for $50$ timesteps.
Lastly, we set $K=3$, when training the $\bm\Sigma_\theta^{\mathrm{mix}}$ method and use for all of our distributional methods the CRPS score as scoring rule.

\subsection{Autoregressive prediction tasks}
\label{app:details_autoregressive}
\paragraph{Burgers' equation}
The Burgers' equation is given as
\begin{equation}
\begin{split}
    \partial_s u(s,x) + \partial_x (u^2(s,x) /2) &= \nu / \pi \partial_{xx}u(s,x), \quad x \in (0,1), \ s\in (0,2]\\
    u(0,x) &= u_0(x), \quad x \in (0,1)
\end{split}    
\end{equation}
where $u \in C([0,S]; H_{\mathrm{per}}^r((0,1); \Ree))$ for any $r>0$, $u_0 \in L_{\mathrm{per}}^2((0,1); \Ree)$ is the initial condition and $\nu \in \Ree_+$ is the diffusion coefficient\footnote{$H^r_{\text{per}}(\mathcal{D}; \mathbb{R}), L^2_{\text{per}}(\mathcal{D}; \mathbb{R})$ denote the periodic Sobolev and $L^2$ spaces, respectively.}. We utilize data from the PDEBENCH repository \citep{NEURIPS2022_0a974713}, which assumes a constant diffusion coefficient, which we choose as $\nu = 0.01$. The data is generated with a periodic boundary condition from a superposition of sinusoidal waves with the temporally and spatially 2nd-order upwind difference scheme for the advection term, and the central difference scheme for the diffusion term.

\paragraph{Kuramoto--Sivashinsky equation}
Recall that the Kuramoto--Sivashinsky (KS-) equation in one spatial dimension is given as:
\begin{alignat*}{2}  
\partial_s u(x,s)+ u \partial_x u(x,s) + \partial_x^2 u(x,s) + \partial_x^4 u(x,s) &= 0, \qquad &&x \in \mathcal{D}, s \in (0,S]\\
u(x,0) &= u_0(x), \qquad &&x \in \mathcal{D}
\end{alignat*}
where $\mathcal{D} \subseteq \mathbb{R}$, $u \in C([0,S]; H_{\mathrm{per}}^4(\mathcal{D}; \mathbb{R}))$, and $u_0 \in L_{\mathrm{per}}^2(\mathcal{D};\mathbb{R})$. We follow the setup in \citet{bultepno} and simulate the KS-equation from random uniform noise $\mathcal{U}(-1,1)$ on a periodic domain $\mathcal{D} = [0,100]$ using the py-pde package \citep{zwicker_py-pde_2020}. We generate 10000 samples with a resolution of $256 \times 50$ and $\Delta s = 2$.

\paragraph{Surface temperature prediction}
For the surface temperature prediction task, we utilize the ERA5 dataset \citep{hersbachERA5GlobalReanalysis2020} provided via the WeatherBench2 benchmark \citep{rasp2024weatherbench2benchmarkgeneration}. Similar to \citet{bultepno}, we use data with a spatial resolution of $0.25^\circ \times 0.25^\circ$ and a time resolution of $6h$. For computational reasons, we restrict the data to a European domain, covering an area from 35°N – 75°N and 12.5°W – 42.5°E with selected user-relevant weather variables (u-component and v-component of 10-m wind speed (U10 and V10), temperature at 2m and 850 hPa (T2M and T850), geopotential height
at 500 hPa (Z500), as well as land-sea mask and orography) that serve as input to the model. In contrast to other diffusion approaches \citep{larsson2025diffusionlamprobabilisticlimitedarea}, we only predict the surface temperature T2M, as this is less computationally demanding. As the initial condition, we always use 00 UTC time and issue a prediction for 06 UTC time.

\paragraph{Model details}
We adapt the U-Net used in \citet{songDenoisingDiffusionImplicit2022, karrasElucidatingDesignSpace2022} to allow for arbitrary input shapes and one-dimensional convolutions. We use 64 feature channels for the first layer and 128 channels for layers 2-4. The diffusion noise is encoded with Fourier embeddings \citep{karrasElucidatingDesignSpace2022}, where the noise is transformed into sine/cosine features at 32 frequencies with base period 16. Afterwards, the features are passed through a 2-layer MLP, which results in a 256-dimensional noise encoding. Finally, this encoding is added to the network through various group norms of the U-Net. The final (regular) diffusion model has 6.5 million and 15.2 million parameters for the 1D and 2D tasks, respectively. While the temperature prediction task uses a fixed prediction horizon, for the PDE tasks, we sample random timesteps $s \in 1, \ldots, S$ for training and evaluation, but split the data such that the test data is unseen during training.

\paragraph{Training details}
For the 1D PDE tasks, we use an effective batch size of 128, a learning rate of $5 \times 10^{-4}$ with the Adam optimizer, early stopping after 150 epochs and a learning rate scheduler that reduces the learning rate by the factor $0.5$ if the validation loss has not improved for $75$ epochs. All models are trained on an NVIDIA RTX3090 with 24GB memory.

For the T2M task, we use an effective batch size of 64, a learning rate of $1 \times 10^{-5}$ with the AdamW optimizer, early stopping after 300 epochs, and a learning rate scheduler that reduces the learning rate by the factor $0.5$ if the validation loss has not improved for $150$ epochs. All models are trained on two NVIDIA RTXA6000 with 48GB memory.

\subsection{Monocular depth estimation} \label{app:experiment-details-depth}

We follow the setup of Marigold \citep{ke2024repurposing} in general. We only adapt the last layer of their model to our methods as described in Appendix~\ref{app:experiment-details-training-details},  the alignment of the inference ensemble, and the used metrics. Details on the Marigold setup and our adaptations are shown in the following.

\paragraph{Model details}
The idea of Marigold \citep{ke2024repurposing} is to adapt Stable Diffusion \citep{rombach2022high} to depth estimation. The frozen VAE is used to encode the image and the corresponding depth map into the latent space for training. As the encoder expects an image with 3 channels, the depth map is replicated into three channels. The embedded RGB image is used as the conditioning $\bm c$, and the embedded depth map gets noise added to serve as $\bm x_t$. To handle this, the first layer of the UNet is duplicated, and the weights are halved to keep the scale of the activations similar. As we need multiple output channels to model the standard deviation as well, we adapt the last layer depending on the used method, as explained in Section~\ref{app:experiment-details-training-details}. We initialize the variance head randomly, while we initialize the weights of the mean prediction with the weights of the last convolutional layer of the Stable Diffusion UNet.

\paragraph{Training details} We follow the exact training procedure of Marigold \citep{ke2024repurposing}. Training is done on the synthetic datasets Hypersim \citep{roberts2021hypersim} and VKitti2 \citep{cabon2020virtual}. We use the same affine-invariant depth normalization and annealed multi-resolution noise as Marigold during training, while using Gaussian noise during inference. We perform 1000 diffusion steps with the DDPM scheduler \citep{hoDenoisingDiffusionProbabilistic2020}. We train using an effective batch size of 32 with Adam \citep{kingma2017adammethodstochasticoptimization} and convergence takes about 3 days on a single Nvidia RTX GPU. We train all models with the same setup and also retrain Marigold ($\bm\delta_\theta$).

\paragraph{Evaluation datasets}
The evaluation is performed on five real-world datasets. NYUv2 \citep{silberman2012nyu} and  ScanNet \citep{dai2017scannet} are both indoor scene datasets captured with an RGB-D Kinect sensor. NYUv2 has a designated test split with 654 images, which we use. For ScenNet, we sample 800 images from the 312 official validation scenes for testing. The street-scene dataset KITTI \citep{geiger2012we} comes with a sparse metric depth captured by a LiDAR sensor and we use the Eigen test split \citep{eigen2014depth} made of 652 images. ETH3D \citep{schops2017multi} and DIODE \citep{vasiljevic2019diode} are high-resolution datasets. We utilize the entire ETH3D dataset and the entire validation split of DIODE, which comprises 325 indoor and 446 outdoor samples.

\paragraph{Evaluation details} 
We use $T=50$ denoising steps and an ensemble size of $N=10$ for every method. Marigold is tackling affine-invariant depth estimation and thus, the predictions have to be transformed before the evaluation. Furthermore, Marigold proposed to use an ensemble prediction to boost the performance and aligned the sample predictions $\{\bm{\hat{d}^1},...,\bm{\hat{d}^N}\}$ among each other by minimizing the following objective:

\[
\min_{\substack{s_1, \ldots, s_N \\ t_1, \ldots, t_N}} 
\left( 
\sqrt{ \frac{1}{b} \sum_{i=1}^{N-1} \sum_{j=i+1}^{N} \| \bm{\hat{d}^{i'}} - \bm{\hat{d}^{j'}} \|_2^2 } 
+ \lambda \mathcal{R} 
\right)
\]
where $\bm{\hat{d}}^{i'}=\hat{s}^i\bm{\hat{d}}^i+\hat{t}^i$ are the scaled and shifted predictions, $\mathcal{R}=|\min(m)|+|1-\max(m)|$ is a regularization term with $\bm{m}(x,y)=\text{median}(\bm{\hat{d}}^{1'}(x,y),...,\bm{\hat{d}}^{N'}(x,y))$. Furthermore, $b=\binom{N}{2}$ and $\lambda$ is a hyperparameter. Then, the median $\bm{m}$ is taken and aligned with the target depth map by a least squares fit $\bm{a}=s\bm{m} + t$. We follow Marigold and perform the same steps for every method to get the point prediction $\bm{a}$. As we are also interested in metrics that consider the whole ensemble, such as ES and CRPS, further align the ensemble by setting $\bm{a}^i = s\bm{\hat{d}}^{i'}+t$, i.e., we align each ensemble member with the same transformation that we computed for the median $\bm{\hat{d}}$. We evaluate the prediction using the UQ metrics ES and CRPS and the metrics reported in Marigold ($\bm{d}=(d_i)_{i=1,...,M}$ denotes the true depth map and $M$ the number of pixels):
\begin{itemize}
    \item Absolute Mean Relativ Error (\textit{AbsRel}): $\frac{1}{M}\Sigma_{i=1}^M|a_i-d_i|/d_i$ 
    \item $\delta_1$ Accuracy: Proportion of pixels satisfying $\max(a_i/d_i,d_i/a_i)<1.25$
\end{itemize}

\section{Additional results}
\label{app:additional_results}

\begin{figure}[ht]
\begin{subfigure}{\linewidth}
        \centering
        \caption{$\bm \delta_\theta$}
    \includegraphics[width=0.9\linewidth]{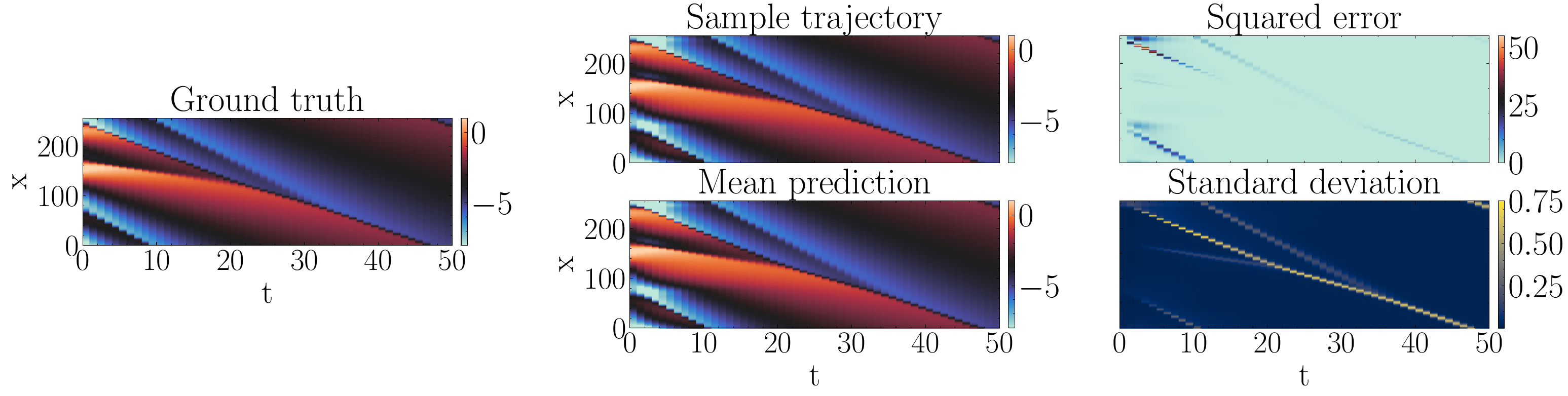}
\end{subfigure}
\begin{subfigure}{\linewidth}
        \centering
        \caption{$\bm\Sigma_\theta^\mathrm{diag}$}
    \includegraphics[width=0.9\linewidth]{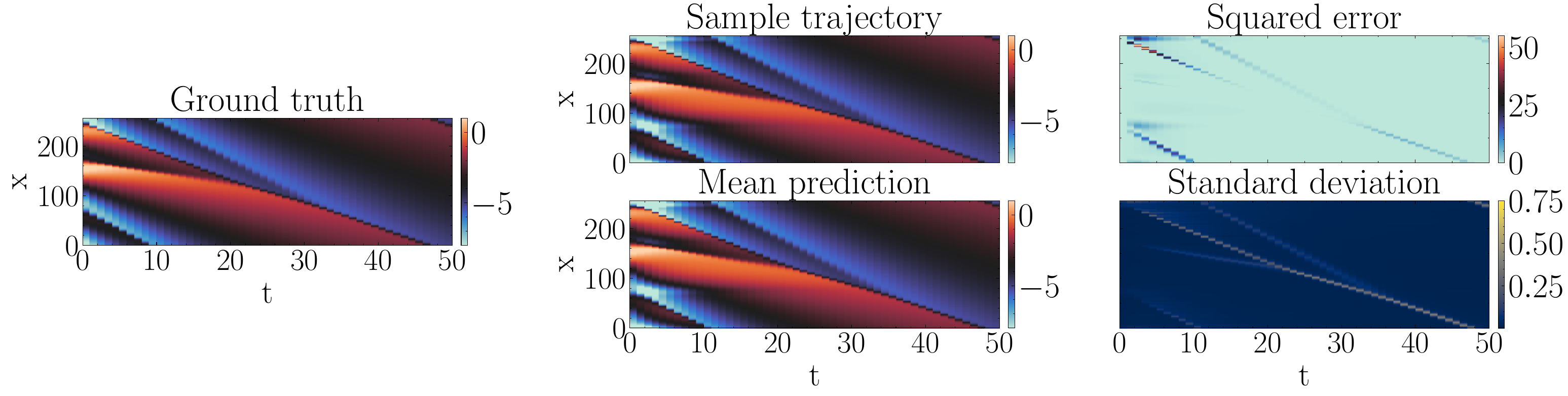}
\end{subfigure}
\begin{subfigure}{\linewidth}
        \centering
        \caption{$\bm\Sigma_\theta^\mathrm{mix}$}
    \includegraphics[width=0.9\linewidth]{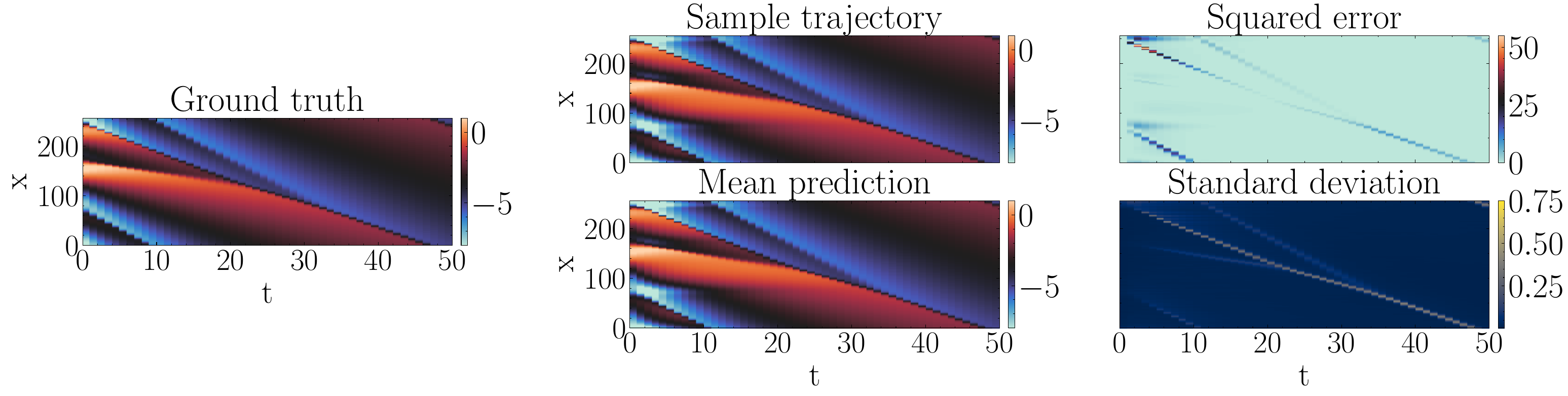}
\end{subfigure}
\begin{subfigure}{\linewidth}
        \centering
        \caption{$\bm\Sigma_\theta^\mathrm{mv}$}
    \includegraphics[width=0.9\linewidth]{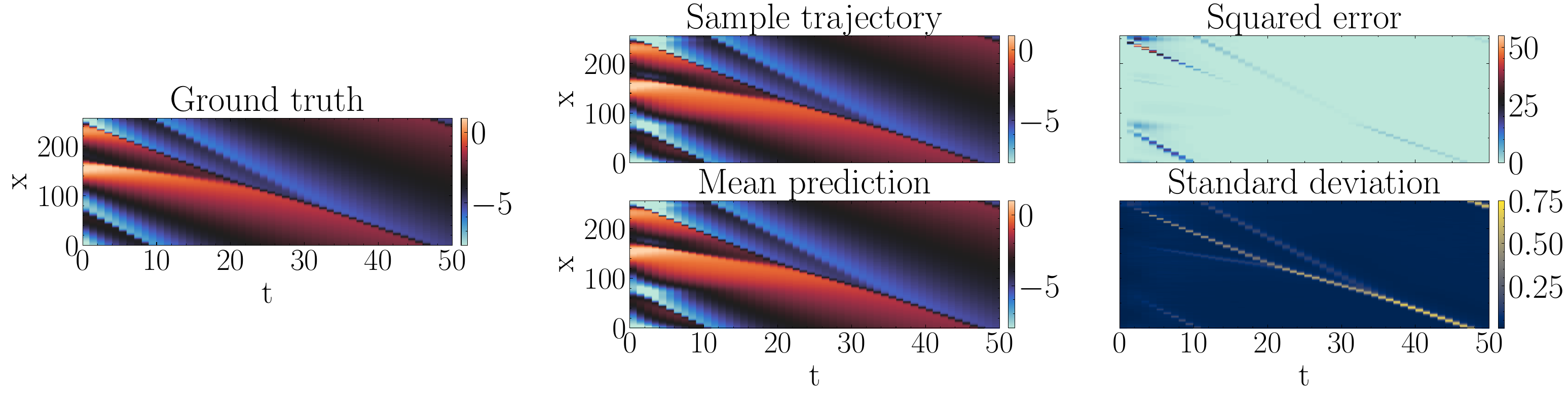}
\end{subfigure}
\begin{subfigure}{\linewidth}
        \centering
        \caption{$\bm\epsilon_t^\mathrm{ES}$}
    \includegraphics[width=0.9\linewidth]{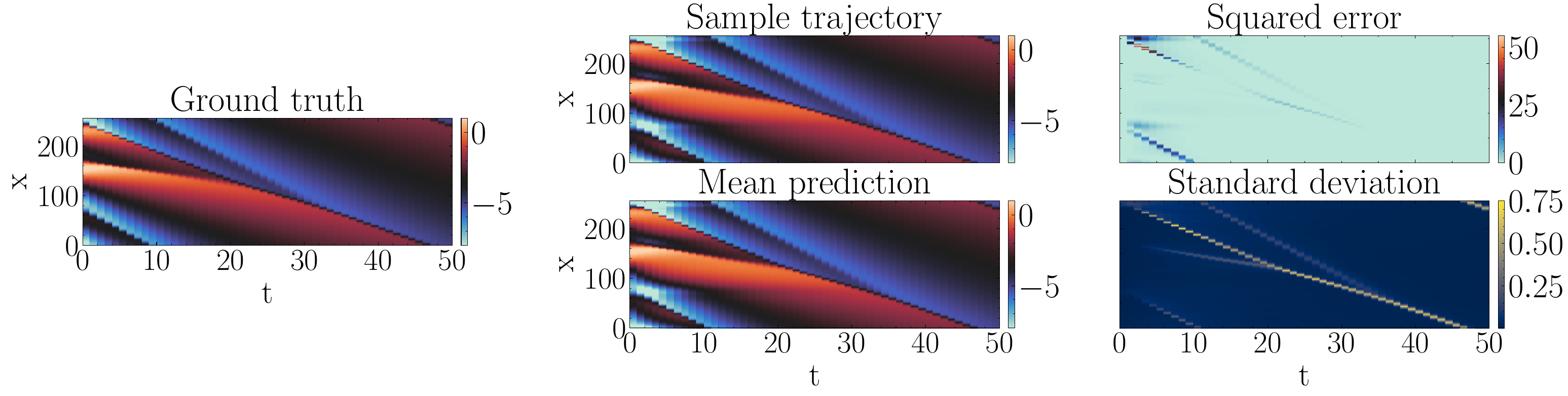}
\end{subfigure}
    \caption{Comparison of the predicted autoregressive trajectories of the Burgers' equation for the different models.}
    \label{fig:burgers_visualization_all}
\end{figure}
\begin{figure}[ht]
\begin{subfigure}{\linewidth}
        \centering
        \caption{$\bm \delta_\theta$}
    \includegraphics[width=0.9\linewidth]{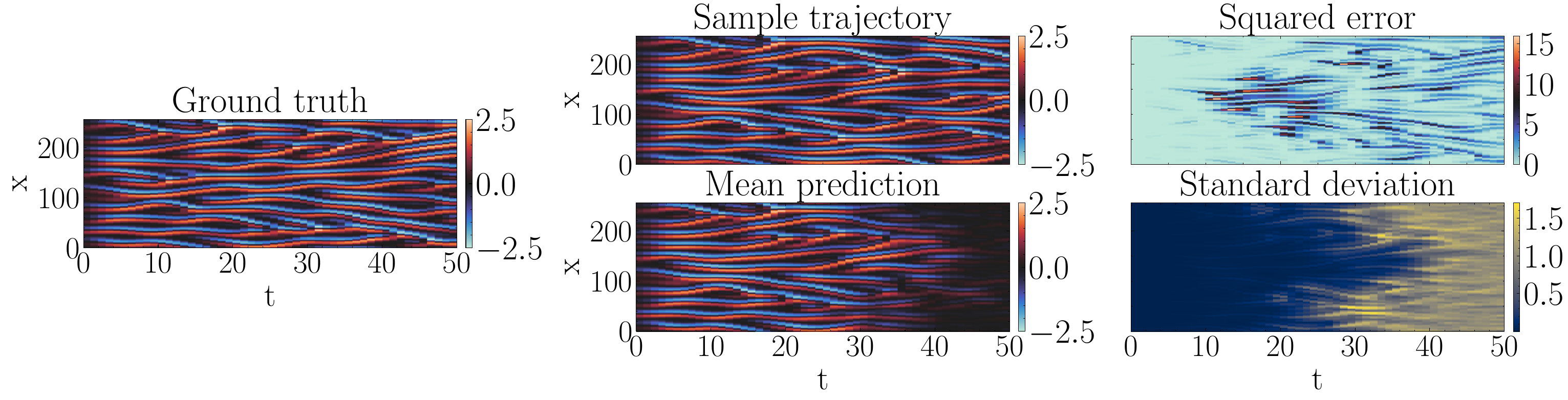}
\end{subfigure}
\begin{subfigure}{\linewidth}
        \centering
        \caption{$\bm\Sigma_\theta^\mathrm{diag}$}
    \includegraphics[width=0.9\linewidth]{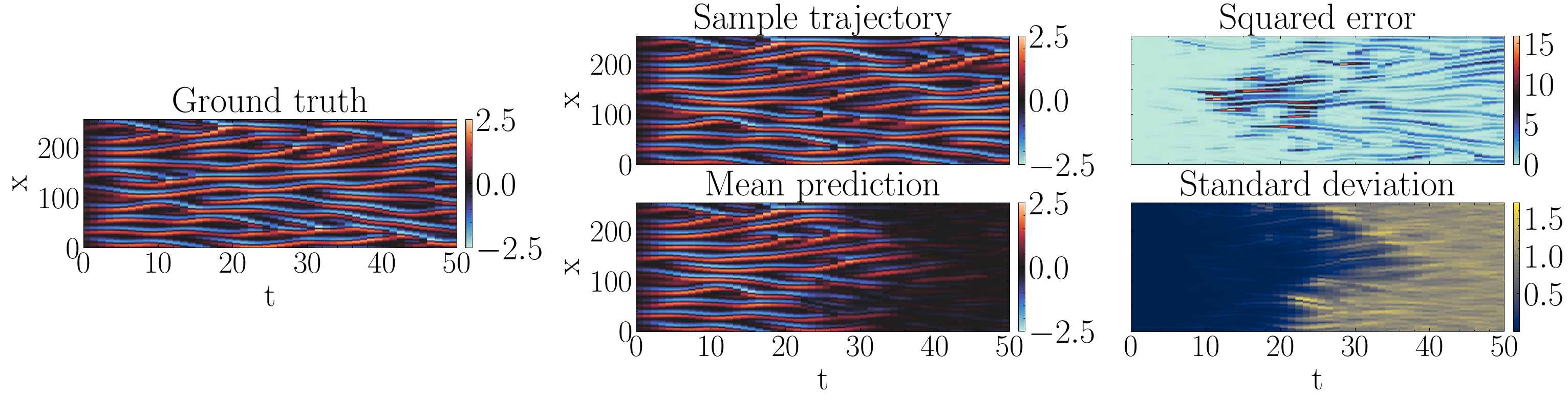}
\end{subfigure}
\begin{subfigure}{\linewidth}
        \centering
        \caption{$\bm\Sigma_\theta^\mathrm{mix}$}
    \includegraphics[width=0.9\linewidth]{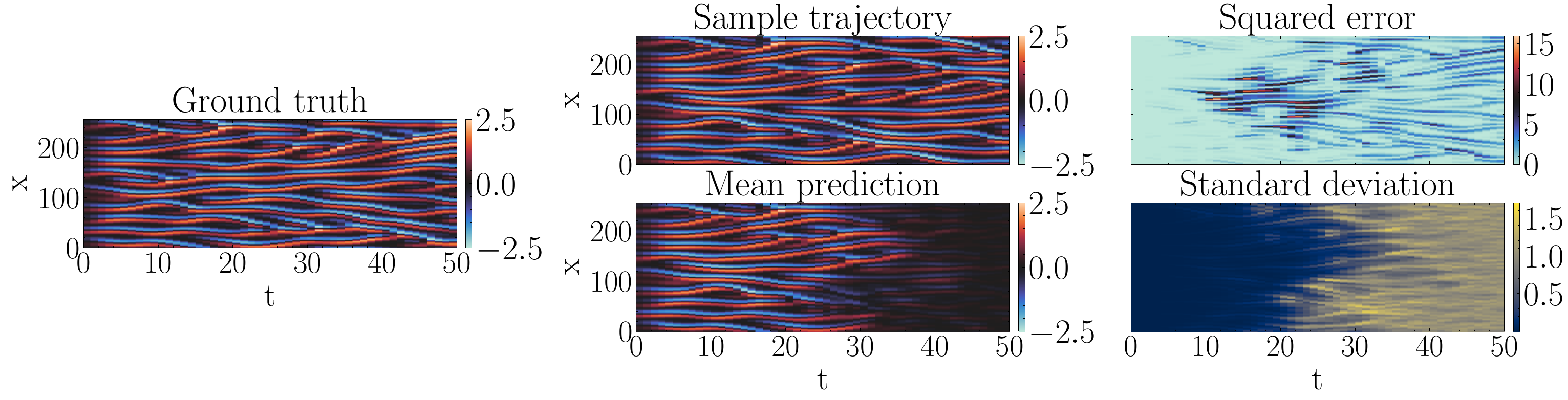}
\end{subfigure}
\begin{subfigure}{\linewidth}
        \centering
        \caption{$\bm\Sigma_\theta^\mathrm{mv}$}
    \includegraphics[width=0.9\linewidth]{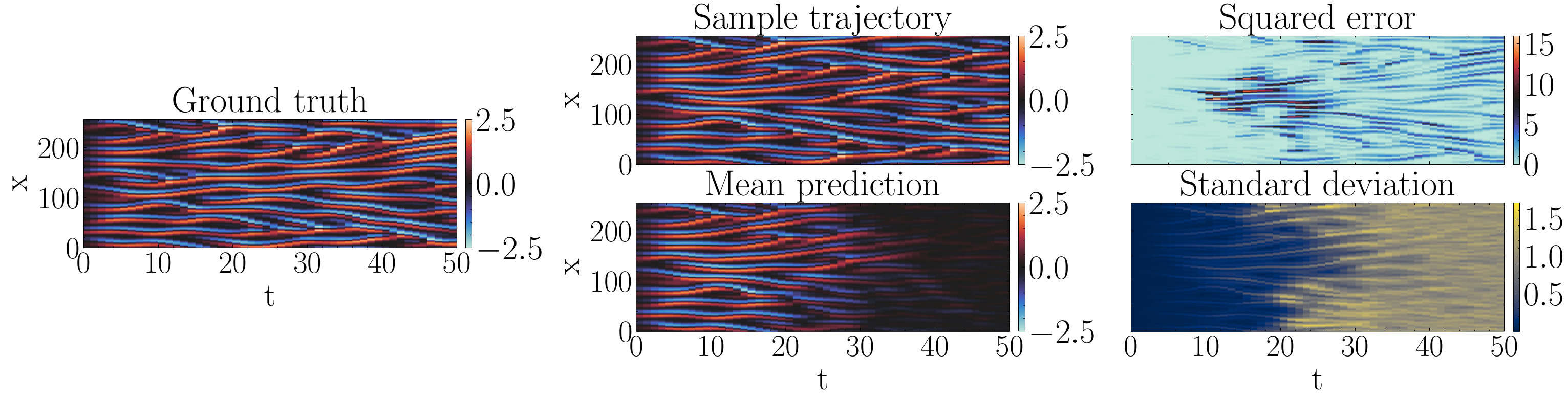}
\end{subfigure}
\begin{subfigure}{\linewidth}
        \centering
        \caption{$\bm\epsilon_t^\mathrm{ES}$}
    \includegraphics[width=0.9\linewidth]{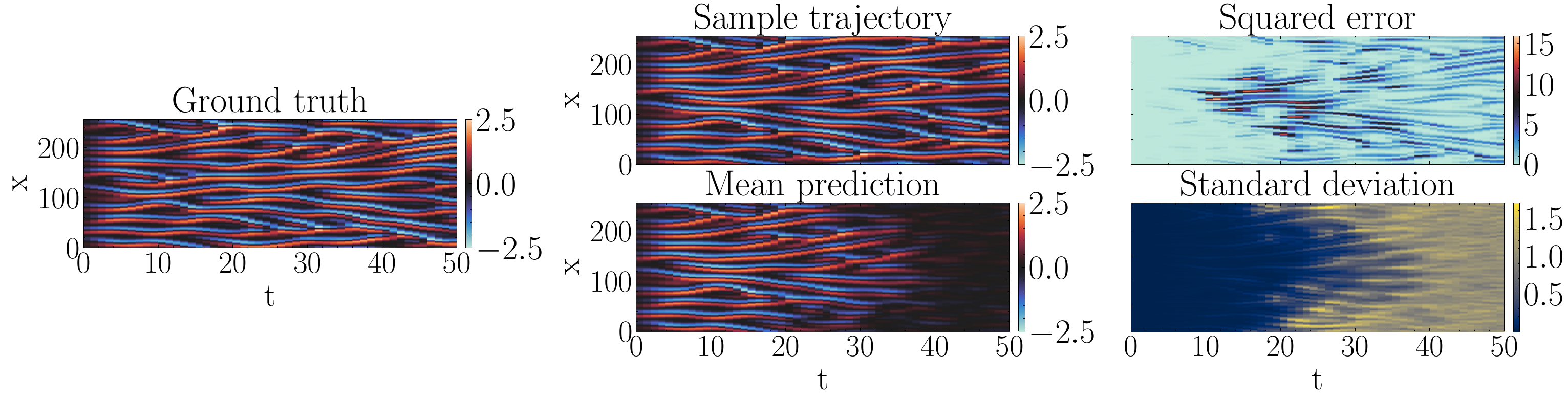}
\end{subfigure}
    \caption{Comparison of the predicted autoregressive trajectories of the KS equation for the different models.}
    \label{fig:ks_visualization_all}
\end{figure}
\begin{figure}[ht]
\begin{subfigure}{\linewidth}
        \centering
        \caption{$\bm \delta_\theta$}
    \includegraphics[width=\linewidth]{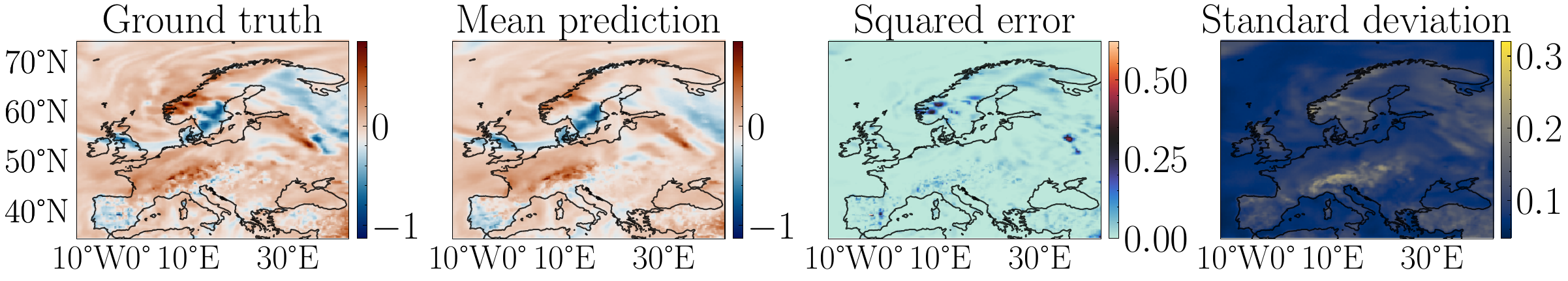}
\end{subfigure}
\begin{subfigure}{\linewidth}
        \centering
        \caption{$\bm\Sigma_\theta^\mathrm{diag}$}
    \includegraphics[width=\linewidth]{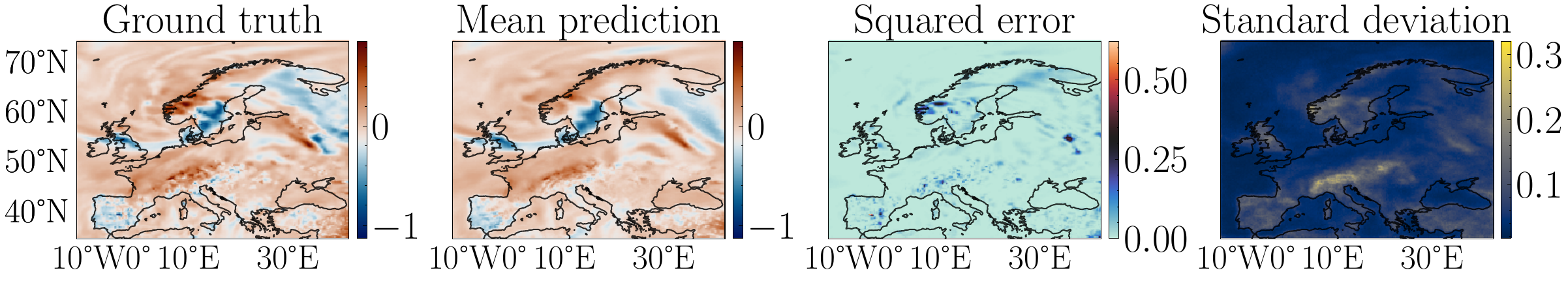}
\end{subfigure}
\begin{subfigure}{\linewidth}
        \centering
        \caption{$\bm\Sigma_\theta^\mathrm{mix}$}
    \includegraphics[width=\linewidth]{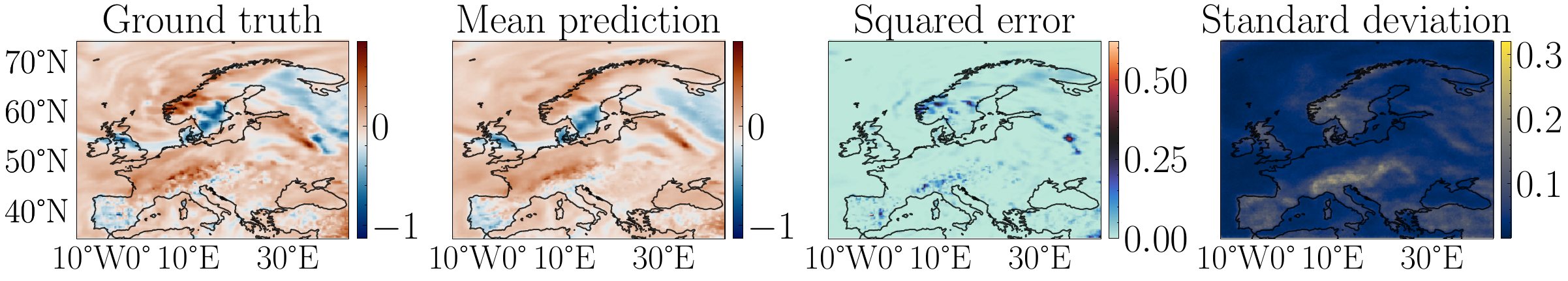}
\end{subfigure}
\begin{subfigure}{\linewidth}
        \centering
        \caption{$\bm\Sigma_\theta^\mathrm{mv}$}
    \includegraphics[width=\linewidth]{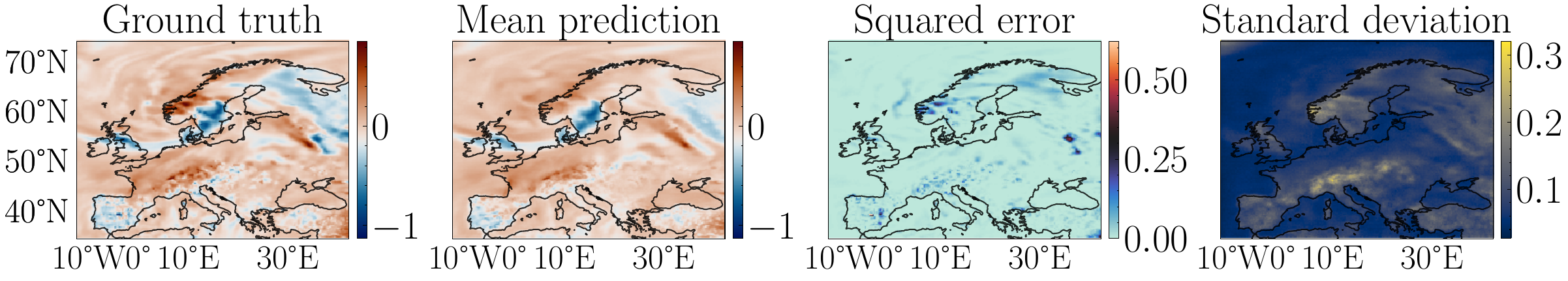}
\end{subfigure}
    \caption{Comparison of the predicted dynamics of the two-meter surface temperature for the different models.}
    \label{fig:t2m_visualization_all}
\end{figure}

This section provides more detailed results of the different experiments. 
Tables~\ref{appendix:add_results:card_rmse_table}, \ref{appendix:add_results:card_crps_table}, \ref{appendix:add_results:card_coverage_table} and \ref{appendix:add_results:card_nll_table} outline several considered metrics for the UCI dataset with standard deviations.
Table~\ref{tab:full_results_autoregressive} shows the results for the different autoregressive prediction tasks, averaged over five runs (except for T2M). While we report the average time per training epoch, the underlying algorithms are not optimized and might heavily depend on the used architecture and the availability of the compute cluster. They should therefore only be seen as a rough estimate for identifying methods that take a significantly longer time for training. Corresponding visualizations of the different methods can be found in Figures~\ref{fig:burgers_visualization_all}, \ref{fig:ks_visualization_all}, and \ref{fig:t2m_visualization_all}. 
Table~\ref{tab:full_results_depth-regression} shows more results of the depth estimation task.

\begin{table}[ht] 
\centering
\footnotesize
\caption{RMSE ($\downarrow$) on the test set for UCI datasets, averaged and with standard deviation.}
\label{appendix:add_results:card_rmse_table}
\begin{tabular}{l|c|c|c|c}
\toprule
 & $\bm\delta_\theta$ & $\bm\Sigma_\theta^{\mathrm{diag}}$ & $\bm\Sigma_\theta^{\mathrm{mix}}$ & $\bm\epsilon_t^\mathrm{ES}$ \\
\midrule
Concrete & 4.81 ± 0.63 & \textbf{4.72 ± 0.7} & 4.73 ± 0.63 & 4.81 ± 0.65 \\
Energy & 0.45 ± 0.06 & 0.45 ± 0.07 & \textbf{0.44 ± 0.06} & 0.45 ± 0.06 \\
Kin8nm & 6.91 ± 0.19 & \textbf{6.88 ± 0.19} & 6.89 ± 0.19 & 6.9 ± 0.19 \\
Naval & 1.35 ± 0.1 & 1.24 ± 0.09 & \textbf{1.23 ± 0.09} & \textbf{1.23 ± 0.09} \\
Power & 3.86 ± 0.17 & 3.64 ± 0.19 & \textbf{3.59 ± 0.17} & 3.78 ± 0.17 \\
Protein & 3.76 ± 0.04 & \textbf{3.71 ± 0.05} & 3.72 ± 0.05 & 3.76 ± 0.04\\
Wine & 0.67 ± 0.08 & 0.67 ± 0.08 & \textbf{0.66 ± 0.07} & \textbf{0.66 ± 0.08} \\
Yacht & 0.74 ± 0.4 & \textbf{0.7 ± 0.31} & 0.78 ± 0.5 & 0.8 ± 0.49 \\
\bottomrule
\end{tabular}
\end{table}

\begin{table}[ht] 
\centering
\footnotesize
\caption{CRPS($\downarrow$) on the test set for UCI datasets, averaged and with standard deviation.}
\label{appendix:add_results:card_crps_table}
\begin{tabular}{l|c|c|c|c}
\toprule
 & $\bm\delta_\theta$ & $\bm\Sigma_\theta^{\mathrm{diag}}$ & $\bm\Sigma_\theta^{\mathrm{mix}}$ & $\bm\epsilon_t^\mathrm{ES}$ \\
\midrule
Concrete & 2.6 ± 0.33 & 2.46 ± 0.37 & \textbf{2.45 ± 0.34} & 2.5 ± 0.34 \\
Energy & 0.3 ± 0.03 & 0.25 ± 0.03 & \textbf{0.24 ± 0.03} & 0.25 ± 0.03 \\
Kin8nm & 3.93 ± 0.11 & \textbf{3.81 ± 0.1} & 3.82 ± 0.1 & 3.83 ± 0.1 \\
Naval & 0.79 ± 0.1 & 0.56 ± 0.06 & 0.54 ± 0.05 & \textbf{0.53 ± 0.05} \\
Power & 2.04 ± 0.05 & 1.84 ± 0.06 & \textbf{1.81 ± 0.06} & 1.93 ± 0.07 \\
Protein & 1.71 ± 0.02 & 1.66 ± 0.03 & \textbf{1.65 ± 0.03} & 1.69 ± 0.02\\
Wine & 0.34 ± 0.06 & 0.34 ± 0.06 & 0.34 ± 0.06 & \textbf{0.33 ± 0.06} \\
Yacht & 0.36 ± 0.17 & \textbf{0.28 ± 0.12} & 0.31 ± 0.18 & 0.32 ± 0.19 \\
\bottomrule
\end{tabular}
\end{table}

\begin{table}[ht] 
\centering
\footnotesize
\caption{$\mathcal{C}_{0.95}$ on the test set for UCI datasets, averaged and with standard deviation.}
\label{appendix:add_results:card_coverage_table}
\begin{tabular}{l|c|c|c|c}
\toprule
 & $\bm\delta_\theta$ & $\bm\Sigma_\theta^{\mathrm{diag}}$ & $\bm\Sigma_\theta^{\mathrm{mix}}$ & $\bm\epsilon_t^\mathrm{ES}$ \\
\midrule
Concrete & 0.57 ± 0.05 & 0.68 ± 0.05 & 0.68 ± 0.06 & \textbf{0.73 ± 0.05} \\
Energy & 0.28 ± 0.12 & \textbf{0.73 ± 0.06} & 0.72 ± 0.06 & 0.72 ± 0.07 \\
Kin8nm & 0.77 ± 0.02 & \textbf{0.86 ± 0.02} & \textbf{0.86 ± 0.01} & \textbf{0.86 ± 0.01} \\
Naval & 0.25 ± 0.09 & 0.93 ± 0.03 & \textbf{0.94 ± 0.02} & 0.92 ± 0.03 \\
Power & 0.82 ± 0.02 & 0.89 ± 0.01 & 0.87 ± 0.02 & \textbf{0.9 ± 0.01} \\
Protein & 0.86 ± 0.01 & \textbf{0.91 ± 0.0} & 0.89 ± 0.01 & \textbf{0.91 ± 0.01} \\
Wine & 0.53 ± 0.12 & 0.76 ± 0.05 & 0.7 ± 0.08 & \textbf{0.79 ± 0.04} \\
Yacht & 0.34 ± 0.12 & 0.78 ± 0.09 & 0.78 ± 0.09 & \textbf{0.79 ± 0.1} \\
\bottomrule
\end{tabular}
\end{table}

\begin{table}[ht] 
\centering
\footnotesize
\caption{NLL($\downarrow$) on the test set for UCI datasets, averaged and with standard deviation.}
\label{appendix:add_results:card_nll_table}
\begin{tabular}{l|c|c|c|c}
\toprule
 & $\bm\delta_\theta$ & $\bm\Sigma_\theta^{\mathrm{diag}}$ & $\bm\Sigma_\theta^{\mathrm{mix}}$ & $\bm\epsilon_t^\mathrm{ES}$ \\
\midrule
Concrete & 10.35 ± 5.99 & 9.63 ± 6.53 & 7.93 ± 4.1 & \textbf{4.97 ± 1.84} \\
Energy & 63.59 ± 85.17 & \textbf{1.6 ± 1.2} & 1.62 ± 1.28 & 1.63 ± 1.27 \\
Kin8nm & -0.99 ± 0.09 & \textbf{-1.22 ± 0.05} & \textbf{-1.22 ± 0.04} & \textbf{-1.22 ± 0.04} \\
Naval & 24.71 ± 26.16 & -7.94 ± 0.13 & \textbf{-7.98 ± 0.13} & -7.93 ± 0.17 \\
Power & 2.98 ± 0.12 & 2.84 ± 0.19 & 2.92 ± 0.3 & \textbf{2.77 ± 0.08} \\
Protein & 3.21 ± 0.36 & 3.19 ± 0.34 & 3.8 ± 0.45 & \textbf{2.91 ± 0.13} \\
Wine & 3903.17 ± 5036.57 & 173.79 ± 211.5 & 865.12 ± 880.86 & \textbf{116.0 ± 138.55} \\
Yacht & 55.75 ± 65.65 & 3.39 ± 7.32 & 6.34 ± 13.65 & \textbf{2.58 ± 4.93} \\
\bottomrule
\end{tabular}
\end{table}

\begin{table}[ht]
    \centering
        \caption{Results for the autoregressive prediction tasks. The RMSE, ES and CRPS are scaled by the factor 1000 (100) for the Burgers' (KS) equation. The best model is highlighted in bold and the standard deviation across the different runs is shown in brackets.}
    \label{tab:full_results_autoregressive}
    \begin{tabular}{|l|l|cccccc|}
    \toprule
        Experiment & Model & $t_{epoch} [s]$ & RMSE$\downarrow$ & ES$\downarrow$ & CRPS$\downarrow$ & NLL$\downarrow$ & $\mathcal{C}_{0.95}$ \\
        \midrule
        \multirow{9}{*}{\textbf{Burgers'}}
        & $\bm\delta_\theta$&  \makecell{6.94 \\ ($\pm$ 0.24)} & \makecell{0.95 \\ ($\pm$ 0.39)} & \makecell{4.26 \\ ($\pm$ 0.71)} & \makecell{0.16 \\ ($\pm$ 0.03)} & \makecell{-7.09 \\ ($\pm$ 0.17)} & \makecell{\textbf{0.94} \\ ($\pm$ 0.01)} \\
        & $\bm\Sigma_\theta^{\mathrm{diag}}$ & \makecell{\textbf{6.81} \\ ($\pm$ 0.16)} & \makecell{0.81 \\ ($\pm$ 0.27)} & \makecell{3.67 \\ ($\pm$ 0.39)} & \makecell{\textbf{0.12} \\ ($\pm$ 0.02)} & \makecell{\textbf{-7.12} \\ ($\pm$ 0.09)} & \makecell{1.00 \\ ($\pm$ 0.00)} \\
        & $\bm\Sigma_\theta^{\mathrm{mix}}$ & \makecell{7.38 \\ ($\pm$ 0.38)} & \makecell{0.81 \\ ($\pm$ 0.27)} & \makecell{3.82 \\ ($\pm$ 0.39)} & \makecell{0.13 \\ ($\pm$ 0.02)} & \makecell{-6.99 \\ ($\pm$ 0.19)} & \makecell{1.00 \\ ($\pm$ 0.00)} \\
        & $\bm\Sigma_\theta^{\mathrm{mv}}$ & \makecell{7.84 \\ ($\pm$ 0.39)} & \makecell{\textbf{0.70} \\ ($\pm$ 0.20)} & \makecell{6.35 \\ ($\pm$ 0.32)} & \makecell{0.24 \\ ($\pm$ 0.02)} & \makecell{-6.10 \\ ($\pm$ 0.04)} & \makecell{1.00 \\ ($\pm$ 0.00)} \\
        & $\bm\epsilon_t^\mathrm{ES}$  & \makecell{10.69 \\ ($\pm$ 0.20)} & \makecell{0.81 \\ ($\pm$ 0.49)} & \makecell{\textbf{3.59} \\ ($\pm$ 1.05)} & \makecell{0.14 \\ ($\pm$ 0.03)} & \makecell{-7.00 \\ ($\pm$ 0.14)} & \makecell{0.99 \\ ($\pm$ 0.00)} \\
        \midrule
        \multirow{9}{*}{\textbf{KS}}
        & $\bm\delta_\theta$ & \makecell{\textbf{6.49 }\\ ($\pm$ 0.12)} & \makecell{0.56 \\ ($\pm$ 0.06)} & \makecell{5.93 \\ ($\pm$ 0.62)} & \makecell{0.24 \\ ($\pm$ 0.02)} & \makecell{-3.99 \\ ($\pm$ 0.17)} & \makecell{0.84 \\ ($\pm$ 0.05)} \\
        & $\bm\Sigma_\theta^{\mathrm{diag}}$ & \makecell{7.54 \\ ($\pm$ 0.28)} & \makecell{0.39 \\ ($\pm$ 0.06)} & \makecell{5.23 \\ ($\pm$ 0.75)} & \makecell{0.21 \\ ($\pm$ 0.03)} & \makecell{-4.04 \\ ($\pm$ 0.14)} & \makecell{1.00 \\ ($\pm$ 0.00)} \\
        & $\bm\Sigma_\theta^{\mathrm{mix}}$ & \makecell{10.46 \\ ($\pm$ 0.11)} & \makecell{\textbf{0.35} \\ ($\pm$ 0.04)} & \makecell{\textbf{4.91} \\ ($\pm$ 0.54)} & \makecell{\textbf{0.20 }\\ ($\pm$ 0.02)} & \makecell{\textbf{-4.08} \\ ($\pm$ 0.09)} & \makecell{1.00 \\ ($\pm$ 0.00)} \\
        & $\bm\Sigma_\theta^{\mathrm{mv}}$  & \makecell{7.23 \\ ($\pm$ 0.51)} & \makecell{0.49 \\ ($\pm$ 0.01)} & \makecell{7.68 \\ ($\pm$ 0.08)} & \makecell{0.34 \\ ($\pm$ 0.01)} & \makecell{-3.59 \\ ($\pm$ 0.02)} & \makecell{0.99 \\ ($\pm$ 0.00)} \\
        & $\bm\epsilon_t^\mathrm{ES}$  & \makecell{10.43 \\ ($\pm$ 0.11)} & \makecell{0.59 \\ ($\pm$ 0.07)} & \makecell{7.34 \\ ($\pm$ 1.51)} & \makecell{0.34 \\ ($\pm$ 0.08)} & \makecell{-3.53 \\ ($\pm$ 0.28)} & \makecell{\textbf{0.98 }\\ ($\pm$ 0.01)} \\
        \midrule
        \multirow{4}{*}{\textbf{T2M}}
        & $\bm\delta_\theta$ & 84.43  (2.10) & 0.77 & 103.21 & 0.40 & 1.05 & 0.97 \\
        & $\bm\Sigma_\theta^{\mathrm{diag}}$& 82.30 (0.80) &\textbf{ 0.71} & 96.48 & \textbf{0.35} & 0.97 & 0.84 \\
        & $\bm\Sigma_\theta^{\mathrm{mix}}$ & 86.38 (1.14) &\textbf{ 0.71} & \textbf{95.68} & \textbf{0.35} & 0.98 & 0.83\\
        & $\bm\Sigma_\theta^{\mathrm{mv}}$  & \textbf{76.38} (1.36) & 0.76 & 101.39 & 0.38 & \textbf{0.93} &\textbf{ 0.94} \\
        \bottomrule
    \end{tabular}
\end{table}

\begin{table}[ht]
    \centering
        \caption{Full depth estimation results. T}
    \label{tab:full_results_depth-regression}
    \begin{tabular}{|l|l|cccc|}
    \toprule
        Experiment & Model & AbsRel $\downarrow$ & $\delta1 \uparrow$ & CRPS $\downarrow$ & ES $\downarrow$ \\
        \midrule
        \multirow{4}{*}{\textbf{NYUv2}}
        & $\bm\delta_\theta$ & 5.96 & 95.95 & 11.32 & 8324.67  \\
        & $\bm\Sigma_\theta^{\mathrm{diag}}$ & 5.90 & 95.94 & 11.32 & 8341.45 \\
        & $\bm\Sigma_\theta^{\mathrm{mix, repl}}$ & 5.89 & 95.99 & 11.35 & 8397.20  \\
        & $\bm\Sigma_\theta^{\mathrm{mv}}$ &  \textbf{5.67} & \textbf{96.15} & \textbf{11.02} & \textbf{8238.10} \\
        \midrule
        \multirow{4}{*}{\textbf{KITTI}}
        & $\bm\delta_\theta$ & 10.32 & 90.11 & 142.92 & 655.03 \\
        & $\bm\Sigma_\theta^{\mathrm{diag}}$ & 10.07 & 90.81 & 138.24 & \textbf{638.46} \\
        & $\bm\Sigma_\theta^{\mathrm{mix, repl}}$ & \textbf{9.89} & \textbf{91.09} & \textbf{137.60} & 640.34  \\
        & $\bm\Sigma_\theta^{\mathrm{mv}}$ & 10.14 & 90.95 & 142.28 & 649.21  \\
        \midrule
        \multirow{4}{*}{\textbf{ETH3D}}
        & $\bm\delta_\theta$ & 6.82 & 95.62 & 29.23 & 906.37 \\
        & $\bm\Sigma_\theta^{\mathrm{diag}}$ & 6.57 & 95.59 & \textbf{27.89} & \textbf{886.84}  \\
        & $\bm\Sigma_\theta^{\mathrm{mix}}$ & 6.72 & 95.66 & 28.66 & 887.65 \\
        & $\bm\Sigma_\theta^{\mathrm{mv}}$ & \textbf{6.47} & \textbf{95.99} & 29.43 & 916.28  \\
        \midrule
        \multirow{4}{*}{\textbf{ScanNet}}
        & $\bm\delta_\theta$ & 7.10 & \textbf{94.67} & 9.19 & 74.28  \\
        & $\bm\Sigma_\theta^{\mathrm{diag}}$ & 6.84 & \textbf{94.67} & 8.86 & \textbf{72.88}  \\
        & $\bm\Sigma_\theta^{\mathrm{mix}}$ & 6.96 & 94.34 & 8.99 & 73.57  \\
        & $\bm\Sigma_\theta^{\mathrm{mv}}$ & \textbf{6.79} & 94.51 & \textbf{8.85} & 73.41  \\
        \midrule
        \multirow{4}{*}{\textbf{DIODE}}
        & $\bm\delta_\theta$ & 30.60 & 77.04 & 191.39 & 2328.79  \\
        & $\bm\Sigma_\theta^{\mathrm{diag}}$ & 31.52 & \textbf{76.41} & 196.08 & 2349.49  \\
        & $\bm\Sigma_\theta^{\mathrm{mix,}}$ & 31.09 & 77.01 & 190.04 & 2302.41 \\
        & $\bm\Sigma_\theta^{\mathrm{mv}}$ & \textbf{29.82} & 77.78 & \textbf{186.63} & \textbf{2284.49}  \\
        \bottomrule
    \end{tabular}
\end{table}

\begin{figure}[ht]
\begin{subfigure}{\linewidth}
        \centering
    \includegraphics[width=0.9\linewidth]{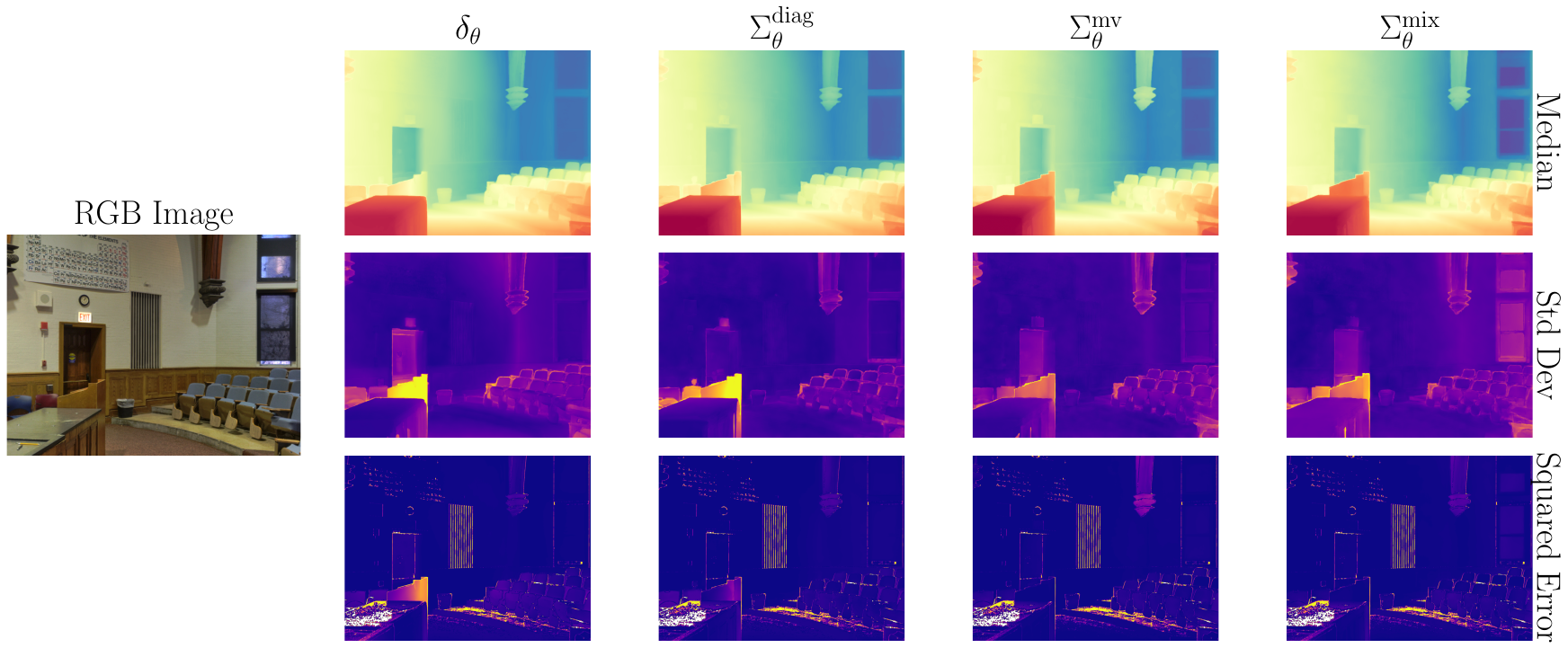}
\vspace{1cm}
\end{subfigure}
\begin{subfigure}{\linewidth}
        \centering
    \includegraphics[width=0.9\linewidth]{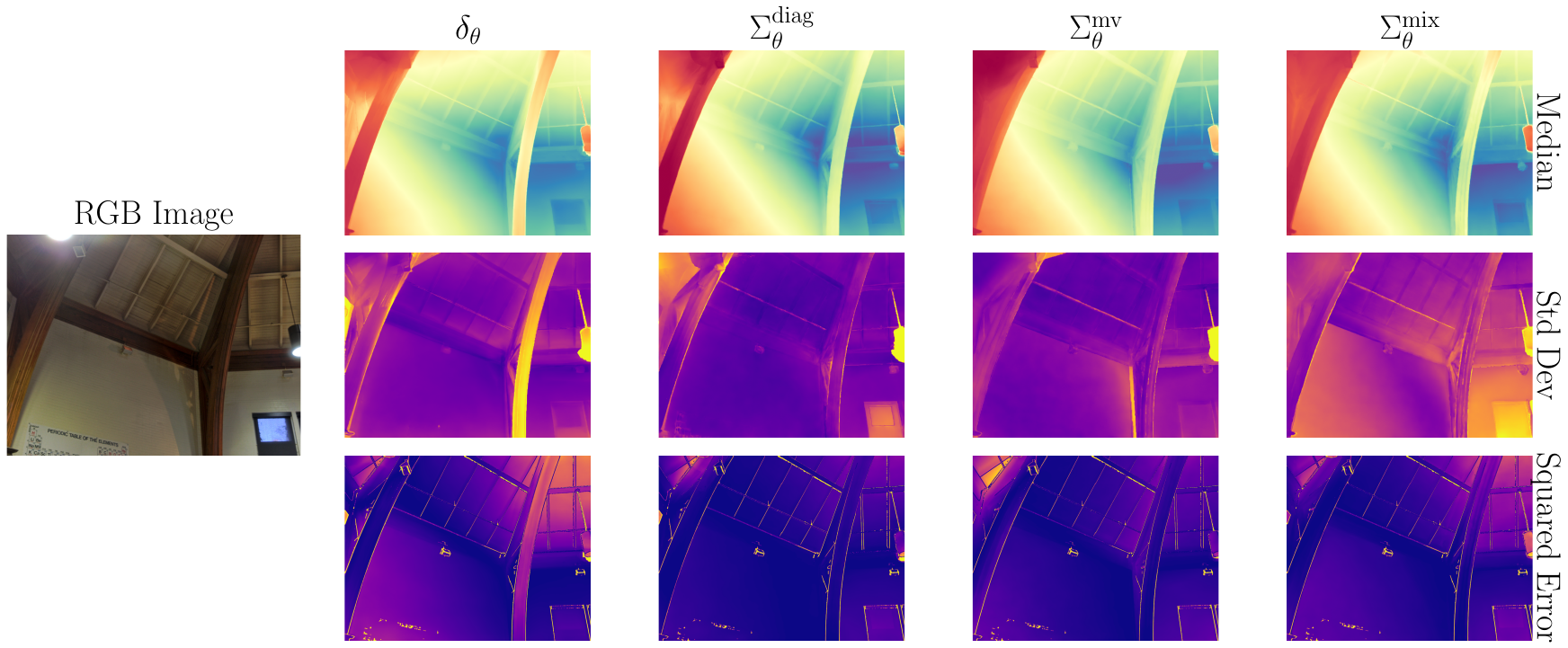}
\vspace{1cm}
\end{subfigure}
\begin{subfigure}{\linewidth}
        \centering
    \includegraphics[width=0.9\linewidth]{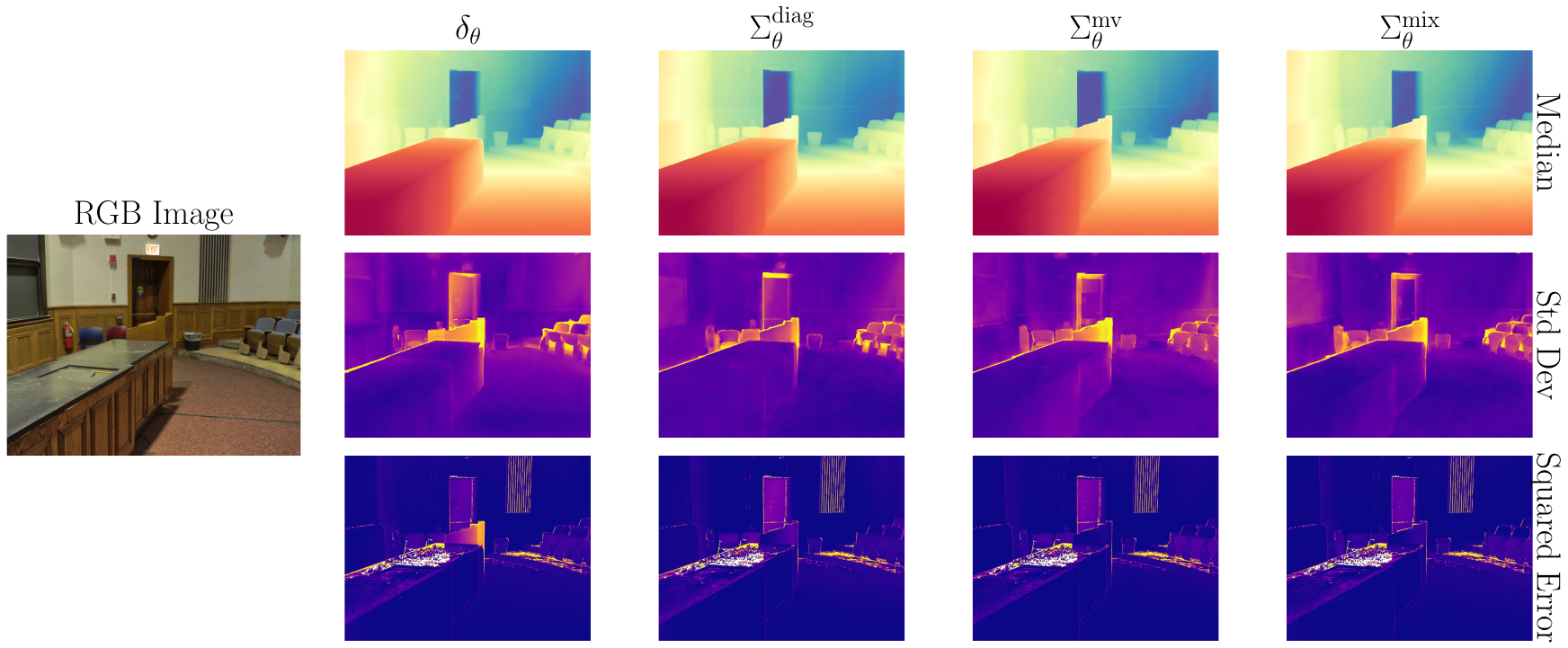}
\end{subfigure}
    \caption{Samples from the Diode dataset \citep{vasiljevic2019diode}.}
    \label{fig:visualization_diode}
\end{figure}

\begin{figure}[ht]
\begin{subfigure}{\linewidth}
        \centering
    \includegraphics[width=\linewidth]{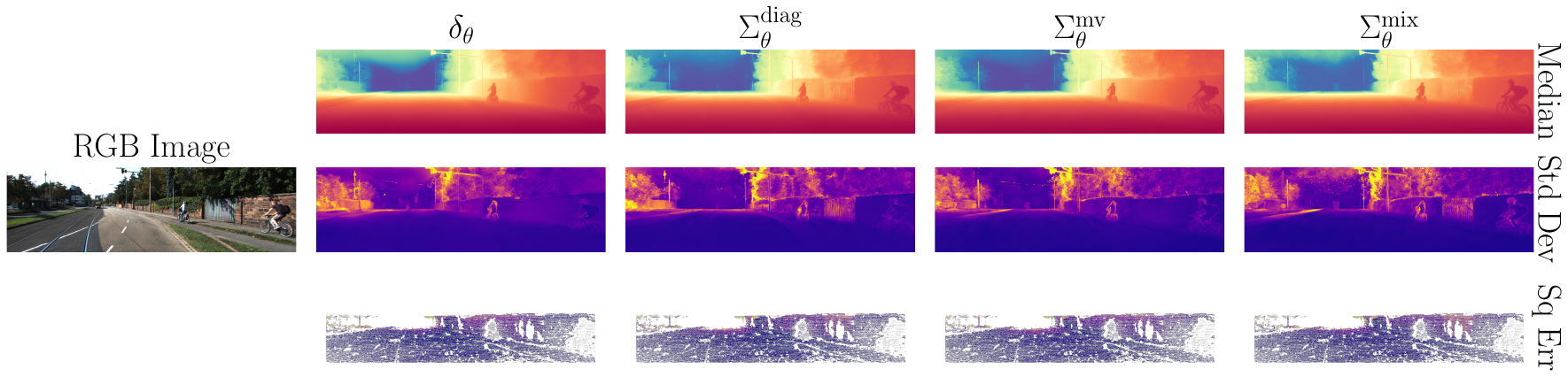}
\vspace{1cm}
\end{subfigure}
\begin{subfigure}{\linewidth}
        \centering
    \includegraphics[width=\linewidth]{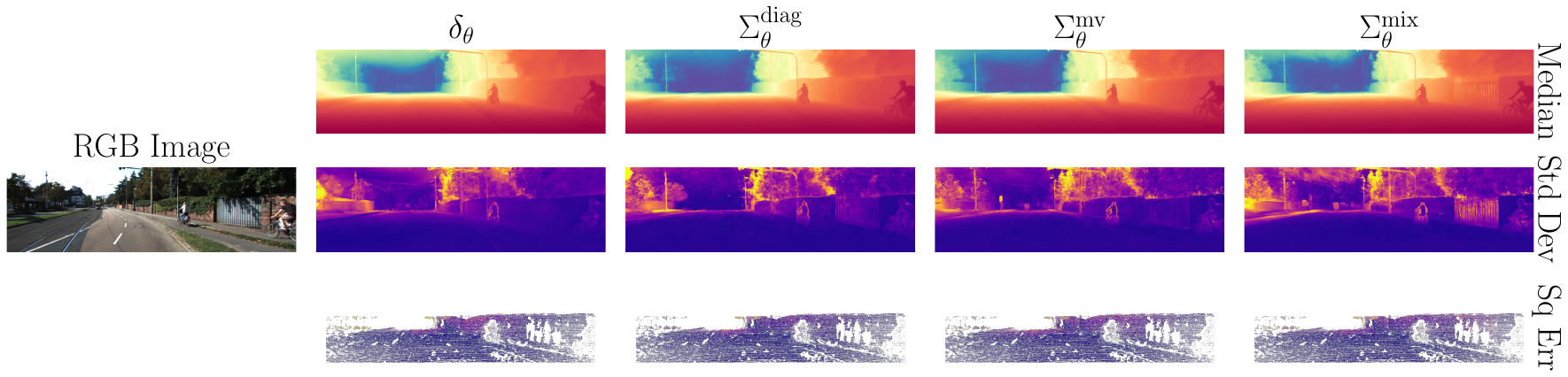}
\vspace{1cm}
\end{subfigure}
\begin{subfigure}{\linewidth}
        \centering
    \includegraphics[width=\linewidth]{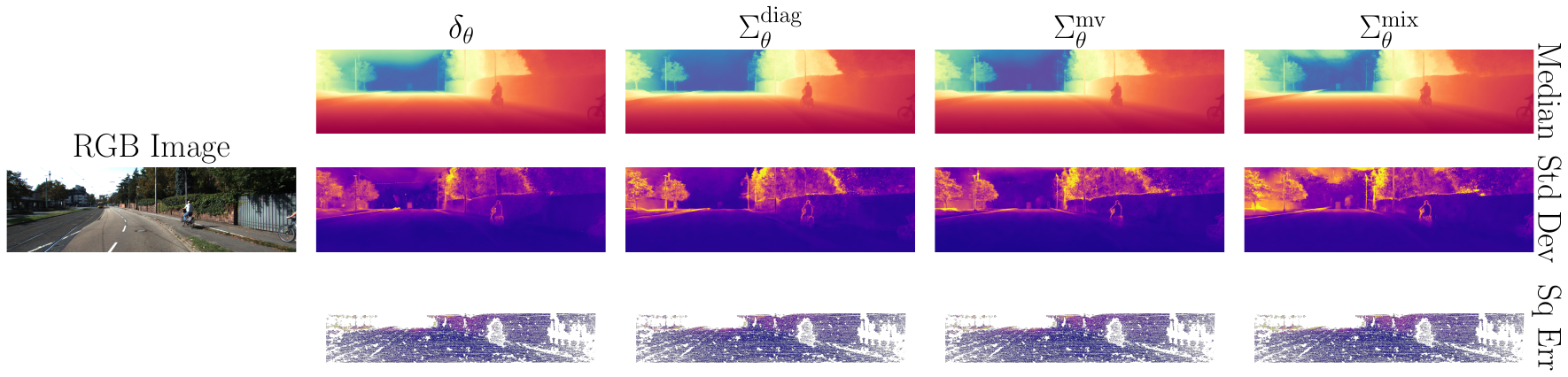}
\end{subfigure}
    \caption{Samples from the Kitti dataset \citep{geiger2012we}.}
    \label{fig:visualization_kitti}
\end{figure}

\begin{figure}[ht]
\begin{subfigure}{\linewidth}
        \centering
    \includegraphics[width=0.9\linewidth]{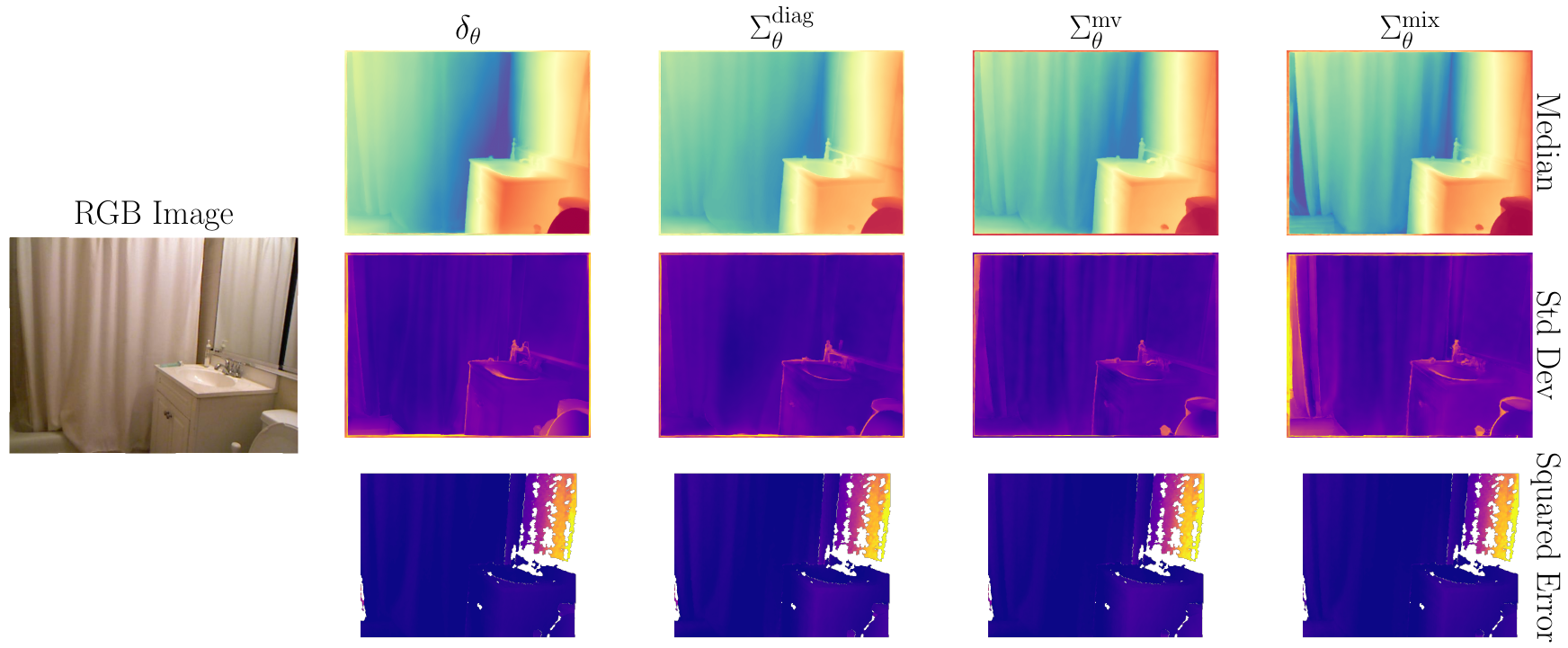}
\vspace{1cm}
\end{subfigure}
\begin{subfigure}{\linewidth}
        \centering
        \caption{$\bm\Sigma_\theta^\mathrm{diag}$}
    \includegraphics[width=0.9\linewidth]{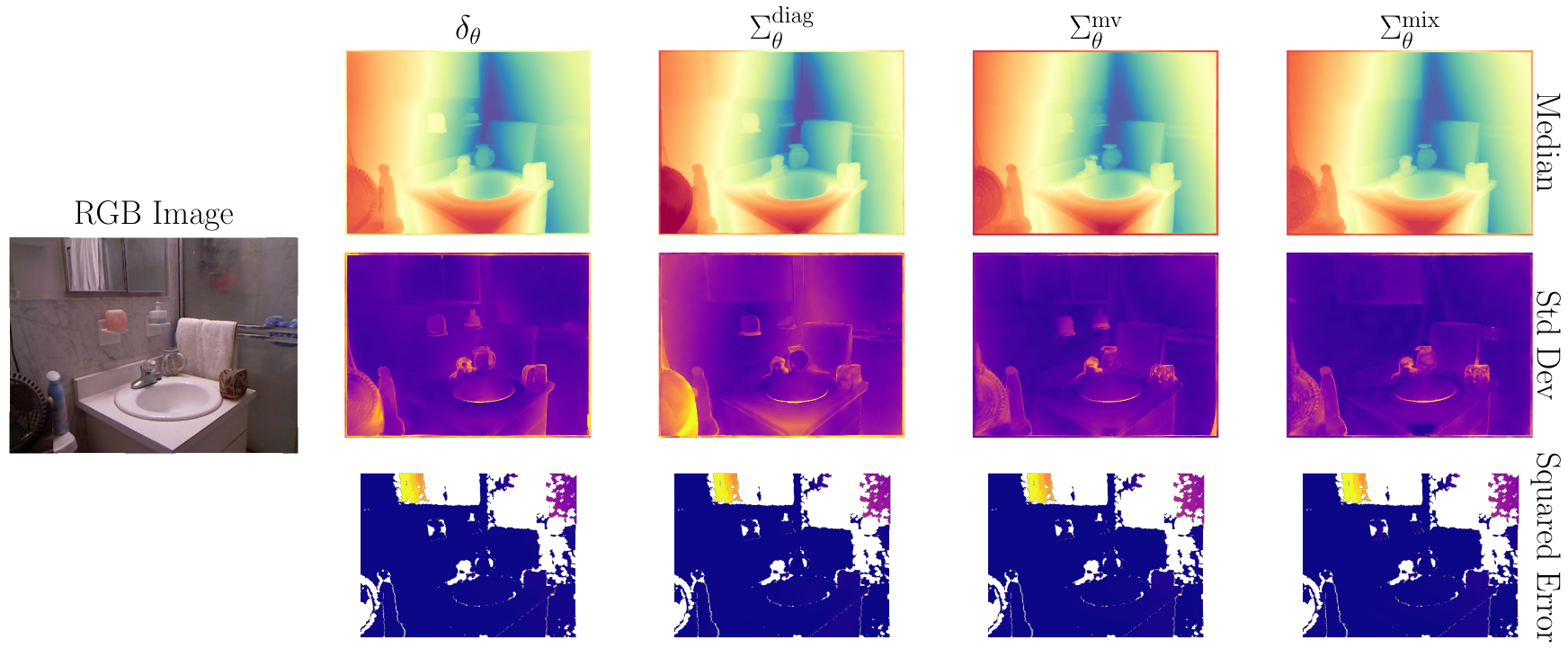}
\vspace{1cm}
\end{subfigure}
\begin{subfigure}{\linewidth}
        \centering
    \includegraphics[width=0.9\linewidth]{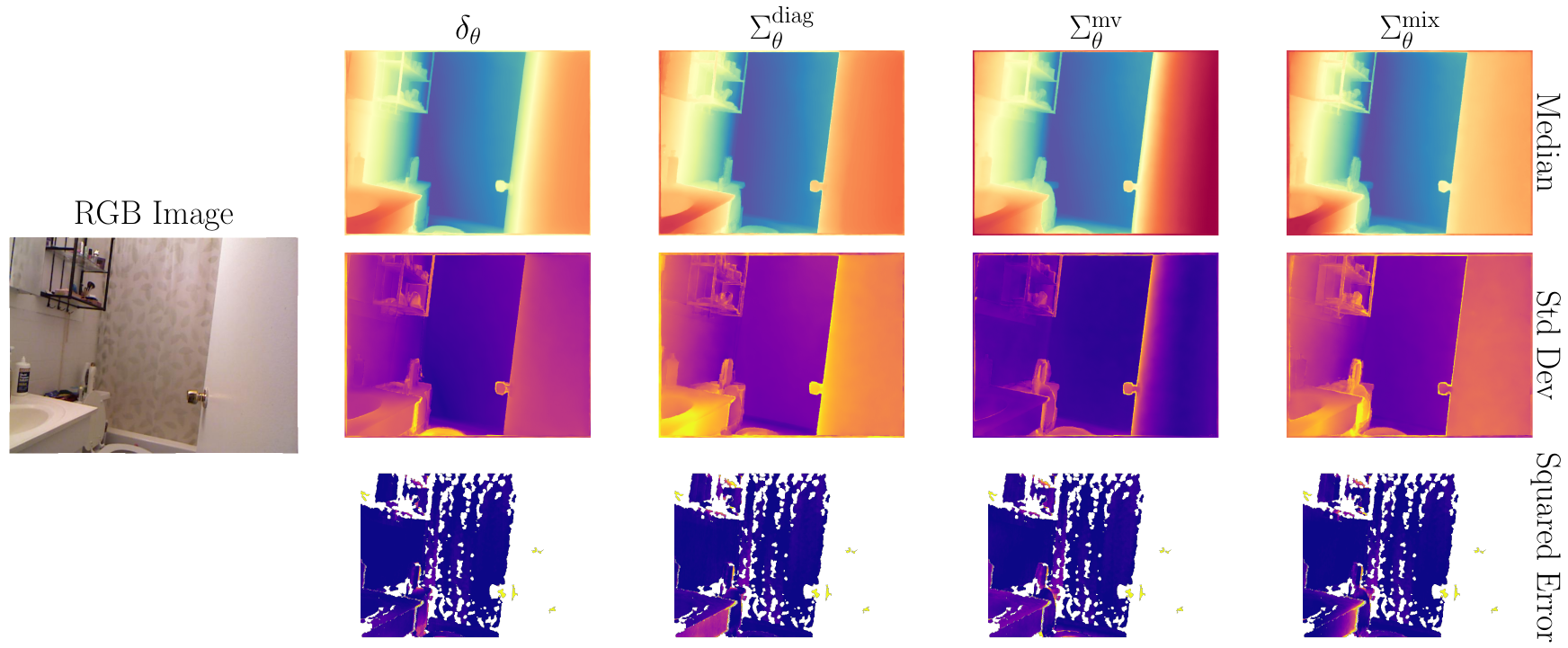}
\end{subfigure}
    \caption{Samples from the NYU dataset \citep{silberman2012nyu}.}
    \label{fig:visualization_nyu}
\end{figure}

\begin{figure}[ht]
\begin{subfigure}{\linewidth}
        \centering
    \includegraphics[width=0.9\linewidth]{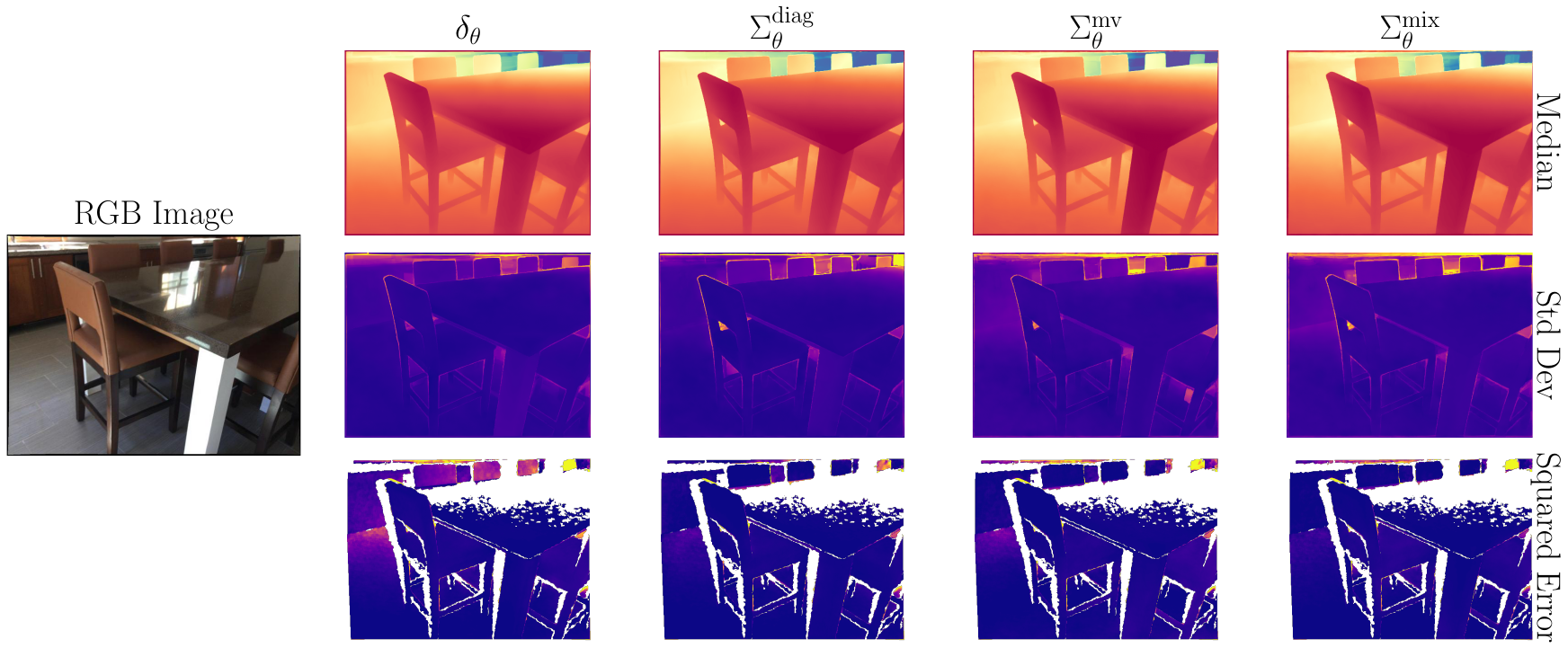}
\vspace{1cm}
\end{subfigure}
\begin{subfigure}{\linewidth}
        \centering
    \includegraphics[width=0.9\linewidth]{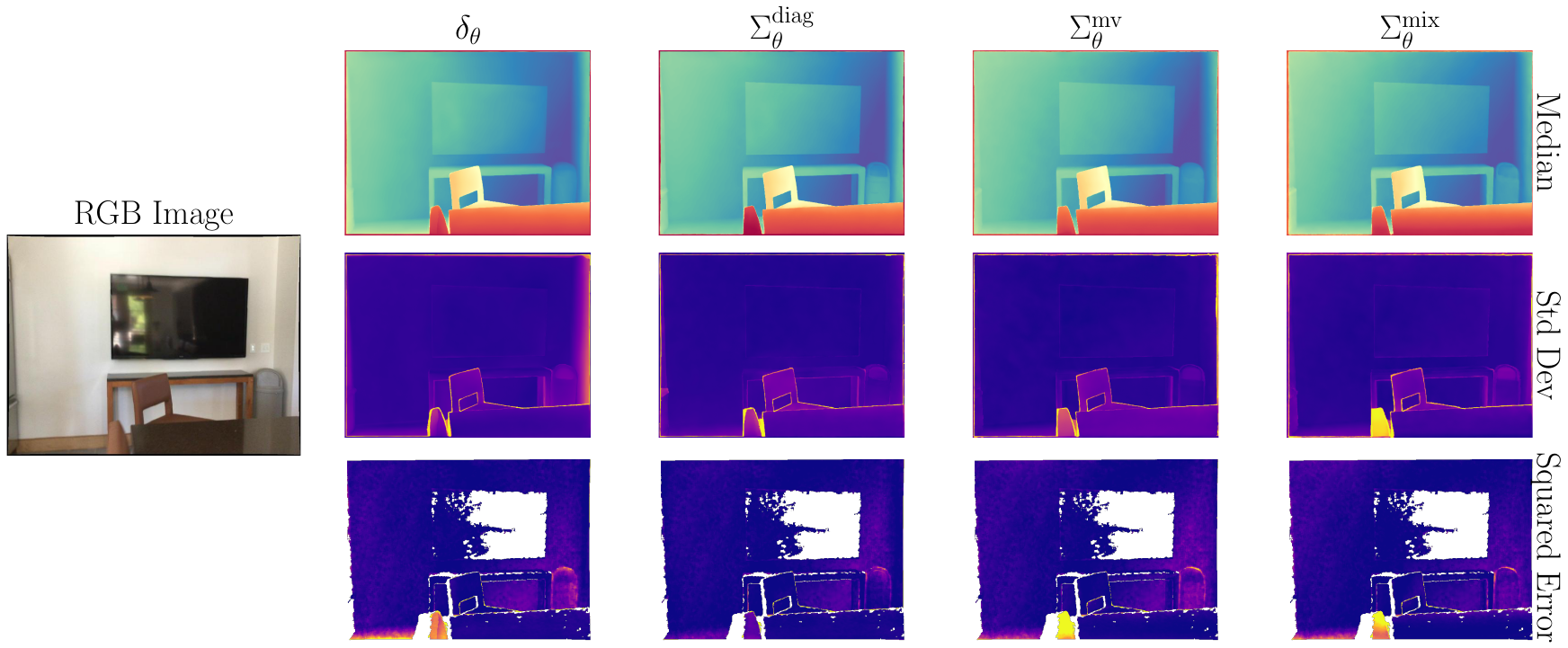}
\vspace{1cm}
\end{subfigure}
\begin{subfigure}{\linewidth}
        \centering
    \includegraphics[width=0.9\linewidth]{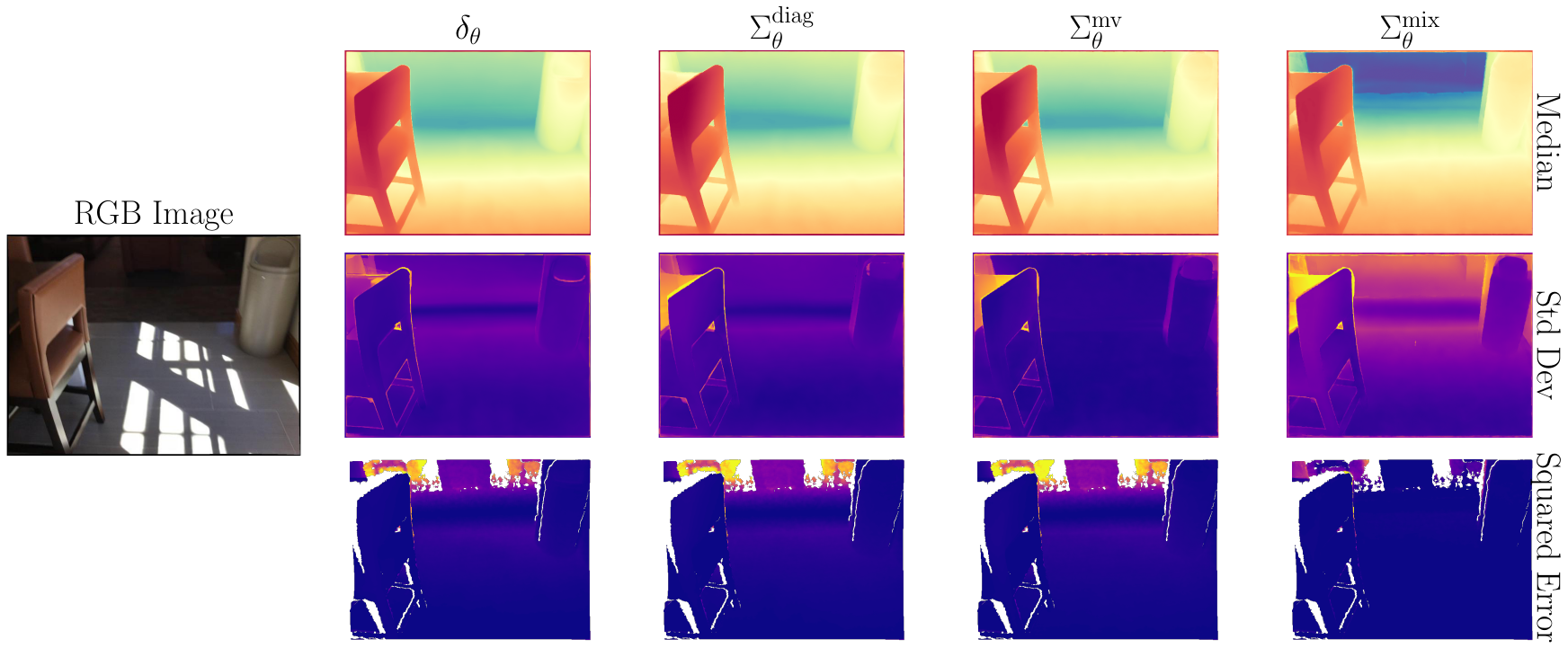}
\end{subfigure}
    \caption{Samples from the ScanNet dataset \citep{dai2017scannet}.}
    \label{fig:visualization_scannet}
\end{figure}

\begin{figure}[ht]
\begin{subfigure}{\linewidth}
        \centering
    \includegraphics[width=0.9\linewidth]{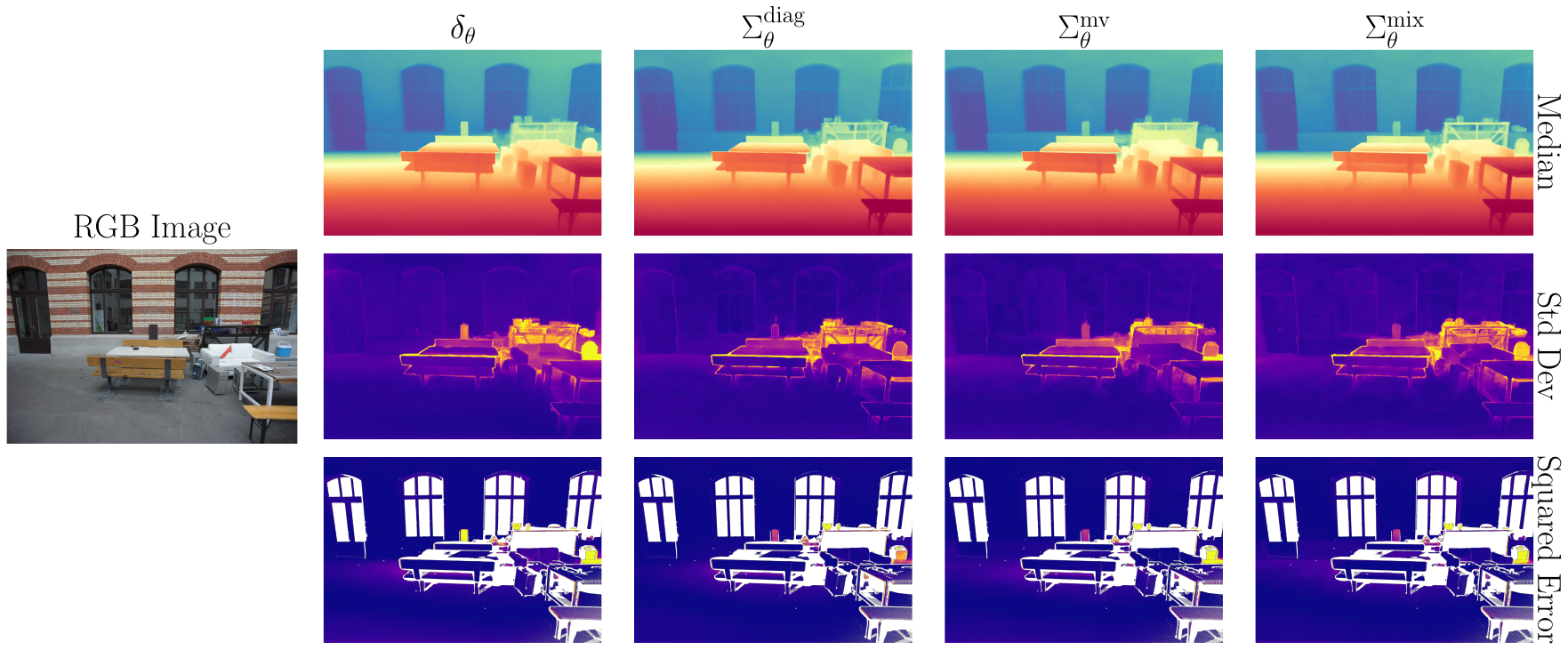}
\vspace{1cm}
\end{subfigure}
\begin{subfigure}{\linewidth}
        \centering
    \includegraphics[width=0.9\linewidth]{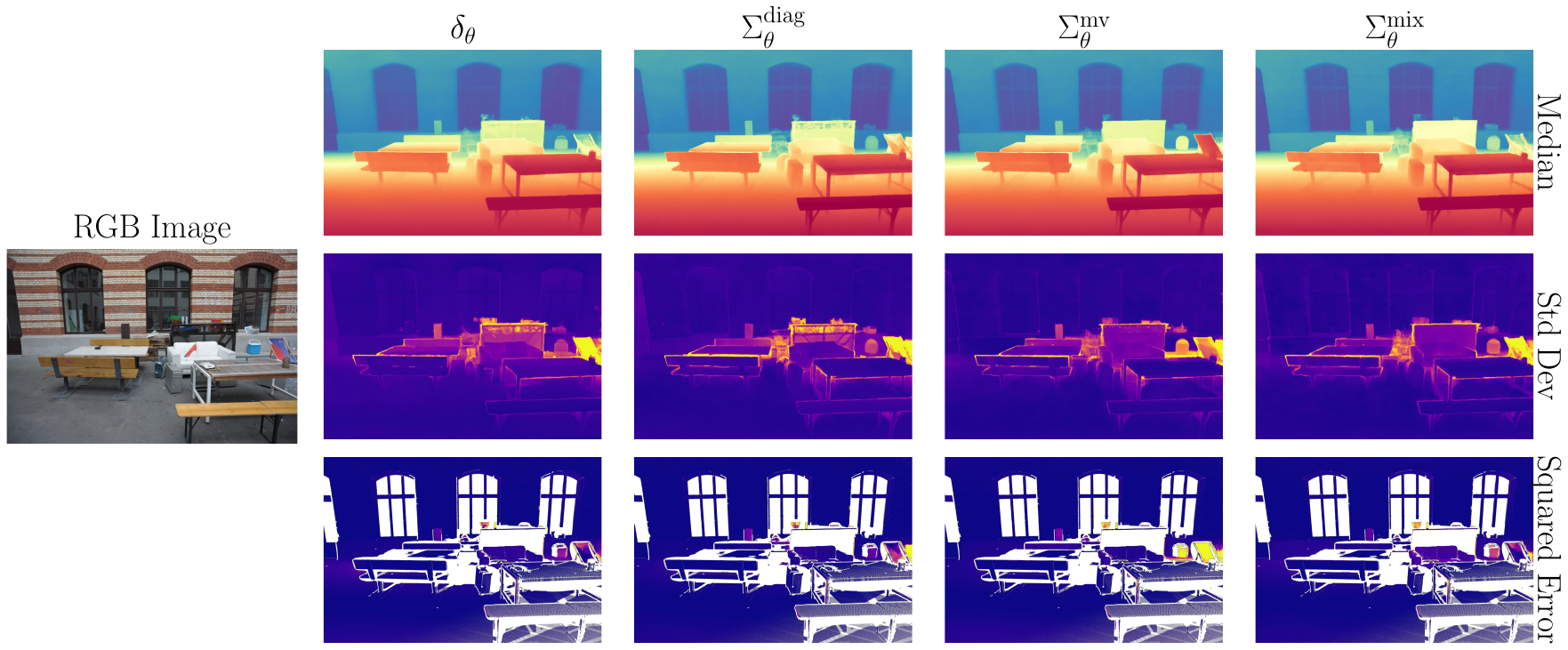}
\vspace{1cm}
\end{subfigure}
\begin{subfigure}{\linewidth}
        \centering
    \includegraphics[width=0.9\linewidth]{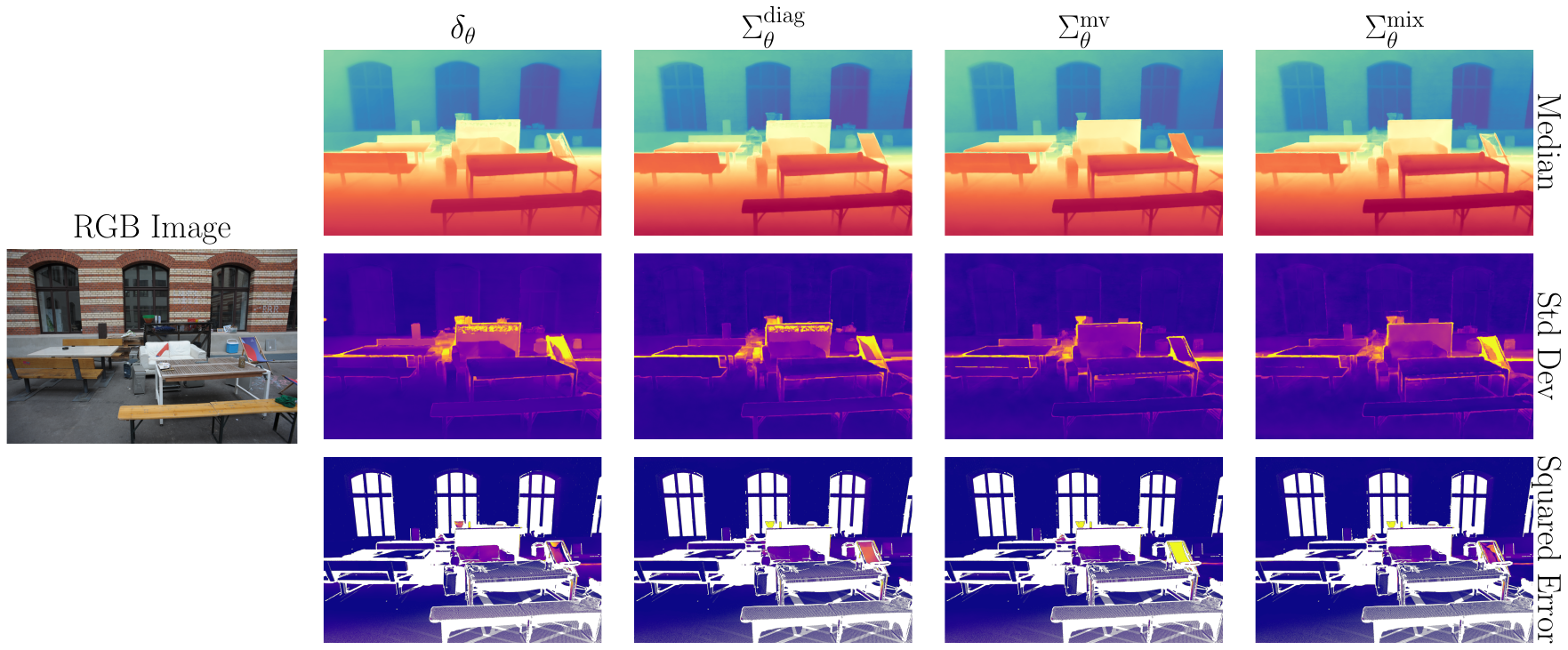}
\end{subfigure}
    \caption{Samples from the ETH3D dataset \citep{schops2017multi}.}
    \label{fig:visualization_eth3d}
\end{figure}

\FloatBarrier
\section{Improving marginal calibration}
\label{app:calibration}
Although our method generally outperforms the standard diffusion model, especially on PDEs, it tends to be over-conservative, leading to overly high coverage and thus suboptimal marginal calibration. A simple remedy is to rescale the covariance matrix of $p_\theta(\bm x_{t-1}|\bm x_t)$. Since the noise is always at least $\sigma_t^2 \mathbf{I}$, adding a learned covariance can produce an overly dispersed distribution. Multiplying the covariance by a scalar reduces this spread and improves calibration. While this deviates from the DDIM framework and warrants further theoretical analysis, we show empirically that it can enhance both calibration and predictive performance.

From Theorem~\ref{methodology:thm:closed_form}, we have
$$p_\theta(\bm x_{t-1}|\bm x_t)=\sum_{k=1} ^K\pi_{\theta,k}\mathcal{N}
        \left(\bm x_{t-1};
         \sqrt{\bar{\alpha}_{t-1}} \hat{\bm x}_0 + 
        \sqrt{ 1 - \bar{\alpha}_{t-1} - \sigma_t^2}\bm\mu_{\theta,k}^\epsilon(\bm x_t,t) ,
        \gamma_t^2 \bm\Sigma_{\theta,k}^\epsilon(\bm x_t,t) + \sigma_t^2 \mathbf{I}
        \right).$$
    
\begin{figure}[ht]
\begin{subfigure}{\linewidth}
        \centering
        \caption{Burgers'}
    \includegraphics[width=\linewidth]{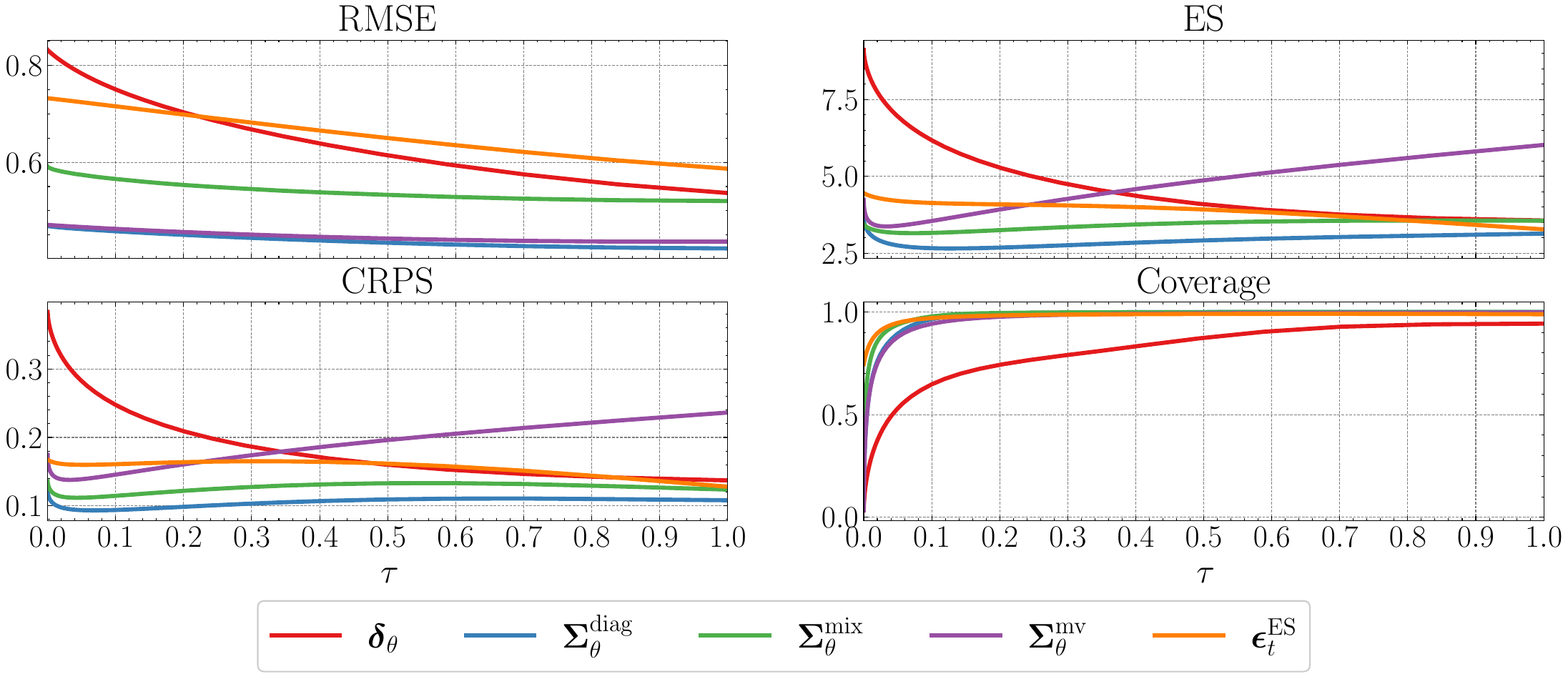}
\end{subfigure}
\begin{subfigure}{\linewidth}
        \centering
        \caption{Kuramoto--Sivashinsky}
    \includegraphics[width=\linewidth]{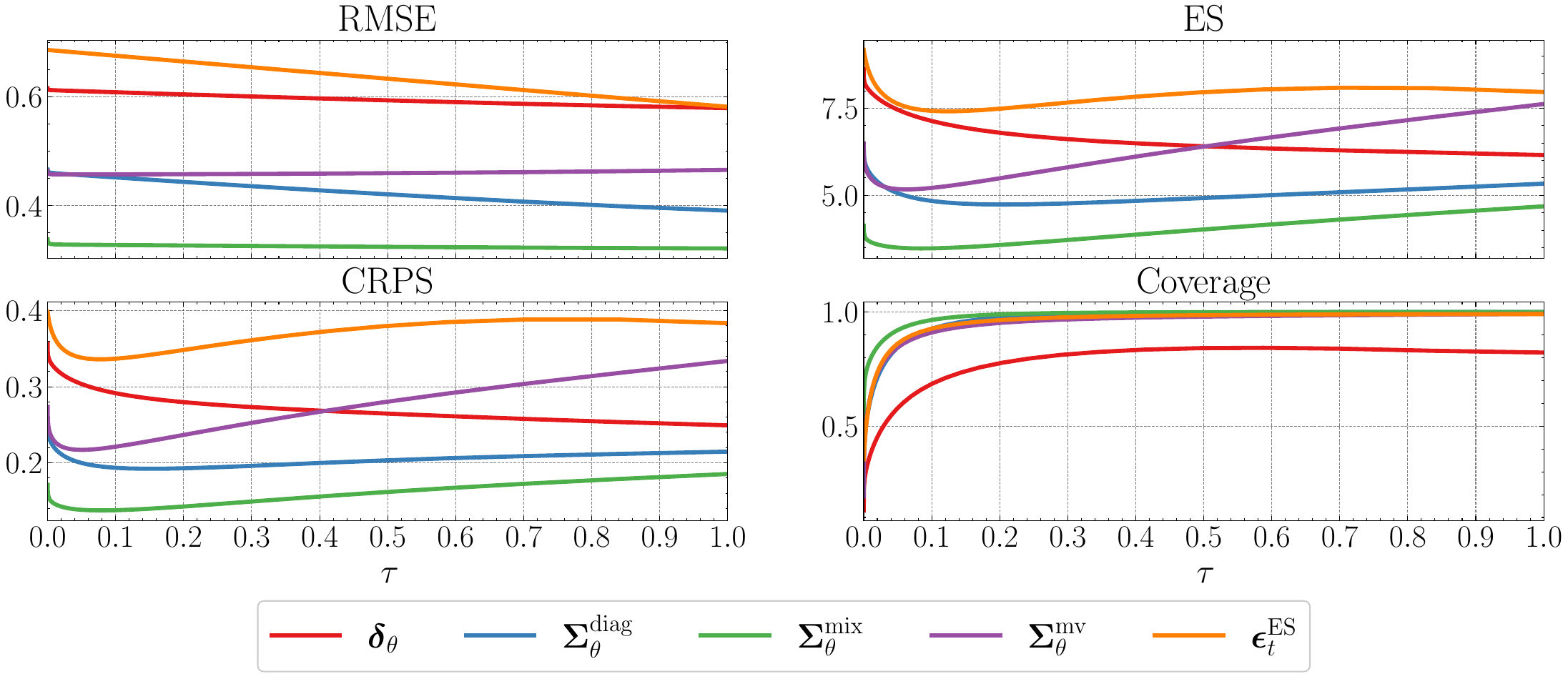}
\end{subfigure}
\caption{Performance of the different diffusion models in dependence on the scaling parameter $\tau$.}
\label{fig:noise_adjustment}
\end{figure}

We define the rescaled covariance as
$$\bm\Tilde{\Sigma}_{\theta,k}^\epsilon \coloneq \tau\left( \gamma_t^2 \bm\Sigma_{\theta,k}^\epsilon(\bm x_t,t) + \sigma_t^2 \mathbf{I} \right), \quad \tau \in (0,1].$$
Here, $\tau=1$ recovers the original covariance, while smaller $\tau$ reduces marginal variances. This adjustment can be applied to all diffusion methods, though for $\bm\delta_\theta$ and $\bm\epsilon_t^\mathrm{ES}$ only the diagonal term $\sigma_t^2 \mathbf{I}$ is rescaled. Figure~\ref{fig:noise_adjustment} illustrates the effect of $\tau$ on different metrics.

Results show that RMSE is typically minimized at $\tau=1$, but for our proposed models ($\bm\Sigma_\theta^\mathrm{diag}$, $\bm\Sigma_\theta^\mathrm{mix}$, $\bm\Sigma_\theta^\mathrm{mv}$), smaller values $\tau \approx 0.05$ improve coverage (close to $0.95$) and yield substantial gains in CRPS and energy score\textemdash{}up to a factor of two. For $\bm\delta_\theta$ performance is best at $\tau = 1$, and for $\bm \epsilon_t^\mathrm{ES}$, only CRPS and energy score show slight benefits for smaller $\tau$.

In summary, rescaling the covariance offers a simple way to improve marginal calibration and predictive performance, though not uniformly across metrics. Choosing $\tau$ optimally remains an open problem, and introducing this parameter moves the model away from the DDIM framework. A theoretical study of the rescaled diffusion process offers a promising direction for future work.

\FloatBarrier
\section{Capturing epistemic uncertainty}
\label{app:epistemic_uncertainty}
When considering predictive uncertainty, one typically considers two sources of uncertainty \citep{hullermeier2021aleatoric}: \emph{epistemic uncertainty} (EU) and \emph{aleatoric uncertainty} (AU). While aleatoric uncertainty describes the inherent randomness in the data-generating process, for example, due to measurement errors and is often referred to as \emph{irreducible} uncertainty, epistemic uncertainty arises from a lack of knowledge or information about the data-generating process and is also referred to as \emph{reducible} uncertainty. While aleatoric uncertainty is naturally represented by a probability distribution, epistemic uncertainty usually requires a higher-order representation, such as a second-order distribution (distribution of a distribution). Although some works propose specialized diffusion architectures for estimation of EU\citep{berry2024shedding, shu2024zeroshotuncertaintyquantificationusing}, standard diffusion models generally lack such access.

Our method addresses this by introducing an additional distribution $p_\theta^\epsilon(\epsilon_t \mid \bm x_t)$ over the noise variable $\epsilon_t$, which serves as a second-order distribution and enables direct estimation of epistemic uncertainty. Recall that for DDPMs, we have
\begin{equation}    
\label{eq:eu_p_theta}
p_\theta(\bm x_{t-1} \mid \bm x_t) = \mathcal{N}(\bm x_{t-1}; \bm\mu_\theta(\bm x_t, t), \bm\Sigma_\theta(\bm x_t ,t)),
\end{equation}
with $\bm\mu_\theta(\bm x_t, t) = \frac{1}{\sqrt{\alpha_t}}\left(\bm x_t - \frac{1-\alpha_t}{\sqrt{1- \bar{\alpha}_t}} \bm\epsilon_\theta(\bm x_t, t) \right)$ and a fixed diagonal $\bm\Sigma_\theta(\bm x_t, t)$.
Here, $\bm\epsilon_\theta(\bm x_t, t)$ is a deterministic denoiser. Iterative sampling then leads to the (approximate) predictive conditional distribution $p_\mathcal{Y}(\cdot \mid \bm c)$, from which aleatoric uncertainty can be assessed \citep{shu2024zeroshotuncertaintyquantificationusing}. Furthermore, by modeling $p_\theta^\epsilon(\epsilon_t \mid \bm x_t)$, our approach induces a distribution over $\bm\mu_\theta(\bm x_t, t)$ and hence a second-order distribution over the Gaussian mean in the transition kernel in \eqref{eq:eu_p_theta}. 

As an example, consider the diagonal covariance approximation, $\bm\Sigma_\theta^\mathrm{diag}$. In this case, EU can be estimated using the variance of the predictive mean:
\[
\underbrace{\mathbb{V}[\bE[X]]}_{\mathrm{EU}} = \bV[\bm\mu_\theta(\bm x_t, t)] \propto \mathrm{diag}(\sigma_\theta^2(\bm x_t, t)).
\]
Here, we focus on marginal epistemic uncertainty, i.e., one estimate per dimension $(i=1,\ldots, D)$, and average across timesteps $T$ to obtain a single value for EU. With this construction, both AU and EU are expressed in the same space as the prediction target, provided that the diffusion process and input/output domain coincide.

\begin{figure}[ht]
\begin{subfigure}{\linewidth}
        \centering
        \caption{Kuramoto--Sivashinsky}
    \includegraphics[width=\linewidth]{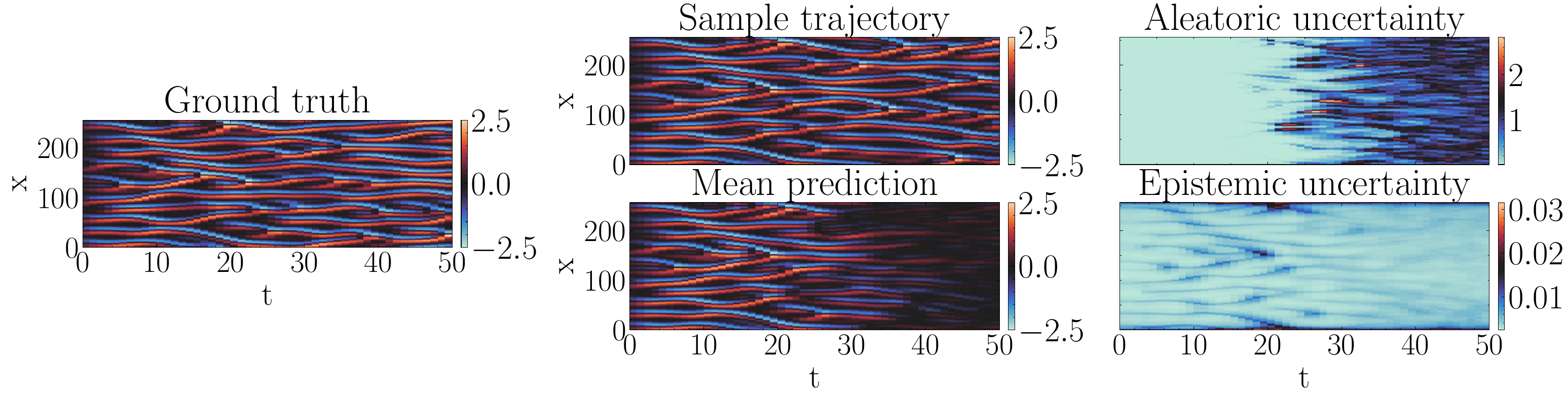}
\end{subfigure}
\begin{subfigure}{\linewidth}
        \centering
        \caption{Burgers'}
    \includegraphics[width=\linewidth]{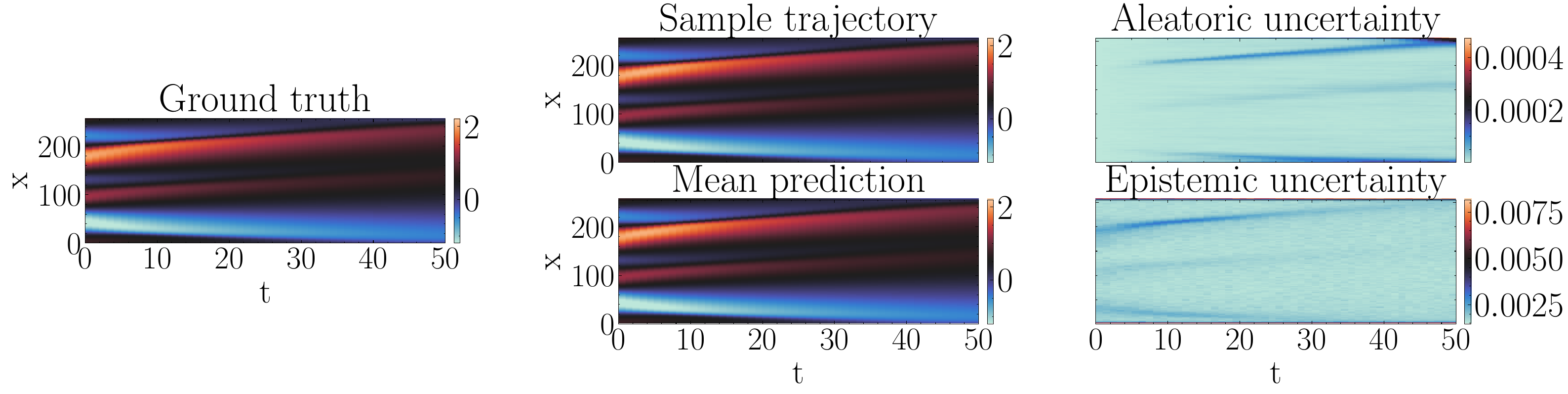}
\end{subfigure}
\begin{subfigure}{\linewidth}
        \centering
        \caption{T2M}
    \includegraphics[width=\linewidth]{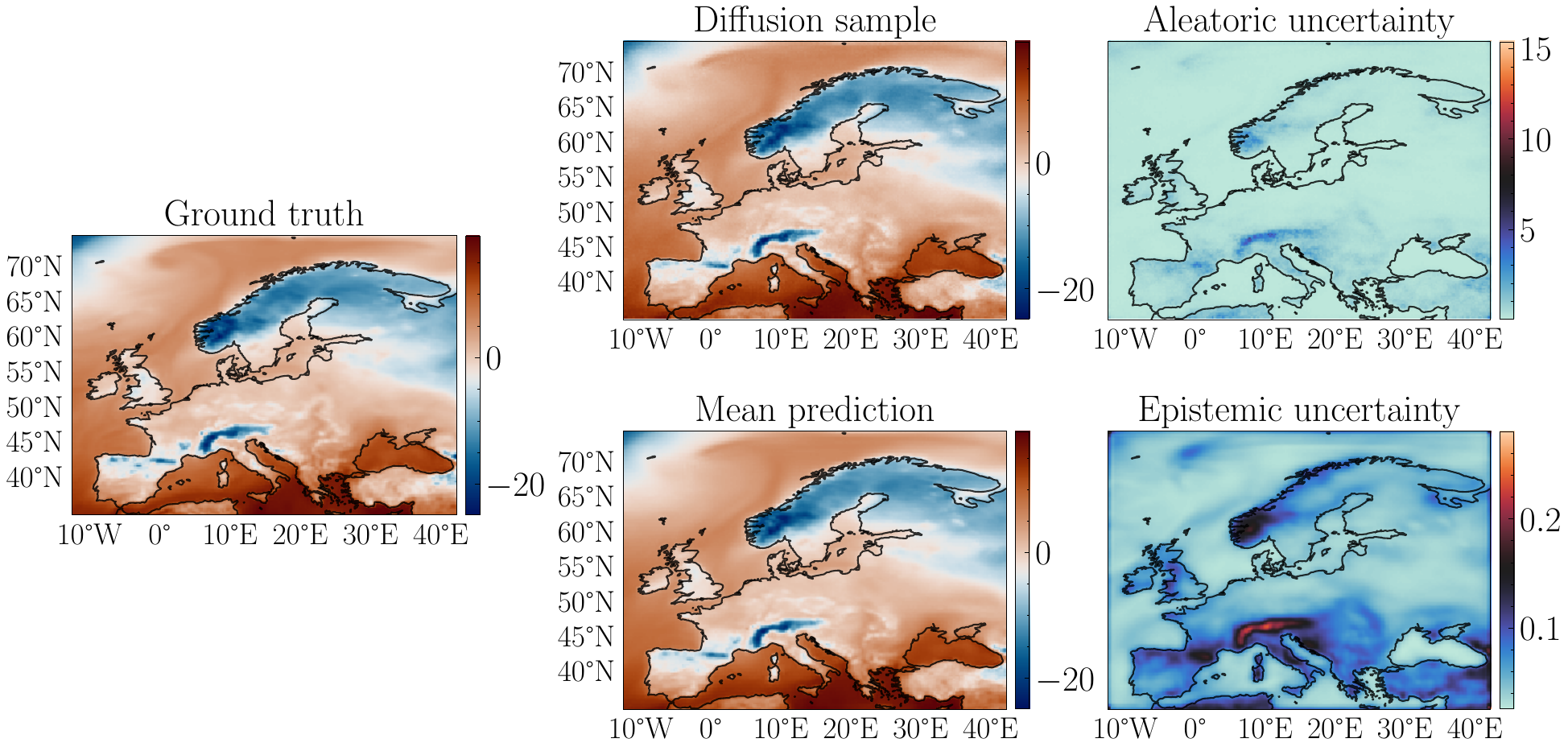}
\end{subfigure}
    \caption{Estimates of aleatoric and epistemic uncertainty for the $\bm\Sigma_\theta^\mathrm{diag}$ model for the three different autoregressive prediction tasks.}
    \label{fig:eu}
\end{figure}

Figure~\ref{fig:eu} shows selected sample predictions and their corresponding uncertainty estimates for the three autoregressive regression tasks. For the Kuramoto--Sivashinsky equation, AU grows over time\textemdash{}consistent with its chaotic dynamics\textemdash{}while EU correlates with different solution branches, especially at early timesteps.
For the Burgers' equation, both uncertainties remain small but vary with spatial location. 
Finally, for surface temperature prediction, AU is elevated in high-altitude regions due to fine-scale variability, while EU highlights these regions even more strongly, with large values over the Alps where predicted temperatures are low. This suggests the model is less certain in these regimes, which could be mitigated by using additional training data.

Overall, our model yields sensible EU estimates, but further theoretical and empirical study is needed—for example, exploring timestep aggregation strategies or alternative EU measures. These remain interesting directions for future work.

\section{Hyperparameters}
\label{app:hyperparameters}
Due to the high computational cost of the models, we are not able to provide an extensive statistical evaluation of all hyperparameters. However, since the proposed methods mainly affect the diffusion process, we can use the same model backbone for all methods and therefore still make them comparable. For some of the proposed methods, though, we introduce additional hyperparameters that need to be tuned. Therefore, we provide some minor hyperparameter evaluation for the prediction task of the Burgers' and the Kuramoto--Sivashinsky equations. Table~\ref{tab:hyperparams_mixednormal_components} shows the performance of the mixed normal model in dependence on the number of components $K$, Table~\ref{tab:hyperparams_sample_components} shows the performance of the sample-based model in dependence on the number of samples $M$, and Table~\ref{tab:hyperparams_mvnormal} shows the performance of the multivariate normal model in dependence on the covariance approximation and the kernel bandwidth $\gamma$. Each table shows the different metrics, as well as the average training time of one epoch. The training process is similar to the main experiment, but with more strict early stopping, to remain computationally feasible. All metrics are evaluated on an (unseen) validation set, and as a selection criterion, we choose the energy score.

\begin{table}[ht]
    \centering
        \caption{Effect of the number of components in the (univariate) mixture normal model on the performance on the validation set. For readability, the metrics RMSE, ES, and CRPS are scaled by the factor 100 (10) for the Burgers' (KS) data.}
    \label{tab:hyperparams_mixednormal_components}
    \begin{tabular}{|l|l|cccccc|}
    \toprule
        Experiment & $K$ & $t_{epoch} [s]$ & RMSE$\downarrow$ & ES$\downarrow$ & CRPS$\downarrow$ & NLL$\downarrow$ & $\mathcal{C}_{0.95}$ \\
        \midrule
        \multirow{6}{*}{\textbf{Burgers'}}
        & 2 & 7.42 (0.18) &\textbf{ 0.0680} & \textbf{0.4790} &\textbf{ 0.0170} & -6.6114 & 0.9996 \\
        & 3 & 7.40 (0.16) & 0.1450 & 0.8400 & 0.0280 & -6.2805 & 0.9993 \\
       & 5 & 7.44 (0.16) & 0.0710 & 0.5030 & 0.0170 & \textbf{-6.6847} & \textbf{0.9992} \\
        & 10& 7.46 (0.17) & 0.1040 & 0.5960 & 0.0200 & -6.5018 & 0.9993 \\
        & 25 &  7.61 (0.14) & 0.0730 & 0.5230 & 0.0180 & -6.5867 & 0.9995 \\
        & 50  & 10.71 (0.22) & 0.0800 & 0.5210 & 0.0180 & -6.6183 & 0.9996 \\
        \midrule
       \multirow{6}{*}{\textbf{KS}}
       & 2 & 7.20 (0.18) & 0.0432 & 0.6043 & 0.0240 & -3.9073 & 0.9998 \\
        & 3 & \textbf{7.12 (0.15)} & 0.0545 & 0.7386 & 0.0301 & -3.6915 & 0.9997 \\
       & 5 & 7.39 (0.19) & 0.0640 & 0.8095 & 0.0328 & -3.6345 & \textbf{0.9995} \\
        & 10 & 7.50 (0.19) & 0.0446 & 0.6356 & 0.0264 & -3.7839 & 0.9997 \\
        & 25& 7.65 (0.16) & 0.0446 & 0.6557 & 0.0265 & -3.7612 & 0.9999 \\
        & 50 & 10.78 (0.18) & \textbf{0.0409} & \textbf{0.5790} & \textbf{0.0237} & \textbf{-3.9075} & 0.9998 \\
        \bottomrule
    \end{tabular}
\end{table}

\begin{table}[ht]
    \centering
        \caption{Effect of the number of samples drawn in the training of the sample-based model on the performance on the validation set. For readability, the metrics RMSE, ES, and CRPS are scaled by the factor 100 (10) for the Burgers' (KS) data.}
    \label{tab:hyperparams_sample_components}
    \begin{tabular}{|l|l|cccccc|}
    \toprule
        Experiment & $M$ & $t_{epoch} [s]$ & RMSE$\downarrow$ & ES$\downarrow$ & CRPS$\downarrow$ & NLL$\downarrow$ & $\mathcal{C}_{0.95}$ \\
        \midrule
        \multirow{4}{*}{\textbf{Burgers'}}
        & 3& \textbf{10.46 (0.25)} & 0.1020 & 0.4870 & \textbf{0.0190} & -6.6803 & 0.9912 \\
       & 5& 15.65 (0.28) & 0.1840 & 0.8440 & 0.0300 & -6.4734 & \textbf{0.9878 }\\
       &10 & 30.18 (0.45) & 0.1610 & 0.6250 & 0.0240 & -6.5091 & 0.9939 \\
       & 25 & 71.31 (0.31) & \textbf{0.0860} & \textbf{0.4560 }& \textbf{0.0190} & -6.6150 & 0.9957 \\
     \midrule
     \multirow{4}{*}{\textbf{KS}}
     & 3 & \textbf{10.47 (0.24)} & \textbf{0.0618} & \textbf{0.6824} & \textbf{0.0304} & \textbf{-3.6989} & \textbf{0.9700} \\
       & 5 & 15.68  (0.20) & 0.0668 & 0.7579 & 0.0347 & -3.5424 & 0.9785 \\
       &10 & 29.82 (0.44) & 0.2015 & 2.1613 & 0.0939 & -2.7003 & 0.9105 \\
       & 25 & 72.56 (0.33) & 0.0635 & 0.7364 & 0.0346 & -3.5335 & 0.9731\\
       \bottomrule
    \end{tabular}
\end{table}

\begin{table}[ht]
    \centering
        \caption{Effect of the covariance matrix approximation and the kernel bandwidth $\gamma$ on the performance on the validation set. The numerical values for $R$ denote the rank in the low-rank approximation, while C denotes the full Cholesky approximation. For the KS data, training with $\gamma = 5$ did not converge. For readability, the metrics RMSE, ES, and CRPS are scaled by the factor 100 (10) for the Burgers' (KS) data.}
    \label{tab:hyperparams_mvnormal}
    \begin{tabular}{|l|l|l|cccccc|}
\toprule
        Experiment  & $\gamma$ & $R$ & $t_{epoch} [s]$ & RMSE$\downarrow$ & ES$\downarrow$ & CRPS$\downarrow$ & NLL$\downarrow$ & $\mathcal{C}_{0.95}$ \\
        \midrule
                \multirow{18}{*}{\textbf{Burgers'}}
        &\multirow{6}{*}{5} & 1 & 7.39 (0.34) & \textbf{0.0480} & \textbf{0.7040} & \textbf{0.0270} & \textbf{-5.8702} & 0.9979 \\
        && 5 & 7.56 (0.11) & 0.1470 & 1.1290 & 0.0480 & -5.4215 & 0.9991 \\
        && 10 & 7.64 (0.20) & 0.2320 & 1.5920 & 0.0630 & -5.3654 & 0.9961 \\
        && 25 & 7.57 (0.14) & 0.3130 & 1.9480 & 0.0700 & -5.3428 & 0.9967 \\
        && 50 & 7.28 (0.37) & 0.0790 & 0.9950 & 0.0390 & -5.6229 & 0.9995 \\
        && C & 9.10 (0.22) & 0.1650 & 5.0860 & 0.1710 & -4.9388 & 0.9937 \\
        \cline{2-9}
        &\multirow{6}{*}{10} & 1 & 7.57 (0.45) & 0.0600 & 0.8280 & 0.0330 & -5.7380 & 0.9977 \\
        && 5 & 7.76 (0.25) & 0.0600 & 0.9130 & 0.0420 & -5.4361 & 0.9996 \\
        && 10 & 7.66 (0.24) & 0.6410 & 3.4170 & 0.1230 & -4.7906 & 0.9938 \\
        && 25 & 7.92 (0.27) & 0.0630 & 0.9490 & 0.0370 & -5.6301 & 0.9997 \\
        && 50 & 7.17 (0.40) & 0.0790 & 0.9790 & 0.0380 & -5.6346 & 0.9996 \\
        && C & 9.02 (0.17) & 0.1790 & 2.9850 & 0.0810 & -5.3116 & \textbf{0.9859} \\
        \cline{2-9}
        &\multirow{6}{*}{25} & 1 & 7.62 (0.18) & 0.0970 & 0.9870 & 0.0410 & -5.5773 & 0.9952 \\
        && 5 & \textbf{7.06 (0.19)} & 0.0990 & 1.0520 & 0.0470 & -5.3813 & 0.9985 \\
        && 10 & \textbf{7.06 (0.09)} & 0.0650 & 0.9450 & 0.0390 & -5.5638 & 0.9997 \\
        && 25 & 7.88 (0.71) & 0.1040 & 1.1440 & 0.0450 & -5.5535 & 0.9993 \\
        && 50 & 7.85 (0.62) & 0.1160 & 1.1010 & 0.0430 & -5.5743 & 0.9990 \\
        && C & 8.95 (0.08) & 0.2200 & 7.7430 & 0.2500 & -4.5096 & 0.9955 \\
        \midrule
        \multirow{12}{*}{\textbf{KS}}
        &\multirow{6}{*}{10} & 1 & 7.51 (0.33) & 0.0505 & \textbf{0.7791} & \textbf{0.0340} & \textbf{-3.5624} & 0.9921 \\
        && 5 & 7.54 (0.32) & 0.6869 & 7.5898 & 0.3386 & -1.4127 & \textbf{0.9719} \\
        && 10 & 7.26 (0.29) & 0.0620 & 0.9294 & 0.0405 & -3.2973 & 0.9987 \\
        && 25 & 7.18 (0.16) & 0.0549 & 0.8870 & 0.0374 & -3.3786 & 0.9991 \\
        && 50 & 7.15 (0.12) & \textbf{0.0493} & 0.8546 & 0.0365 & -3.3813 & 0.9993 \\
        && C & 9.01 (0.10) & 0.1147 & 4.3415 & 0.1133 & -2.9118 & 0.9797 \\
        \cline{2-9}
        &\multirow{6}{*}{25} & 1 & \textbf{7.13 (0.12)} & 0.2035 & 2.3734 & 0.1039 & -2.5079 & 0.9851 \\
        && 5 & 7.38 (0.34) & 0.0578 & 0.8802 & 0.0412 & -3.1876 & 0.9990 \\
        && 10 & 7.53 (0.34) & 0.0612 & 0.9406 & 0.0421 & -3.2333 & 0.9989 \\
        && 25 & 7.54 (0.32) & 0.0549 & 0.8918 & 0.0395 & -3.2972 & 0.9991 \\
        && 50 & 7.54 (0.32) & 0.0768 & 1.0469 & 0.0467 & -3.1729 & 0.9978 \\
        && C& 9.07 (0.11) & 0.1773 & 7.8864 & 0.2815 & -1.9158 & 0.9958 \\
    \bottomrule
    \end{tabular}
\end{table}

In a second stage, given the previous optimal hyperparameters, we tested for each model, whether performance would improve when changing the parameter $\beta_T$ of the noise scheduler. We tested two different values, based on recent studies; $\beta_T = 0.35$ \citep{hanCARDClassificationRegression2022} and $\beta_T=0.2$ \citep{kohl_benchmarking_2024}. Each model was run for two different seeds, the averaged results can be found in Table~\ref{tab:hyperparams_beta}. As for the weather forecasting task hyperparameter optimization is computationally too demanding, we choose values based on the previous results, selecting values based on a trade-off between performance and computational complexity.

\begin{table}[ht]
    \centering
        \caption{ For readability, the metrics RMSE, ES, and CRPS are scaled by the factor 100 (10) for the Burgers' (KS) data.}
    \label{tab:hyperparams_beta}
    \begin{tabular}{|l|l|l|ccccc|}
\toprule
        Experiment  & Model & $\beta_T$ & RMSE$\downarrow$ & ES$\downarrow$ & CRPS$\downarrow$ & NLL$\downarrow$ & $\mathcal{C}_{0.95}$ \\
        \midrule
        \multirow{10}{*}{\textbf{Burgers'}}
        &\multirow{2}{*}{$\bm\delta_\theta$}
        & 0.2 & 0.1856 & 0.9785 & 0.0371 & -6.2573 & 0.8240 \\
        && \textbf{0.35} & 0.0640 & 0.4133 & 0.0155 & -7.0553 & 0.9131 \\
        \cline{2-8}
        &\multirow{2}{*}{$\bm\Sigma_\theta^{\mathrm{diag}}$}
        & \textbf{0.2}  & 0.1098 & 0.7814 & 0.0314 & -6.1350 & 0.9944 \\
        && 0.35  & 0.1688 & 1.1605 & 0.0452 & -5.7837 & 0.9906 \\
        \cline{2-8}
        &\multirow{2}{*}{$\bm\Sigma_\theta^{\mathrm{mix}}$}
        & \textbf{0.2 } & 0.0581 & 0.4219 & 0.0146 & -6.8363 & 0.9994 \\
        && 0.35 & 0.0784 & 0.5105 & 0.0172 & -6.6333 & 0.9995 \\
        \cline{2-8}
        &\multirow{2}{*}{$\bm\Sigma_\theta^{\mathrm{mv}}$}
        & \textbf{0.2}& 0.0524 & 0.6523 & 0.0255 & -6.0578 & 0.9970 \\
        && 0.35 & 0.0806 & 0.8573 & 0.0342 & -5.7823 & 0.9967 \\
        \cline{2-8}
        &\multirow{2}{*}{$\bm\epsilon_t^\mathrm{ES}$}
        & \textbf{0.2} & 0.0637 & 0.3675 & 0.0148 & -6.8602 & 0.9903 \\
        && 0.35& 0.0806 & 0.4386 & 0.0176 & -6.6754 & 0.9922 \\   
        \cline{1-8}
        
        \multirow{10}{*}{\textbf{KS}}
        &\multirow{2}{*}{$\bm\delta_\theta$}
        & 0.2 & 0.1896 & 2.1152 & 0.0976 & -3.0426 & 0.8237 \\
        && \textbf{0.35}& 0.0636 & 0.6545 & 0.0263 & -3.9476 & 0.8643 \\
        \cline{2-8}
        &\multirow{2}{*}{$\bm\Sigma_\theta^{\mathrm{diag}}$}
        & \textbf{0.2}& 0.0549 & 0.7443 & 0.0324 & -3.5836 & 0.9985 \\
        && 0.35 & 0.0718 & 0.9108 & 0.0390 & -3.4225 & 0.9982 \\
        \cline{2-8}
        &\multirow{2}{*}{$\bm\Sigma_\theta^{\mathrm{mix}}$}
        & \textbf{0.2}& 0.0475 & 0.6448 & 0.0258 & -3.8172 & 0.9996 \\
        && 0.35  & 0.0556 & 0.7336 & 0.0300 & -3.6820 & 0.9995 \\
        \cline{2-8}
        &\multirow{2}{*}{$\bm\Sigma_\theta^{\mathrm{mv}}$}
        & 0.2 & 0.0987 & 1.1955 & 0.0515 & -3.3161 & 0.9903 \\
        && \textbf{0.35} & 0.0584 & 0.8510 & 0.0376 & -3.5145 & 0.9893 \\
        \cline{2-8}
        &\multirow{2}{*}{$\bm\epsilon_t^\mathrm{ES}$}
        & 0.2  & 0.1145 & 1.2534 & 0.0558 & -3.2355 & 0.9432 \\
        && \textbf{0.35} & 0.0814 & 0.9029 & 0.0410 & -3.4059 & 0.9620 \\     
    \bottomrule
    \end{tabular}
\end{table}

\clearpage
\subsection{Comparison of scoring rules}
In principle, our framework allows for arbitrary scoring rule loss functions for training. While \citet{bortoli2025distributional} show that certain choices of $k$ for the kernel score lead to loss functions that can recover the original diffusion loss in the limit, they also show empirical findings that scoring rules without this property can lead to similar results. To further motivate our use of the energy (and kernel) score, we show a brief comparison against a very common scoring rule, the log-score (or negative log-likelihood), which is defined as 
 \[
 S_\mathrm{log}(p, \bm y) \coloneq - \log p(\bm y),
 \]
 for a probability density $p$. The log-score is commonly used to train neural networks for parametric distributions such as predictive Gaussians \citep{374138, lakshminarayanan_simple_2017}. We compare $S_\mathrm{log}$ against the energy score $S_\mathrm{ES}$ for the two PDE prediction tasks and the $\bm\Sigma_\theta^{\mathrm{diag}}$ model, for which closed-form expressions are available for both loss functions. Table~\ref{tab:log_score_comparison} shows that the energy score leads to significantly better results across all metrics for both PDEs. The performance of the model trained with $S_\mathrm{log}$ is up to a factor of 20 times worse, highlighting that training using the energy score can be advantageous, which goes along recent literature \citep{Rasp.2018,10.1093/jrsssb/qkae108,alet2025skillfuljointprobabilisticweather}.
\begin{table}[ht]
    \centering
        \caption{Comparison of different scoring rules for the $\bm\Sigma_\theta^{\mathrm{diag}}$ model. The best model is highlighted in bold and the standard deviation across different model runs is given in brackets.}
    \label{tab:log_score_comparison}
    \begin{tabular}{|l|l|cccccc|}
    \toprule
        Experiment & $S$ & $t_{epoch} [s]$ & RMSE$\downarrow$ & ES$\downarrow$ & CRPS$\downarrow$ & NLL$\downarrow$ & $\mathcal{C}_{0.95}$ \\
        \midrule
        \multirow{3}{*}{\textbf{Burgers'}}
        & ES & \makecell{\textbf{6.81} \\ ($\pm$ 0.16)} & \makecell{\textbf{0.81} \\ ($\pm$ 0.27)} & \makecell{\textbf{3.67} \\ ($\pm$ 0.39)} & \makecell{\textbf{0.12} \\ ($\pm$ 0.02)} & \makecell{\textbf{-7.12} \\ ($\pm$ 0.09)} & \makecell{1.00 \\ ($\pm$ 0.00)} \\
        & log  & \makecell{7.25 \\ ($\pm$ 0.22)} & \makecell{17.31 \\ ($\pm$ 6.47)} & \makecell{65.88 \\ ($\pm$ 28.98)} & \makecell{1.32 \\ ($\pm$ 0.52)} & \makecell{-5.79 \\ ($\pm$ 0.28)} & \makecell{1.00 \\ ($\pm$ 0.00)} \\
        \midrule
        \multirow{3}{*}{\textbf{KS}}
        & ES & \makecell{7.54 \\ ($\pm$ 0.28)} & \makecell{\textbf{0.39} \\ ($\pm$ 0.06)} & \makecell{\textbf{5.23} \\ ($\pm$ 0.75)} & \makecell{\textbf{0.21} \\ ($\pm$ 0.03)} & \makecell{\textbf{-4.04} \\ ($\pm$ 0.14)} & \makecell{1.00 \\ ($\pm$ 0.00)} \\
        & log & \makecell{\textbf{7.47} \\ ($\pm$ 0.66)} & \makecell{2.11 \\ ($\pm$ 1.57)} & \makecell{23.19 \\ ($\pm$ 15.69)} & \makecell{0.50 \\ ($\pm$ 0.14)} & \makecell{-3.35 \\ ($\pm$ 0.23)} & \makecell{1.00 \\ ($\pm$ 0.00)} \\
        \bottomrule
    \end{tabular}
\end{table}

\end{document}

%% file: packages_and_commands.tex
\usepackage{amsmath}
\usepackage{amssymb}
\usepackage{amsthm}
\usepackage{bm}
\usepackage{subcaption}
\usepackage{hyperref}
\usepackage{xcolor}
\usepackage{array}
\usepackage{makecell}
\usepackage{longtable}
\usepackage{mathtools}
\usepackage{enumitem}
\usepackage{tcolorbox}
\usepackage{booktabs}
\usepackage{natbib}

\newcommand{\bE}{\mathbb{E}}
\newcommand{\Ree}{\mathbb{R}}
\newcommand{\bV}{\mathbb{V}}

\newcommand{\bP}{\mathbb{P}}
\newcommand{\bQ}{\mathbb{Q}}
\newcommand{\cP}{\mathcal{P}}

\newcommand{\cN}{\mathcal{N}}

\newtheorem{theorem}{Theorem}

%% file: preprint/DistrDiffusionDrawing.tex
\begingroup%
  \makeatletter%
  \providecommand\color[2][]{%
    \errmessage{(Inkscape) Color is used for the text in Inkscape, but the package 'color.sty' is not loaded}%
    \renewcommand\color[2][]{}%
  }%
  \providecommand\transparent[1]{%
    \errmessage{(Inkscape) Transparency is used (non-zero) for the text in Inkscape, but the package 'transparent.sty' is not loaded}%
    \renewcommand\transparent[1]{}%
  }%
  \providecommand\rotatebox[2]{#2}%
  \newcommand*\fsize{\dimexpr\f@size pt\relax}%
  \newcommand*\lineheight[1]{\fontsize{\fsize}{#1\fsize}\selectfont}%
  \ifx\svgwidth\undefined%
    \setlength{\unitlength}{492.60078267bp}%
    \ifx\svgscale\undefined%
      \relax%
    \else%
      \setlength{\unitlength}{\unitlength * \real{\svgscale}}%
    \fi%
  \else%
    \setlength{\unitlength}{\svgwidth}%
  \fi%
  \global\let\svgwidth\undefined%
  \global\let\svgscale\undefined%
  \makeatother%
  \begin{picture}(1,0.21105912)%
    \lineheight{1}%
    \setlength\tabcolsep{0pt}%
    \put(0,0){\includegraphics[width=\unitlength,page=1]{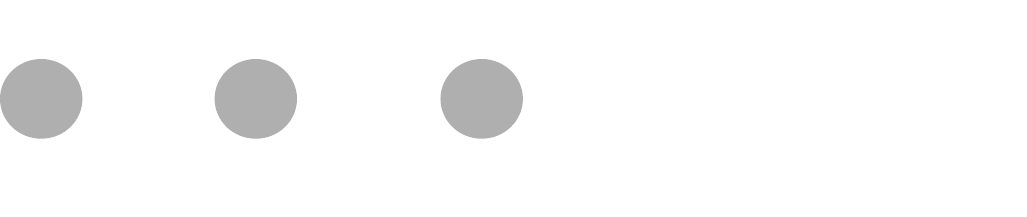}}%
    \put(0.46822194,0.10920759){\color[rgb]{0,0,0}\makebox(0,0)[t]{\lineheight{1.25}\smash{\begin{tabular}[t]{c}$x_0$\end{tabular}}}}%
    \put(0.03901807,0.10920759){\color[rgb]{0,0,0}\makebox(0,0)[t]{\lineheight{0}\smash{\begin{tabular}[t]{c}$x_T$\end{tabular}}}}%
    \put(0.2481312,0.10920759){\color[rgb]{0,0,0}\makebox(0,0)[t]{\lineheight{1.25}\smash{\begin{tabular}[t]{c}$x_t$\end{tabular}}}}%
    \put(0,0){\includegraphics[width=\unitlength,page=2]{DistrDiffusionDrawing.pdf}}%
    \put(0.32090563,0.16088144){\color[rgb]{0,0,0}\makebox(0,0)[rt]{\lineheight{1.25}\smash{\begin{tabular}[t]{r}$\sqrt{1-\bar{\alpha}_t} \epsilon$\end{tabular}}}}%
    \put(0.07133872,0.18636092){\color[rgb]{0,0,0}\makebox(0,0)[lt]{\lineheight{1.25}\smash{\begin{tabular}[t]{l}$\sqrt{1-\bar{\alpha}_T} \epsilon$\end{tabular}}}}%
    \put(0.2626825,0.00853878){\color[rgb]{0,0,0}\makebox(0,0)[t]{\lineheight{1.25}\smash{\begin{tabular}[t]{c}$p_\theta^\epsilon(\cdot \mid x_T)$\end{tabular}}}}%
    \put(0.37036247,0.0456448){\color[rgb]{0,0,0}\makebox(0,0)[lt]{\lineheight{1.25}\smash{\begin{tabular}[t]{l}$p_\theta^\epsilon(\cdot \mid x_t)$\end{tabular}}}}%
    \put(0,0){\includegraphics[width=\unitlength,page=3]{DistrDiffusionDrawing.pdf}}%
    \put(0.58402959,0.19010036){\color[rgb]{0,0,0}\makebox(0,0)[lt]{\lineheight{1.25}\smash{\begin{tabular}[t]{l}$p_\theta^\epsilon(\cdot \mid x_t)$\end{tabular}}}}%
    \put(0.86701237,0.00529646){\color[rgb]{0,0,0}\makebox(0,0)[lt]{\lineheight{1.25}\smash{\begin{tabular}[t]{l}$\epsilon_t$\end{tabular}}}}%
    \put(0.88492662,0.16175225){\color[rgb]{0,0,0}\makebox(0,0)[lt]{\lineheight{1.25}\smash{\begin{tabular}[t]{l}Mixture\end{tabular}}}}%
    \put(0.88423962,0.12875575){\color[rgb]{0,0,0}\makebox(0,0)[lt]{\lineheight{1.25}\smash{\begin{tabular}[t]{l}Gaussian\end{tabular}}}}%
    \put(0.88478674,0.09575925){\color[rgb]{0,0,0}\makebox(0,0)[lt]{\lineheight{1.25}\smash{\begin{tabular}[t]{l}Dirac\end{tabular}}}}%
    \put(0.88274385,0.06469013){\color[rgb]{0,0,0}\makebox(0,0)[lt]{\lineheight{1.25}\smash{\begin{tabular}[t]{l}Ground \\Truth\end{tabular}}}}%
    \put(0,0){\includegraphics[width=\unitlength,page=4]{DistrDiffusionDrawing.pdf}}%
  \end{picture}%
\endgroup%

%% file: ICLR_Submission/methodology.tex
Our goal is to improve uncertainty quantification in regression tasks with diffusion models by generalizing the diffusion loss, thereby enabling more accurate approximation of the conditional distribution $p_\mathcal{Y}(\cdot \mid \bm c)$.

\subsection{Learning the full distribution in each denoising step}

Most diffusion frameworks, following DDPM, fix the variance of the noise distribution in each denoising step. This design was originally motivated by two observations: (i) learning the variance often destabilized training, and (ii) variance modeling showed little benefit for image generation benchmarks, for example, with respect to the FID \cite{heusel2018ganstrainedtimescaleupdate}.  

However, subsequent work \citep{nicholImprovedDenoisingDiffusion2021} demonstrated that learning the variance improves likelihood estimates, indicating that recovering only the mean is insufficient for faithfully approximating the conditional distribution. Furthermore, the Gaussian approximation of $p(\bm x_{t-1} \mid \bm x_t)$ is only valid when the number of timesteps $T$ is large. Yet, large $T$ is computationally costly, and recent results suggest that using as few as 20--50 steps can yield superior performance in regression tasks \citep{kohl_benchmarking_2024, priceProbabilisticWeatherForecasting2025}, particularly with improved noise schedulers and solvers \citep{karrasElucidatingDesignSpace2022, 9879691}.  

These observations motivate us to go beyond estimating the first two moments and instead learn the full distribution of $\epsilon_t$. Concretely, we reinterpret $\epsilon_\theta$: rather than treating it as a point estimate of $\epsilon_t$, we view it as a random variable. This perspective naturally suggests replacing the mean-squared error loss $L_{\text{VLB}}$ with a criterion that compares probability distributions rather than point predictions. To this end, we adopt the framework of strictly proper scoring rules \citep{https://doi.org/10.1111/sjos.12168, JMLR:v25:23-0038}.

Formally, let $\cP$ be a convex set of probability measures on $\mathcal{Y}$ and $\bP, \bQ \in \cP$.  A scoring rule $S(\bP, \bm y)$ is a function that measures the discrepancy between a predictive distribution $\bP$ and an observation $\bm y \in \mathcal{Y}$ \citep{Gneiting.2007}. The expected score is defined as $S(\bP, \bQ) \coloneq \bE_{ Y \sim \bQ}[S(\bP,  Y)]$. A scoring rule is called \emph{proper} if $S(\bQ, \bQ) \leq S(\bP, \bQ)$ for all $\bP, \bQ \in \cP$, and \emph{strictly proper} if equality holds if and only if $\bQ = \bP$. Therefore, strictly proper scoring rules ensure that the true distribution is the unique optimum and have been successfully applied in training neural networks for various tasks \citep{JMLR:v25:23-0038, bultepno, Chen.2022}. Motivated by this, we propose the loss
\begin{equation} \label{eq:L_SR}
    L_{SR}=\mathbb{E}_{t,x_0,\epsilon_t}[S(p_\theta^\epsilon(\cdot\mid \bm x_t),\epsilon_t)]
\end{equation}
where $p_\theta^\epsilon$ is a neural network–based model of the predictive distribution of $\epsilon_t$ and $S$ denotes a strictly proper scoring rule. This objective enables learning general noise distributions while ensuring that training accounts for the entire distribution rather than just its moments. \\
Section~\ref{sec:theoretical-justification} connects this formulation to recent theoretical advances that justify our approach, and Section~\ref{sec:methodology-models-and-losses} details design choices for both the noise distribution $p_\theta^\epsilon$ and scoring rule $S$.

\subsection{Theoretical justification} \label{sec:theoretical-justification}
Concurrent work by \citet{bortoli2025distributional} reached a similar conclusion, proposing the use of proper scoring rules as a principled way to learn the full posterior distribution over noisy samples. Consider the reverse transition
\begin{equation}    \label{background:diffusion:backward_formula}
    p(\bm x_{t-1} \mid \bm x_t) 
    = \int_{\Ree^d} p(\bm x_{t-1} \mid \bm x_0, \bm x_t)\, p(\bm x_0 \mid \bm x_t)\, d\bm x_0,
\end{equation}
where the intractable posterior $p(\bm x_0 \mid \bm x_t)$ must be approximated.  

In standard DDPM/DDIM frameworks, this posterior is replaced by a point mass,  
$p(\bm x_0 \mid \bm x_t) \approx \delta_{\hat{\bm x_0}(t,\bm x_t)}$,  
where $\hat{\bm x}(t,\bm x_t)$ is the prediction of the denoising network. Trained with a regression loss, the denoiser recovers  
$\hat{\bm x}(t,\bm x_t) \approx \bE[X_0 \mid X_t=\bm x_t]$, yielding the DDIM formulation.  
For $T \to \infty$, this procedure recovers the data distribution. However, when using only a small number of timesteps---a setting of practical interest---the approximation may no longer hold.  

To address this, \citet{bortoli2025distributional} proposed learning the full posterior distribution $p(\bm x_0 \mid \bm x_t)$ via proper scoring rules. They mainly focus on \emph{generalized kernel scores}, which are of the form

\begin{equation}    
\label{eq:generalized_kernel_score}
S_{\lambda, \rho}(p,\bm y) = \bE_p [\rho(X,\bm y)]-\frac{\lambda}{2} \bE_{p \otimes p}[\rho(X,X')] - \frac{1}{2}\rho(\bm y,\bm y),
\end{equation}
for kernel $\rho$. Special cases include the \emph{energy score} \citep{Gneiting.2007} for  
$\rho(\bm x',\bm x)=\|\bm x'-\bm x\|^\beta$, $\beta \in (0,2)$,  
and the  \emph{(Gaussian) kernel score} with  
$\rho(\bm x',\bm x)=-\exp(-\|\bm x-\bm x'\|^2/\gamma^2)$.  
Importantly, \citet{bortoli2025distributional} show that both recover the classical diffusion regression loss in the limit when combining \eqref{eq:L_SR} with \eqref{eq:generalized_kernel_score}, a property termed \emph{diffusion compatibility}. Furthermore, even for scoring rules lacking this property, the authors report substantial empirical gains. 

A second design choice concerns the parametrization of $p(\bm x_0 \mid \bm x_t)$.  
Following \citet{10.1093/jrsssb/qkae108}, \citet{bortoli2025distributional} propose to concatenate Gaussian noise to the neural network input, thereby generating $M$ samples of the target distribution. Training is then carried out using an unbiased estimator of Equation~\eqref{eq:generalized_kernel_score}. This strategy provides a flexible, nonparametric estimate of $p(\bm x_0 \mid \bm x_t)$, but at significant computational cost: multiple samples $M$ increase both the input dimensionality and runtime. Empirically, training is reported to be $1.3\times$ to $7\times$ slower than standard diffusion models, even for moderate $M$ \citep{bortoli2025distributional}, and may become prohibitive for large-scale datasets.

\subsection[Parametrization of p]{Parametrization of $p_\theta^\epsilon(\cdot \mid \bm x_t)$}
\label{sec:methodology-models-and-losses}
As an alternative to nonparametric sampling-based approaches, we propose to model $p_\theta^\epsilon(\cdot \mid \bm x_t)$ directly through a parametrized distribution. This aligns naturally with scoring rule minimization, since closed-form expressions for the training objective are available. A schematic overview is given in Figure~\ref{fig:overview}.  
Depending on the choice of parametrization, one obtains a trade-off between computational efficiency and flexibility, which may vary across tasks: while simple Gaussian models may suffice in some applications, more complex structures are beneficial for multimodal or correlated noise patterns.  

Specifically, we consider the general Gaussian mixture form
\begin{equation}
    \label{eq:noise_parametrization}
    p_\theta^\epsilon(\epsilon_{t} \mid \bm x_t)=\sum_{k=1} ^K\pi_{\theta,k}\mathcal{N}(\epsilon_{t};\bm \mu_{\theta,k}^\epsilon(\bm x_t,t),\bm\Sigma_{\theta,k}^\epsilon(\bm x_t,t)),
\end{equation}
with component means $\bm\mu_{\theta,k}^\epsilon(\bm x_t,t)$, positive-definite covariance matrices $\bm\Sigma_{\theta,k}^\epsilon(\bm x_t,t)$, and mixture weights $\pi_{\theta,k}\in[0,1]$ satisfying $\sum_{k=1}^K \pi_{\theta,k}=1$. From this general specification, we highlight three concrete parametrizations that offer a trade-off between the expressivity of the distribution and the simplicity of the model training:

\textbf{Univariate Gaussian:} For $K=1$ and $\bm \Sigma_\theta(\bm x_t,t) = \mathrm{diag}(\sigma_\theta^2(\bm x_t,t)), \ \sigma_\theta^2(\bm x_t, t)\in\mathbb{R}^{d_y}_{>0}$, we obtain a univariate Gaussian parametrization for each coordinate in the noise space.

\textbf{Univariate Gaussian mixture.}  
    Setting $\bm\Sigma_{\theta,k}(\bm x_t,t) = \mathrm{diag}(\sigma_{\theta,k}^2(\bm x_t,t))$ with $\sigma_{\theta,k}^2(\bm x_t, t)\in\mathbb{R}^{d_y}_{>0}$ and $K>1$ enables multimodal modeling, as Gaussian mixtures can approximate arbitrary continuous densities to arbitrary precision under mild assumptions \citep{bishop,inbook}, offering significantly greater expressivity at moderate cost.  
    
\textbf{Multivariate Gaussian:}  Choosing $K=1$ with full covariance $\bm \Sigma_\theta(\bm x_t,t)$ allows modeling correlations between components of $\epsilon_t$. Since a full parametrization is infeasible for high-dimensional tasks, we consider two approximations: \emph{Cholesky-based} and \emph{low-rank plus diagonal}, which are described in detail in Appendix~\ref{app:covariance_approximations}.

All three model variants are compatible with the scoring rule framework, where we focus on $\lambda=1$ in Equation~\eqref{eq:generalized_kernel_score}, corresponding to a strictly proper scoring rule. For each case, we obtain closed-form expressions for all models in Appendix~\ref{app:closed_form}. Moreover, our parametrizations admit a closed-form backward distribution, which enables straightforward sampling.  

\begin{theorem} \label{methodology:thm:closed_form}
    Let $p_\theta^\epsilon(\epsilon_t \mid \bm x_t)$ be given as in Equation~\eqref{eq:noise_parametrization}.
    Then the reverse distribution in Equation~\eqref{background:diffusion:backward_formula} admits the closed form
    \begin{equation*}
        \label{eq:closed_form_expression}
        p_\theta(\bm x_{t-1}\mid\bm x_t)=\sum_{k=1} ^K\pi_{\theta,k}\mathcal{N}
        \left(\bm x_{t-1};
         \sqrt{\bar{\alpha}_{t-1}} \hat{\bm x}_0 + 
        \lambda_t\bm\mu_{\theta,k}^\epsilon(\bm x_t,t) ,
        \gamma_t^2 \bm\Sigma_{\theta,k}^\epsilon(\bm x_t,t) + \sigma_t^2 \mathbf{I}
        \right),
    \end{equation*}
    where $\hat{\bm x}_0 = \frac{\bm x_t - \sqrt{1 - \bar{\alpha}_t} \bm\mu_{\theta,k}^\epsilon(\bm x_t, t)}
    {\sqrt{\bar{\alpha}_t}}$, $\lambda_t \coloneq \sqrt{1 - \bar{\alpha}_{t-1} - \sigma_t^2}$ and $\gamma_t \coloneq \lambda_t  - 
    \frac{\sqrt{1 - \bar{\alpha}_t}}{\sqrt{\alpha_t}}$.
\end{theorem}

The proof can be found in Appendix~\ref{app:proof_of_thm1}.

Applying this theorem to our univariate Gaussian instantiation and setting $\sigma_t = \tilde{\beta}_t$, i.e. $\eta = 1$, we obtain a DDPM model that learns the variance for each output dimension, similar to what was proposed by \cite{nicholImprovedDenoisingDiffusion2021}.
Remarkably, in both formulations, the variance is greater than $\tilde{\beta}_t$.
Notice, however, that our approach was derived from a principled way of modeling the diffusion noise as a normal distribution and can use any scoring rule as a loss function.